\documentclass{article}

\PassOptionsToPackage{numbers, compress}{natbib}

\usepackage[main, final]{neurips_2026}

\usepackage[utf8]{inputenc} 
\usepackage[T1]{fontenc}    
\usepackage{hyperref}       
\usepackage{url}            
\usepackage{booktabs}       
\usepackage{amsfonts}       
\usepackage{nicefrac}       
\usepackage{microtype}      
\usepackage{xcolor}         

\usepackage{microtype}
\usepackage{graphicx}
\usepackage{booktabs}
\usepackage{hyperref}
\usepackage{amsmath}
\usepackage{amssymb}
\usepackage{mathtools}
\usepackage{amsthm}
\usepackage{algorithm}
\usepackage{algorithmic}
\usepackage{multirow}
\usepackage{xcolor}
\usepackage{colortbl}
\usepackage{subcaption}
\usepackage{bm}
\usepackage{float}
\usepackage{makecell}

\newtheorem{theorem}{Theorem}[section]
\newtheorem{proposition}[theorem]{Proposition}

\usepackage[textsize=tiny]{todonotes}

\newcommand{\R}{\mathbb{R}}
\newcommand{\E}{\mathbb{E}}
\newcommand{\vect}[1]{\mathbf{#1}}
\newcommand{\mat}[1]{\mathbf{#1}}
\newcommand{\method}{SDFlow}

\usepackage{enumitem}
\setlist[itemize]{topsep=0.3pt, parsep=0pt, itemsep=0.4pt, partopsep=0.2pt}
\setlist[enumerate]{topsep=0.7pt, parsep=0pt, itemsep=0.4pt, partopsep=0.2pt}
\makeatletter
\renewcommand{\section}{\@startsection{section}{1}{\z@}%
	{-1.3ex \@plus -0.4ex \@minus -.2ex}%
	{0.78ex \@plus .15ex}%
	{\normalfont\Large\bfseries}}
\renewcommand{\subsection}{\@startsection{subsection}{2}{\z@}%
	{-0.95ex \@plus -0.30ex \@minus -.18ex}%
	{0.55ex \@plus .12ex}%
	{\normalfont\large\bfseries}}
\renewcommand{\subsubsection}{\@startsection{subsubsection}{3}{\z@}%
	{-0.78ex \@plus -0.22ex \@minus -.15ex}%
	{0.42ex \@plus .10ex}%
	{\normalfont\normalsize\bfseries}}
\makeatother

\title{SDFlow: Similarity-Driven Flow Matching for Time Series Generation}

\author{
	Wei Li$^{1,2}$\thanks{Equal contribution.}, ~
	Shibo Feng$^{3,*}$, ~
	Pengcheng Wu$^{3}$, ~
	Xingyu Gao$^{4,5}$, ~
	Min Wu$^{6}$, ~
	Peilin Zhao$^{1}$\thanks{Corresponding Author} \\
	$^{1}$Shanghai Jiao Tong University, \quad
	$^{2}$Shanghai University, \quad
	$^{3}$Nanyang Technological University \\
	$^{4}$Institute of Microelectronics, Chinese Academy of Sciences \\
	$^{5}$University of Chinese Academy of Sciences \\
	$^{6}$Institute for Infocomm Research, A*STAR, Singapore \\
	{\ttfamily\small liwei008009@163.com},
	{\ttfamily\small \{Shibo001, pengchengwu\}@ntu.edu.sg},
	{\ttfamily\small gxy9910@gmail.com} \\
	{\ttfamily\small wumin@i2r.a-star.edu.sg, peilinzhao@sjtu.edu.cn}
}

\begin{document}

	\maketitle

	\begin{abstract}
		
		Vector quantization (VQ) with autoregressive (AR) token modeling is a widely adopted and highly competitive paradigm for time-series generation. However, such models are fundamentally limited by exposure bias: during inference, errors can accumulate across sequential predictions, leading to pronounced quality degradation in long-horizon generation. To address this, we propose \method{} (\textbf{S}imilarity-\textbf{D}riven \textbf{Flow} Matching), a non-autoregressive framework that operates entirely in the frozen VQ latent space and enables parallel sequence generation via flow matching. We tackle three key challenges in making this transition: (1) eliminating exposure bias by replacing step-wise token prediction with a global transport map; (2) mitigating the high-dimensionality of VQ token spaces via a low-rank manifold decomposition with a learned anchor prior over the latent manifold; and (3) incorporating discrete supervision into continuous transport dynamics by introducing a categorical posterior over codebook indices within a variational flow-matching formulation. Extensive experiments show that \method{} achieves state-of-the-art performance, improving Discriminative Score and substantially reducing Context-FID, particularly for challenging long-sequence generation. Moreover, \method{} provides significant inference speedups over autoregressive baselines, offering both high fidelity and computational efficiency. Code is available at \url{https://github.com/William-Liwei/SDFlow}
		
	\end{abstract}

	\section{Introduction}
	\label{sec:intro}
	
	Time series generation underpins critical applications spanning finance~\citep{ding2020hierarchical}, healthcare~\citep{penfold2013use}, and energy systems~\citep{lim2021time}. Among generative paradigms, variational autoencoders (VAEs) offer fast one-step generation that bypasses the lengthy iterative sampling required by diffusion-based models while remaining competitive in both forecasting and generation quality. Within this lightweight paradigm, vector quantization approaches such as VQ-VAE~\citep{van2017neural} and VQGAN~\citep{esser2021taming} have proven particularly effective, compressing multivariate sequences into finite codebooks that remove redundancy, suppress noise, and enable scalable high-fidelity reconstruction.

	\begin{figure*}[t!]
		\centering
		\begin{subfigure}{0.30\textwidth}
			\includegraphics[width=\textwidth]{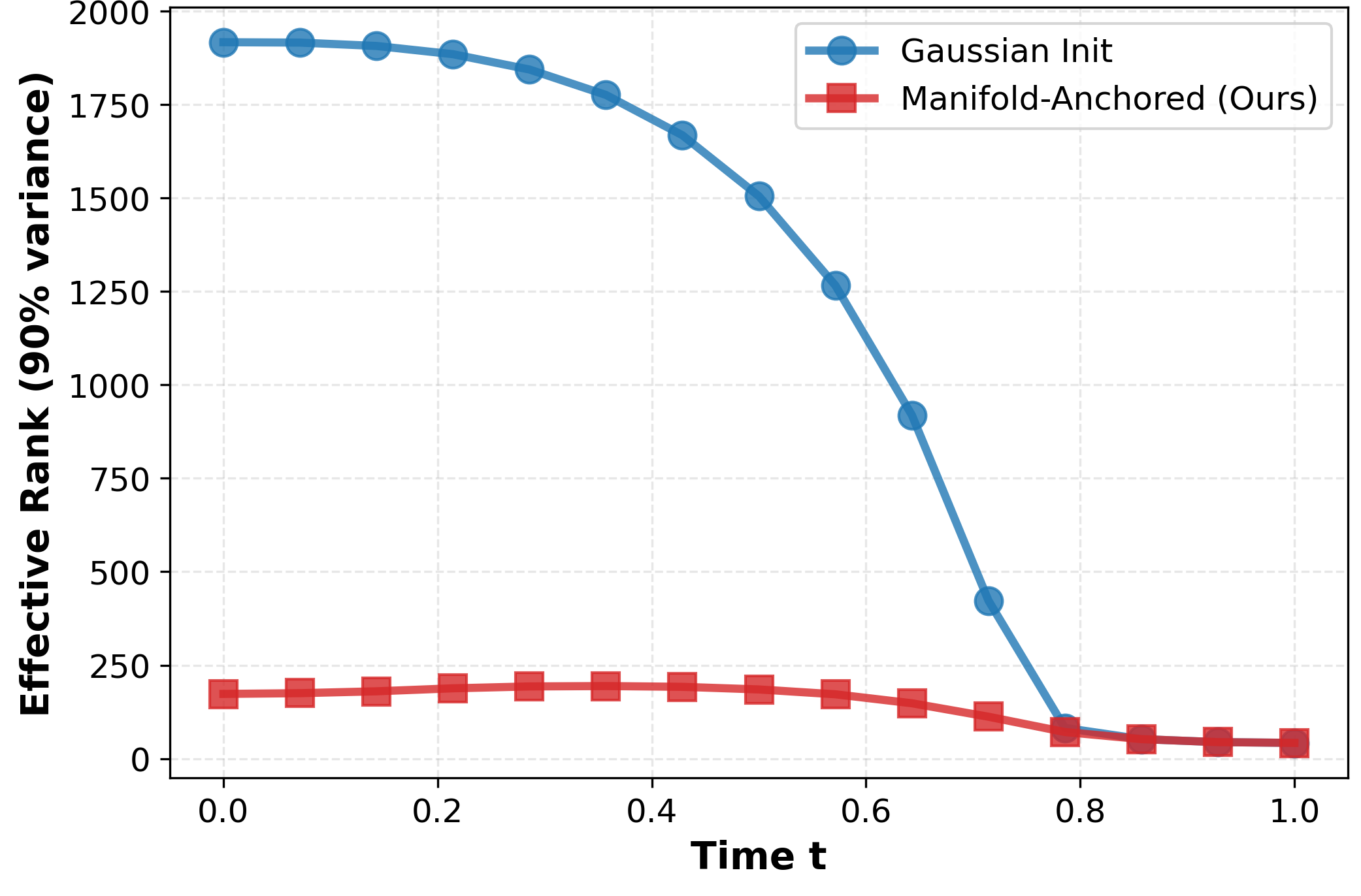}
			\caption{Space: Intrinsic Geometry}
			\label{fig:teaser_space}
		\end{subfigure}
		\hfill
		\begin{subfigure}{0.30\textwidth}
			\includegraphics[width=\textwidth]{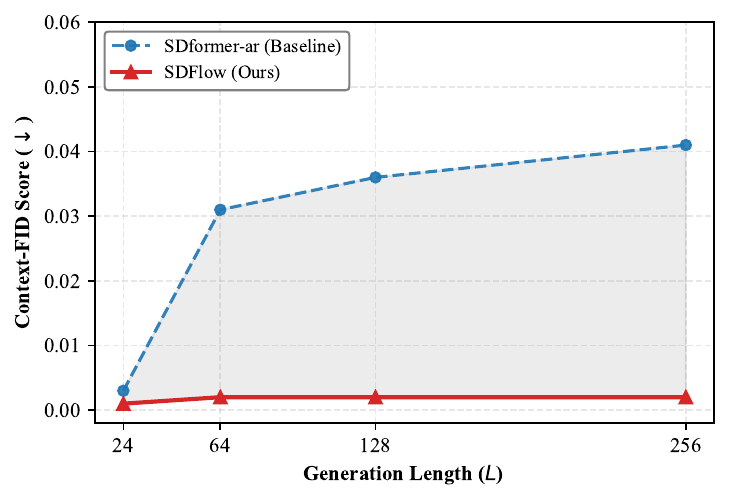}
			\caption{Time: Temporal Consistency}
			\label{fig:teaser_time}
		\end{subfigure}
		\hfill
		\begin{subfigure}{0.37\textwidth}
			\includegraphics[width=\textwidth]{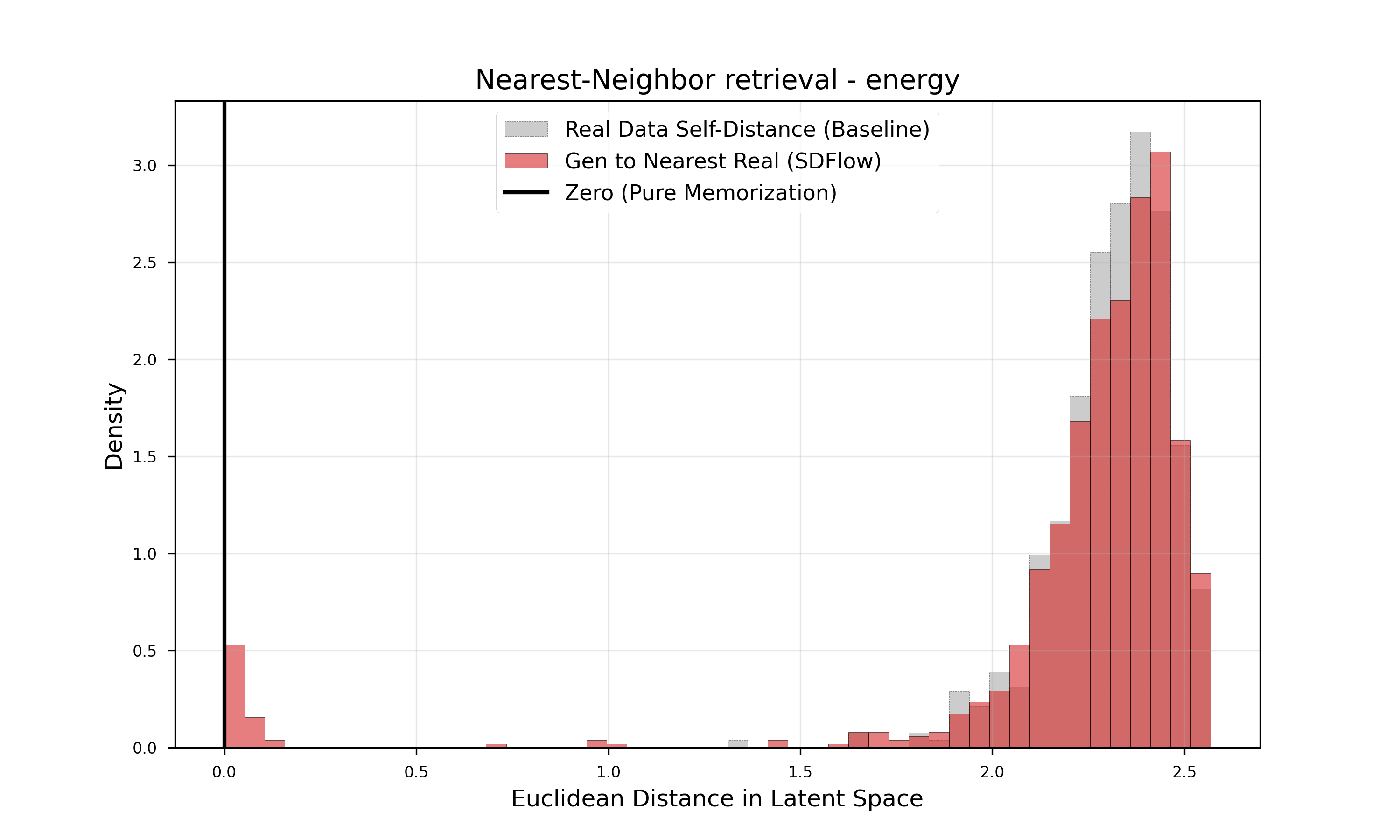}
			\caption{Generalization: Novelty}
			\label{fig:teaser_gen}
		\end{subfigure}
		\caption{
			\textbf{The Three Pillars of \method{}.} 
			\textbf{(a) Space:} Gaussian initialization (\textcolor{blue}{blue}) starts from a high-rank space far from the data, whereas our manifold-anchored approach (\textcolor{red}{red}) initializes within the intrinsic low-rank subspace, making transport computationally tractable. 
			\textbf{(b) Time:} Unlike autoregressive baselines (\textcolor{blue}{blue}) that suffer from exposure bias on long sequences, \method{} (\textcolor{red}{red}) maintains consistent high fidelity regardless of length. 
			\textbf{(c) Generalization:} Nearest-neighbor distance distributions show that the generated-to-nearest-real distances (\textcolor{red}{red}) closely match the real-to-nearest-real distances (\textcolor{gray}{gray}), supporting non-copying behavior.
		}
		\label{fig:teaser}
	\end{figure*}

	SDformer~\citep{chen2024sdformer} recently established state-of-the-art performance through VQ-VAE tokenization with similarity-driven quantization combined with autoregressive transformer generation. However, advancing this paradigm to robust long-term generation faces three challenges:

	\textbf{1. Exposure Bias in Autoregressive Modeling.} 
	AR models suffer from a fundamental train-test discrepancy~\citep{bengio2015scheduled,schmidt2019generalization}. During training, the model conditions on ground-truth tokens, but during inference, it must rely on its own predictions. Errors accumulate sequentially during inference, causing significant degradation in long-horizon generation. This is empirically verified in Figure~\ref{fig:teaser}(b), where the baseline's fidelity deteriorates sharply as sequence length increases.
	
	\textbf{2. The Curse of Dimensionality.} 
	While non-AR flow matching avoids sequential errors, SDFlow still operates in the frozen VQ latent space rather than raw data space. As shown in Figure~\ref{fig:teaser}(a), the latent data manifold is extremely low-rank ($r \ll D$) compared to the high-dimensional ambient space. Standard Gaussian initialization starts far from this manifold, making the learning of transport dynamics computationally intractable.
	
	\textbf{3. Discrete Supervision for Continuous Transport.} VQ-VAE latents possess a dual discrete-continuous structure: each position corresponds to a discrete codebook index with an associated continuous embedding. Purely continuous flow matching ignores the categorical structure, while purely discrete approaches collapse geometry. An effective solution must provide categorical supervision while maintaining continuous transport dynamics. Crucially, the relevant generalization failure mode is latent-anchor collapse rather than raw data-space interpolation. SDFlow learns a low-rank anchor manifold and uses a kernel-smoothed prior only as a tractable way to initialize flow near that scaffold. As shown in Figure \ref{fig:teaser}(c), generated samples maintain substantial distances from training data, suggesting nontrivial synthesis and generalization.
	
	To address this triad of challenges, we propose \method{}, making three key innovations. 
	\textbf{First}, we learn a low-rank manifold decomposition that discovers the intrinsic subspace structure, enabling manifold-anchored initialization that reduces transport distance. 
	\textbf{Second}, following variational flow matching~\citep{eijkelboom2024variational} 
	, we employ categorical posteriors over codebook indices that provide explicit discrete supervision through cross-entropy loss while computing velocity fields in continuous embedding space. 
	\textbf{Third}, we use a learned anchor prior over the low-dimensional anchor coordinates as an auxiliary sampler for topology-preserving initialization.
	
	In summary, our contributions include:
	\begin{itemize}
		\item \textbf{Analysis:} We identify exposure bias and the curse of dimensionality as fundamental barriers in VQ-based time series generation.
		
		\item \textbf{Method:} \method{} combines low-rank manifold discovery with categorical flow matching for non-autoregressive parallel generation.
		
		\item \textbf{Evaluation:} SDFlow achieves the state-of-the-art performance in short- and long-term generation tasks, while offering 3--10$\times$ speedup improvements.
	\end{itemize}
	
	\section{Related Work}
	\label{sec:related}
	
	\textbf{Time Series Generation.}
	GAN-based methods like TimeGAN~\citep{yoon2019time} and COT-GAN~\citep{xu2020cot} pioneered deep generative modeling for time series but suffer from training instability and mode collapse. Diffusion approaches~\citep{kong2021diffwave,coletta2024constrained,yuan2024diffusion,tashiro2021csdi,alcaraz2022diffusion,feng2024latent} achieve high fidelity but require hundreds of denoising steps, limiting practical applicability. SDformer~\citep{chen2024sdformer} established state-of-the-art through discrete tokens with autoregressive transformers, demonstrating the power of VQ-VAE representations for temporal data. We build upon this foundation while addressing the fundamental limitations of sequential generation.
	
	\textbf{Discrete Token Modeling and Exposure Bias.}
	VQ-VAE~\citep{van2017neural} and VQGAN~\citep{esser2021taming} enable discrete representations for continuous data, forming the backbone of modern generative models. Autoregressive methods like DALL-E~\citep{ramesh2021zero} and GPT-based approaches generate tokens sequentially with powerful language model priors. However, all autoregressive methods suffer from exposure bias~\citep{bengio2015scheduled}: models are trained via teacher forcing with ground-truth conditioning but must rely on self-generated tokens at inference. This train-test mismatch leads to cascading errors where early mistakes compound through the sequence~\citep{schmidt2019generalization}. Various mitigation strategies have been proposed, including scheduled sampling~\citep{bengio2015scheduled} and random token replacement~\citep{zhang2023t2m}, but these only partially address the fundamental issue.
	Non-AR alternatives like MaskGIT~\citep{chang2022maskgit} employ parallel decoding through iterative refinement, reducing but not eliminating sequential dependencies. 
	
	\textbf{Flow Matching and Variational Extensions.}
	Flow matching~\citep{lipman2023flow,liu2023flow,albergo2023building,tong2024improving} provides simulation-free training for continuous normalizing flows, enabling efficient learning of transport maps. Recent work has also explored flow-based methods for time series, including rectified flows~\citep{hu2025flowts} and stochastic interpolants~\citep{albergo2023stochastic,chen2024flow}. Variational flow matching (VFM)~\citep{eijkelboom2024variational} reframes the problem as variational inference, enabling flexible posterior families beyond Gaussian. For discrete or quantized data, VFM yields categorical posteriors that provide direct supervision while maintaining continuous transport---an approach termed CatFlow and successfully applied to molecular generation. 
	Recently, Purrception~\citep{matisan2025purrception} integrates discrete categorical targets to steer continuous generative trajectories in image synthesis.

	\section{Preliminaries}
	\label{sec:prelim}
	
	We establish notation and review the technical foundations underlying our approach.
	
	\textbf{Problem Formulation.}
	Let $\mat{X}_{1:\ell} = (\vect{x}_1, \ldots, \vect{x}_\ell) \in \R^{\ell \times d}$ denote a multivariate time series of length $\ell$ with $d$ observed dimensions. The unconditional generation problem is defined as:
	\begin{equation}
		\text{Input: } \vect{Z}_0 \sim \pi_0; \quad \text{Output: } \hat{\mat{X}}_{1:\ell} = \mathcal{G}(\vect{Z}_0) \in \R^{\ell \times d},
	\end{equation}
	where $\vect{Z}_0 \in \R^{\ell \times d}$, $\pi_0 = \mathcal{N}(\vect{0}, \mat{I})$ is a tractable prior, and $\mathcal{G}$ maps samples from the prior to the target data distribution. The generative model can be implemented using GANs~\citep{yoon2019time}, VAEs~\citep{desai2021timevae}, or diffusion models~\citep{yuan2024diffusion}, which effectively capture complex temporal dependencies. During training, $\mathcal{G}$ is optimized to minimize discrepancy between generated $\hat{\mat{X}}_{1:\ell}$ and real data $\mat{X}_{1:\ell}$, ensuring high fidelity and diversity in generated sequences.
	
	\textbf{Similarity-Driven Vector Quantization.}
	Following~\citep{chen2024sdformer}, we employ a VQ-VAE tokenizer that maps continuous time series to discrete token sequences. An encoder $\mathcal{E}$ produces latent vectors $\mat{H}_{1:L} = \mathcal{E}(\mat{X}_{1:\ell}) \in \R^{L \times d_c}$, where $L = \ell/s$ for temporal downsampling rate $s$. Each latent vector is quantized to its nearest neighbor in a learned codebook $\mathcal{C} = \{\vect{c}_k\}_{k=0}^{K-1} \subset \R^{d_c}$ using \emph{similarity-driven} quantization:
	\vspace{-0.1cm}
	\begin{equation}
		\label{eq:similarity_driven}
		y_i = Q(\vect{h}_i) := \arg\max_{k=0,\ldots,K-1} \frac{\vect{h}_i}{\|\vect{h}_i\|} \cdot \frac{\vect{c}_k}{\|\vect{c}_k\|}.
	\end{equation}
	This cosine similarity measure, with unit-normalized codes, improves codebook utilization by preventing collapse to high-norm regions. The encoder output is normalized to unit length, simplifying quantization to $y_i = \arg\max_k \vect{h}_i \cdot \vect{c}_k$. Dequantization recovers embeddings via $\tilde{\vect{h}}_i = Q^{-1}(y_i) := \vect{c}_{y_i}$, and the decoder $\mathcal{D}$ reconstructs the time series: $\tilde{\mat{X}}_{1:\ell} = \mathcal{D}(\tilde{\mat{H}}_{1:L})$. Training minimizes reconstruction loss plus embedding regularization with exponential moving average (EMA) codebook updates and inactive code resetting.
	
	\textbf{Exposure Bias in Autoregressive Token Modeling.}
	SDformer's second stage employs autoregressive token modeling to learn the distribution at the discrete token level. During training, shifted tokens $\mat{Y}^{in}_{1:L} = ([\text{BOS}], y_1, \ldots, y_{L-1})$ serve as input to predict targets $\mat{Y}_{1:L}$, where $[\text{BOS}]$ denotes the beginning-of-sequence token. The training objective minimizes negative log-likelihood:
	\begin{equation}
		\mathcal{L}_{ar} = -\E\left[\sum_i \log P(y_i | \mat{Y}^{in}_{1:i})\right].
	\end{equation}
	At inference, generation proceeds autoregressively from $[\text{BOS}]$, sampling each token from the predicted distribution. This creates a fundamental train-test discrepancy: training conditions on ground-truth tokens while inference conditions on self-generated tokens. Early prediction errors propagate through the sequence, causing cascading degradation---a phenomenon particularly severe for long sequences where errors accumulate over many steps.
	
	\textbf{Flow Matching.}
	Flow matching~\citep{lipman2023flow} learns a time-dependent velocity field $\vect{v}_\theta: \R^D \times [0,1] \to \R^D$ that generates samples by integrating the ordinary differential equation:
	\begin{equation}
		\frac{d\vect{z}_t}{dt} = \vect{v}_\theta(\vect{z}_t, t), \quad \vect{z}_0 \sim p_0, \quad \vect{z}_1 \sim p_{\text{data}}.
	\end{equation}
	For the linear interpolation path $\vect{z}_t = (1-t)\vect{z}_0 + t\vect{z}_1$, the conditional velocity field is $\vect{v}_t(\vect{z}|\vect{z}_1) = \vect{z}_1 - \vect{z}_0$. Training minimizes:
	\begin{equation}
		\mathcal{L}_{\text{FM}} = \E_{t, \vect{z}_0, \vect{z}_1}\left[ \|\vect{v}_\theta(\vect{z}_t, t) - (\vect{z}_1 - \vect{z}_0)\|^2 \right].
	\end{equation}
	This formulation provides simulation-free training and enables parallel generation---no sequential dependencies, thus completely eliminating exposure bias.
	
	\section{Method}
	\label{sec:method}
	
\begin{figure*}[t]
		\centering
		\includegraphics[width=0.9\textwidth]{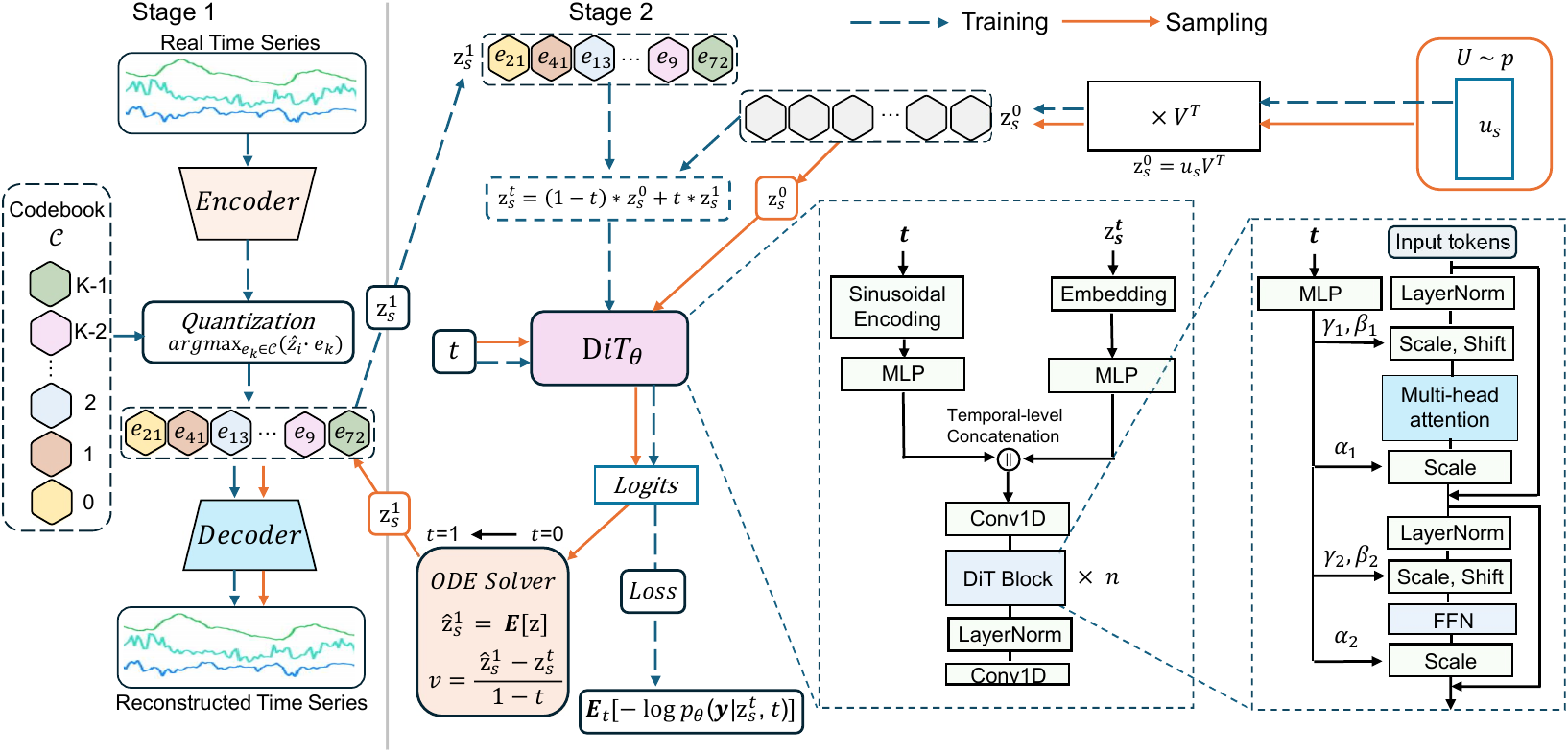}
		\caption{Overview of the \method{} framework. \textbf{Stage 1} pre-trains a VQ-VAE tokenizer with similarity-driven quantization (frozen during Stage 2). \textbf{Stage 2} learns manifold-anchored flow matching in the frozen VQ latent space: low-rank decomposition discovers the intrinsic anchor manifold, a learned anchor prior provides topology-preserving initialization, and categorical posteriors over codebook indices enable discrete supervision with continuous transport.}
		\label{fig:architecture}
	\end{figure*}

	\method{} is a two-stage framework illustrated in Figure~\ref{fig:architecture}. Stage 1 pre-trains a VQ-VAE tokenizer with similarity-driven quantization (frozen during Stage 2). Stage 2 learns manifold-anchored flow matching with categorical supervision. We now detail each component.

	\subsection{Similarity-Driven VQ-VAE Tokenization}
	\label{sec:vqvae}
	
	We adopt the similarity-driven VQ-VAE architecture from~\cite{chen2024sdformer}, which has demonstrated strong performance for time series. The encoder uses 1D convolutions with residual blocks to map input time series $\mat{X}_{1:\ell} \in \R^{\ell \times d}$ to latent vectors $\mat{H}_{1:L} \in \R^{L \times d_c}$, where $L = \ell/s$ is the compressed sequence length. As shown in Eq.~\eqref{eq:similarity_driven}, The key innovation is similarity-driven quantization: rather than Euclidean distance, quantization maximizes cosine similarity between encoder outputs and codebook vectors.
	This normalization-based approach prevents codebook collapse where only a few high-norm codes are utilized. Training employs the standard VQ-VAE objective:
	\begin{equation}
		\mathcal{L}_{\text{VQ}} = \|\mat{X}_{1:\ell} - \tilde{\mat{X}}_{1:\ell}\|_2^2 + \frac{\lambda}{L} \sum_{i=1}^L \left(1 - \vect{h}_i \cdot \text{sg}(\tilde{\vect{h}}_i)\right),
	\end{equation}
	where $\text{sg}(\cdot)$ denotes stop-gradient and the second term is the embedding loss. Codebook updates use exponential moving average (EMA) with inactive code resetting to maintain high utilization.
	
	Once trained, the tokenizer is frozen and all subsequent learning occurs in the discrete token space, ensuring the latent structure remains stable during flow matching training.
	
	\subsection{The Geometric Mismatch}
	\label{sec:curse}
	
	Consider the VQ-VAE latent space with ambient dimension $D = L \times d_c > 3,000$ for typical configurations. Standard flow matching from $\vect{z}_0 \sim \mathcal{N}(\vect{0}, \mat{I}_D)$ fails due to two geometric issues: (1) \textit{Norm mismatch}: unit-normalized VQ codes yield $\|\vect{z}^*\| = \sqrt{L}$, while Gaussian samples concentrate at $\|\vect{z}_0\| \approx \sqrt{D}$, creating a norm ratio of $\sqrt{d_c} \approx 22$ (where $d_c=512$ is the embedding dimension), implying the noise prior is physically much larger than the data manifold; (2) \textit{Manifold sparsity}: the analysis (Figure~\ref{fig:teaser}) reveals that only dozens of components capture 90\% variance when $t=1$ despite $D > 3,000$, 
	which means that isotropic samples have negligible probability of landing near the data manifold. Our low-rank decomposition addresses both issues through manifold-anchored initialization.
	To address the dimensionality challenge, we learn a low-rank factorization of the latent space:
	\begin{equation}
		\mat{Z} \approx \mat{U}\mat{V}^\top, \quad \mat{U} \in \R^{M \times r}, \; \mat{V} \in \R^{D \times r},
	\end{equation}
	where $\mat{V}$ is a global basis and $\mat{U}$ is a learnable set of low-rank support coordinates for the encoded Stage-2 VQ latents. We do not prefit a separate SVD or reconstruction loss; $\mat{U},\mat{V}$ are optimized end-to-end with the flow objective and coordinate regularization. These supports only parameterize latent initialization: no raw time series are stored or retrieved, and generation samples a continuous coordinate that is subsequently transported by the learned flow. The extra memory is $O(Mr+Dr)$ with $r\ll D$, and Table~\ref{tab:ablation} shows that $M$ can be substantially reduced.
	
	\textbf{Coordinate Regularization.}
	To ensure well-behaved coordinate distributions suitable for density estimation, we regularize toward zero mean and unit variance:
	\begin{equation}
		\mathcal{L}_{\text{reg}} = \lambda_\mu \mathcal{L}_\mu + \lambda_\sigma \mathcal{L}_\sigma = \lambda_\mu \|\bar{\vect{u}}\|^2 + \lambda_\sigma (|\text{std}(\mat{U}) - 1|),
	\end{equation}
	where $\bar{\vect{u}} = \frac{1}{N}\sum_i \vect{u}_i$ is the mean coordinate and $\text{std}(\mat{U})$ is the standard deviation across samples.

	\textbf{Topology-Preserving Latent Anchor Prior.}
	Rather than fitting the transport initializer in raw data space or relying on an uninformative isotropic Gaussian prior, we construct a learned anchor prior over the low-rank latent anchors, implemented with kernel smoothing:
	\begin{equation}
		p_{\text{anchor}}(\vect{u}) = \frac{1}{N} \sum_{i=1}^N \mathcal{K}_h(\vect{u} - \vect{u}_i),
	\end{equation}
	where the bandwidth $h$ is not an arbitrary noise scale, but a topological parameter governing manifold connectivity. Crucially, since the coordinates are globally regularized to unit variance ($\text{std}(\mat{U}) \approx 1$), the average nearest-neighbor distance $\bar{d}_{\text{NN}}$ is empirically small relative to the coordinate scale. By enforcing $h \propto \bar{d}_{\text{NN}}$, this geometrically calibrated bandwidth bridges local gaps between latent supports without over-smoothing the global topology. Thus, the kernel smoother is an auxiliary continuous sampler over the support set, not a lookup generator; the learned flow carries the final generative dynamics.
	
	\textbf{Manifold-Anchored Initialization.}
	Given a sampled coordinate $\vect{u} \sim p_{\text{anchor}}$, the flow initialization is $\vect{z}_0 = \text{normalize}_{\ell_2}(\vect{u}\mat{V}^\top)$. Because $\vect{u}$ is drawn from a continuous support within the highly bottlenecked $r$-dimensional subspace ($r \ll D$), this initialization acts as a principled manifold interpolation mechanism. The learned anchor scaffold shapes where generation starts, while the transport map $v_\theta$ is responsible for producing final samples beyond those anchors.
	
	\subsection{Categorical Flow Matching for VQ-VAE Latents}
	\label{sec:vq-vfm}
	
	In VQ-VAE latent spaces, endpoints $\vect{z}_1$ must be codebook embeddings $\{\vect{c}_k\}_{k=1}^K$. Following variational flow matching~\citep{eijkelboom2024variational} 
	, we model the posterior over endpoints as categorical:
	\begin{align}
		q_\theta(c | \vect{z}_t) &= \text{Cat}(c | \pi_\theta^t(\vect{z}_t)),  \\
		\pi_\theta^{t,k}(\vect{z}_t) &= \frac{\exp(f_\theta^k(\vect{z}_t, t) / \tau)}{\sum_{j=1}^K \exp(f_\theta^j(\vect{z}_t, t) / \tau)},
	\end{align}
	where $f_\theta(\vect{z}_t, t) \in \R^{L \times K}$ outputs logits over codebook indices at each position and $\tau$ is a temperature parameter.
	
	This categorical posterior yields a continuous velocity field as a probability-weighted combination of codebook embeddings:
	Define the posterior mean embeddings
	\[
	\mu^*(z_t):=\sum_{k=1}^K p(c_k\mid z_t)\,c_k,
	\qquad
	\mu_\theta(z_t):=\sum_{k=1}^K q_\theta(c_k\mid z_t)\,c_k,
	\]
	and the induced velocity fields (for $t\in[0,1)$)
	\[
	v^*(z_t,t):=\frac{\mu^*(z_t)-z_t}{1-t},
	\qquad
	v_\theta(z_t,t):=\frac{\mu_\theta(z_t)-z_t}{1-t}.
	\]
	
	where $\mu_\theta(\vect{z}_t) = \sum_k \pi_\theta^k(\vect{z}_t) \vect{c}_k$ is the expected codebook embedding under the predicted distribution. This formulation elegantly bridges discrete and continuous: the model predicts a distribution over discrete codes, but transport occurs smoothly in embedding space.
	
	\textbf{Cross-Entropy Training Objective.}
	The variational flow matching objective reduces to cross-entropy between predicted posteriors and ground-truth indices:
	\begin{equation}
		\mathcal{L}_{\text{CE}} = -\E_{t, \vect{z}_t}\left[\log q_\theta(c^* | \vect{z}_t)\right] = -\E_{t, \vect{z}_t}\left[\sum_{i=1}^L \log \pi_\theta^{c_i^*}(\vect{z}_t)\right],
	\end{equation}
	where $c^*$ denotes the observed VQ endpoint label for each sampled path, i.e., the empirical hard-label realization of the endpoint posterior. This provides direct discrete supervision---unlike MSE regression that ignores categorical structure---while maintaining geometric awareness through probability-weighted embedding combinations.
	
	\textbf{Advantages of Categorical Supervision.}
	Cross-entropy loss provides three benefits: (1) \emph{explicit token-level supervision} that directly teaches which codebook entry to select; (2) \emph{uncertainty quantification} over plausible codes, enabling soft assignments during transport; and (3) \emph{temperature-controlled generation} where lower $\tau$ sharpens predictions for higher fidelity while higher $\tau$ increases diversity.
	
	\textbf{Why SDFlow Generalizes.} The kernel-smoothed anchor prior is a continuous initializer over low-rank VQ-latent coordinates, not a raw-data generator; Appendix~\ref{sec:appendix_kernel_theory} provides the theoretical perspective, while Section~\ref{sec:ablation} reports held-out latent-flow stress tests.
	
	\subsection{Architecture and Training}
	\label{sec:arch}
	
	\textbf{Flow Network Architecture.}
	We employ a Diffusion Transformer (DiT)~\citep{peebles2023scalable} adapted for sequential data. The input is the interpolated latent $\vect{z}_t \in \R^{L \times d_c}$ treated as a sequence of $L$ tokens. Time conditioning uses adaptive layer normalization (AdaLN): a learned MLP transforms the scalar $t$ into scale and shift parameters that modulate layer statistics. A learnable global token is prepended to aggregate sequence-level information. The output is logits over the $K$-dimensional codebook at each position.
	The training loss combines categorical cross-entropy with coordinate and structure regularization:
	\begin{equation}
		\mathcal{L} = \mathcal{L}_{\text{CE}} + \lambda_\mu \mathcal{L}_\mu + \lambda_\sigma \mathcal{L}_\sigma + 10\mathcal{L}_{\text{str}}.
	\end{equation}
	where $\lambda_\mu$ and $\lambda_\sigma$ keep $\mat{U}$ near zero mean and unit variance, and $\mathcal{L}_{\text{str}}$ aligns normalized within-batch pairwise distances; it is used only during training. During training, we sample $u\sim\mathrm{Uniform}(0,1)$ and set $t=1-\cos(\pi u^2/2)$.

	\textbf{Inference Procedure.}
	Generation proceeds in five steps: (1) Sample coordinate $\vect{u} \sim p_{\text{anchor}}$ with bandwidth $h$; (2) Compute anchor $\vect{z}_0 = \text{normalize}(\vect{u}\mat{V}^\top)$; (3) Integrate the learned ODE forward using Euler steps with temperature-scaled softmax; (4) Quantize final latent to nearest codebook entries via cosine similarity; (5) Decode through the frozen VQ-VAE decoder to obtain generated time series. The kernel-smoothed prior only provides the initialization scaffold; the learned flow carries the sample to the final output.
	
	\subsection{Theoretical Foundation}
	\label{sec:theory}

	\begin{theorem}[Transport Complexity: Gaussian vs.\ Manifold-Anchored Initialization]
		\label{thm:transport_distance}
		Let $\mathcal{D}$ be the data distribution supported on $\mathbb{R}^D$. Assume the data effectively lies near an $r$-dimensional linear subspace $\mathcal{M}$ spanned by a semi-orthogonal basis $\mat{V}\in\mathbb{R}^{D\times r}$ ($r \ll D$), such that any data sample $\vect{z} \sim \mathcal{D}$ can be decomposed as $\vect{z} = \mat{V}\vect{u}^* + \vect{\epsilon}$, where $\vect{u}^* \in \mathbb{R}^r$ is the intrinsic coordinate and $\|\vect{\epsilon}\|^2 \le \epsilon^2$ is the bounded approximation error.
		Let $\vect{z}_0$ denote the initialization point for the flow. The expected squared transport distance $\mathbb{E}[\|\vect{z} - \vect{z}_0\|^2]$ behaves as follows:
		
		\begin{enumerate}
			\item \textbf{Gaussian Initialization.} 
			If $\vect{z}_0 \sim \mathcal{N}(\vect{0}, \mat{I}_D)$ and the data is bounded (i.e., $\mathbb{E}[\|\vect{z}\|^2] = C < \infty$), then the transport distance scales linearly with the ambient dimension:
			\begin{equation}
				\label{eq:gauss_dist}
				\mathbb{E}_{\vect{z}, \vect{z}_0}\big[\|\vect{z} - \vect{z}_0\|^2\big] = D + C = \Theta(D).
			\end{equation}
			
			\item \textbf{Manifold-Anchored Initialization.} 
			Let $\vect{z}_0 = \mat{V}\vect{u}$, where $\vect{u}$ is sampled via a topology-preserving anchor prior with bandwidth $h$. Assuming the prior ensures local sampling such that $\mathbb{E}[\|\vect{u} - \vect{u}^*\|^2] \le h^2$, the transport distance is independent of the ambient dimension $D$:
			\begin{equation}
				\label{eq:anchor_dist}
				\mathbb{E}_{\vect{z}, \vect{z}_0}\big[\|\vect{z} - \vect{z}_0\|^2\big] \le \sigma_{\max}^2(\mat{V})h^2 + \epsilon^2 = O(h^2 + \epsilon^2).
			\end{equation}
		\end{enumerate}
	\end{theorem}

	\paragraph{Theoretical Implication.}
	Theorem~\ref{thm:transport_distance} highlights the fundamental advantage of our approach. Eq.~\eqref{eq:gauss_dist} shows that standard flow matching must transport probability mass across a squared distance proportional to the ambient dimension $D$ (typically $>3000$ in VQ-VAE spaces), creating an intractable optimization landscape. In contrast, Eq.~\eqref{eq:anchor_dist} demonstrates that SDFlow reduces the transport cost to a term controlled by the intrinsic manifold properties ($h$ and $\epsilon$), effectively bypassing the curse of dimensionality.
	\textbf{This transforms a global transport problem into a local interpolation problem, drastically reducing the difficulty of learning the velocity field.}
	
	\paragraph{Kernel-Smoothed Anchor Prior.}
	The kernel smoother is used only to define a continuous initialization distribution over the learned low-rank coordinates, not to generate time series by itself. Appendix~\ref{sec:appendix_kernel_theory} further shows that this prior can be interpreted as an explicit RBF mixture over latent anchors, and that kernel smoothing enjoys the standard KDE rate $O(N^{-4/(r+4)})$ on the intrinsic coordinate space rather than the much slower raw-space rate $O(N^{-4/(D+4)})$ with $D\gg r$. This explains why local latent support smoothing is appropriate for initialization while the learned flow remains responsible for generation.
	
	\begin{proposition}[Consistency of Categorical Flow Matching]
		\label{prop:consistency}
		Let $\mathcal{C}=\{c_k\}_{k=1}^K\subset\mathbb{R}^d$ be a codebook with bounded norm $\max_k \|c_k\|_2 \le R$.
		Let $p(c\mid z_t)$ be the true categorical posterior over codebook indices and $q_\theta(c\mid z_t)$ be its model prediction.
		
		Let the time-weighted cross-entropy objective be
		\[
		\mathcal{L}_{\mathrm{wCE}}(\theta)
		:=\mathbb{E}_{t\sim\tau,\,z_t}\!\left[\frac{1}{(1-t)^2}
		\mathbb{E}_{c\sim p(\cdot\mid z_t)}[-\log q_\theta(c\mid z_t)]\right],
		\]
		and denote its Bayes-optimal value
		\[
		\mathcal{L}_{\mathrm{wCE}}^*
		:=\mathbb{E}_{t\sim\tau,\,z_t}\!\left[\frac{H(p(\cdot\mid z_t))}{(1-t)^2}\right].
		\]
		Then minimizing this weighted categorical objective controls the mean-squared error of the induced velocity field:
		\begin{equation}
			\label{eq:ce_to_velocity_mse}
			\mathbb{E}_{t\sim\tau,\,z_t}\big[\|v_\theta(z_t,t)-v^*(z_t,t)\|_2^2\big]
			\;\le\;
			2R^2 \cdot \Big(\mathcal{L}_{\mathrm{wCE}}(\theta)-\mathcal{L}_{\mathrm{wCE}}^*\Big).
		\end{equation}
		For the unweighted CE used in practice, the same argument gives a time-weighted KL control; with $t\le 1-\delta$ or a bounded importance weight, this reduces to the usual CE control up to a finite constant.
	\end{proposition}
	
	\paragraph{Explanation.}
	Proposition~\ref{prop:consistency} shows that categorical supervision is not merely a classification surrogate: the time-weighted categorical cross-entropy controls the velocity-field MSE that standard flow matching seeks to minimize. In practice, our cosine time sampling avoids the singular region near $t=1$, making the unweighted CE a stable surrogate. The complete proofs can be found in Appendix~\ref{app:proof_transport_distance} and Appendix~\ref{app:proof_consistency}.

	\section{Experiments}
	\label{sec:experiments}
	\subsection{Experimental Setup}
	We conduct comprehensive experiments to evaluate \method{} on both standard benchmarks and challenging long-sequence generation scenarios.
	We evaluate on four datasets spanning diverse domains and complexities, following the protocol established by prior work~\citep{yoon2019time,chen2024sdformer}. Datasets include: Sines, Stocks, ETTh and Energy. We adopt three complementary metrics following standard practice: Discriminative Score (DS), Predictive Score (PS) and Context-FID. All experiments report mean and standard deviation over 5 random seeds.
	We compare against representative methods from each paradigm:
	\textbf{GAN-based}: TimeGAN~\citep{yoon2019time}, COT-GAN~\citep{xu2020cot};
	\textbf{VAE-based}: TimeVAE~\citep{desai2021timevae};
	\textbf{Diffusion-based}: DiffWave~\citep{kong2021diffwave}, DiffTime~\citep{coletta2024constrained}, Diffusion-TS~\citep{yuan2024diffusion};
	\textbf{Discrete token}: SDformer-m (masked), SDformer-ar (autoregressive)~\citep{chen2024sdformer}, TimeMAR-L~\citep{xu2026timemar}.
	\textbf{Flow-based}: FlowTS~\citep{hu2025flowts}.
	
	We use the same VQ-VAE tokenizer as SDformer for fair comparison, with codebook size $K=512$, latent dimension $d_c=512$, and downsampling rate $s=4$. For \method{}, we use rank $r=256$ with one DiT layer for $L=24$, and rank $r=1024$ with three DiT layers for $L=256$. The low-rank coordinates are normalized before constructing the anchor prior; its geometry-calibrated bandwidth is described in Appendix~\ref{sec:appendix_kernel_theory}. More implementation details are provided in Appendix~\ref{app:setup}.

	\begin{table*}[!t]
		\centering
		\small
		\resizebox{\textwidth}{!}{
			\begin{tabular}{@{}ll cccccccccc@{}}
				\toprule
				\textbf{Metric} & \textbf{Dataset} & \textbf{TimeGAN} & \textbf{COT-GAN} & \textbf{TimeVAE} & \textbf{DiffWave} & \textbf{DiffTime} & \textbf{Diff-TS} & \textbf{FlowTS} & \textbf{SDformer-m} & \textbf{SDformer-ar} & \textbf{Ours} \\
				\midrule
				\multirow{4}{*}{\shortstack{\textbf{DS} $\downarrow$}}
				& \textbf{Sines}  & .011$\pm$.008 & .254$\pm$.137 & .041$\pm$.044 & .017$\pm$.008 & .013$\pm$.006 & \underline{.006$\pm$.007} & \textbf{.005$\pm$.005} & .008$\pm$.004 & \underline{.006$\pm$.004} & \underline{.006$\pm$.006} \\
				& \textbf{Stocks} & .102$\pm$.021 & .230$\pm$.016 & .145$\pm$.120 & .232$\pm$.061 & .097$\pm$.016 & .067$\pm$.015 & .019$\pm$.013 & .020$\pm$.011 & \underline{.010$\pm$.006} & \textbf{.003$\pm$.003} \\
				& \textbf{ETTh}   & .114$\pm$.055 & .325$\pm$.099 & .209$\pm$.058 & .190$\pm$.008 & .100$\pm$.007 & .061$\pm$.009 & .011$\pm$.015 & .022$\pm$.001 & \underline{.003$\pm$.001} & \textbf{.002$\pm$.003} \\
				& \textbf{Energy} & .236$\pm$.012 & .498$\pm$.002 & .499$\pm$.000 & .493$\pm$.004 & .445$\pm$.004 & .122$\pm$.003 & \underline{.053$\pm$.010} & .062$\pm$.006 & \textbf{.006$\pm$.004} & \textbf{.006$\pm$.002} \\
				\midrule
				\multirow{4}{*}{\shortstack{\textbf{PS} $\downarrow$}}
				& \textbf{Sines}  & \underline{.093$\pm$.019} & .100$\pm$.000 & \underline{.093$\pm$.000} & \underline{.093$\pm$.000} & \underline{.093$\pm$.000} & \underline{.093$\pm$.000} & \textbf{.092$\pm$.000} & \underline{.093$\pm$.000} & \underline{.093$\pm$.000} & \textbf{.092$\pm$.000} \\
				& \textbf{Stocks} & .038$\pm$.001 & .047$\pm$.001 & .039$\pm$.000 & .047$\pm$.000 & .038$\pm$.001 & \textbf{.036$\pm$.000} & \textbf{.036$\pm$.000} & \underline{.037$\pm$.000} & \underline{.037$\pm$.000} & \textbf{.036$\pm$.000} \\
				& \textbf{ETTh}   & .124$\pm$.001 & .129$\pm$.000 & .126$\pm$.004 & .130$\pm$.001 & .121$\pm$.004 & \underline{.119$\pm$.002} & \textbf{.118$\pm$.005} & \underline{.119$\pm$.002} & \textbf{.118$\pm$.002} & \textbf{.118$\pm$.001} \\
				& \textbf{Energy} & .273$\pm$.004 & .259$\pm$.000 & .292$\pm$.000 & .251$\pm$.000 & .252$\pm$.000 & \underline{.250$\pm$.000} & \underline{.250$\pm$.000} & \underline{.250$\pm$.000} & \textbf{.249$\pm$.000} & \textbf{.249$\pm$.000} \\
				\midrule
				\multirow{4}{*}{\shortstack{\textbf{C-FID} $\downarrow$}}
				& \textbf{Sines}  & .101$\pm$.014 & 1.337$\pm$.068& .307$\pm$.060 & .014$\pm$.002 & .006$\pm$.001 & .006$\pm$.000 & .002$\pm$.000 & .010$\pm$.002 & \underline{.001$\pm$.000} & \textbf{.0004$\pm$.000} \\
				& \textbf{Stocks} & .103$\pm$.013 & .408$\pm$.086 & .215$\pm$.035 & .232$\pm$.032 & .236$\pm$.074 & .147$\pm$.025 & .015$\pm$.003 & .034$\pm$.008 & \underline{.002$\pm$.000} & \textbf{.001$\pm$.000} \\
				& \textbf{ETTh}   & .300$\pm$.013 & .980$\pm$.071 & .805$\pm$.186 & .873$\pm$.061 & .299$\pm$.044 & .116$\pm$.010 & .024$\pm$.001 & .019$\pm$.003 & \underline{.008$\pm$.001} & \textbf{.001$\pm$.000} \\
				& \textbf{Energy} & .767$\pm$.103 & 1.039$\pm$.028& 1.631$\pm$.142& 1.031$\pm$.131& .279$\pm$.045 & .089$\pm$.024 & .031$\pm$.004 & .041$\pm$.005 & \underline{.003$\pm$.000} & \textbf{.001$\pm$.000} \\
				\bottomrule
			\end{tabular}
		}
		\caption{Unconditional generation results on standard benchmarks ($L=24$). Best results in \textbf{bold}, second best \underline{underlined}. Lower is better for all metrics. Results show mean $\pm$ std over 5 runs.}
		\label{tab:main_short}
	\end{table*}

	\subsection{Main Results}
	\label{sec:main_results}
	
	\subsubsection{Short-Sequence Generation (\small$L=24$).} 
	Table~\ref{tab:main_short} presents results on the standard short-sequence benchmark. \method{} achieves competitive or superior performance across all datasets. Notably, on Energy, \method{} matches the state-of-the-art DS (0.006) while achieving dramatically superior Context-FID (0.001 vs. 0.003), indicating better distributional coverage beyond fooling discriminators. The t-SNE visualizations (Figure~\ref{fig:t_sne}) confirm that generated samples align with real data, filling the manifold without complete overlap.

	\subsubsection{Long-Sequence Generation (\small $L: 64, 128 , 256$).}

	\begin{table*}[t]
		\centering
		\small
		\resizebox{\textwidth}{!}{
			\begin{tabular}{@{}l ccc ccc ccc@{}}
				\toprule
				& \multicolumn{3}{c}{\textbf{Discriminative Score $\downarrow$}} & \multicolumn{3}{c}{\textbf{Predictive Score $\downarrow$}} & \multicolumn{3}{c}{\textbf{Context-FID $\downarrow$}} \\
				\cmidrule(lr){2-4} \cmidrule(lr){5-7} \cmidrule(lr){8-10}
				\textbf{Method} & $L=64$ & $L=128$ & $L=256$ & $L=64$ & $L=128$ & $L=256$ & $L=64$ & $L=128$ & $L=256$ \\
				\midrule
				\multicolumn{10}{c}{\textit{ETTh Dataset}} \\
				\midrule
				TimeGAN & 0.227 $\pm$ .078 & 0.188 $\pm$ .074 & 0.442 $\pm$ .056 & 0.132 $\pm$ .008 & 0.153 $\pm$ .014 & 0.220 $\pm$ .008 & 1.130 $\pm$ .102 & 1.553 $\pm$ .169 & 5.872 $\pm$ .208 \\
				COT-GAN & 0.296 $\pm$ .348 & 0.451 $\pm$ .080 & 0.461 $\pm$ .010 & 0.135 $\pm$ .003 & 0.126 $\pm$ .001 & 0.129 $\pm$ .000 & 3.008 $\pm$ .277 & 2.639 $\pm$ .427 & 4.075 $\pm$ .894 \\
				TimeVAE & 0.171 $\pm$ .142 & 0.154 $\pm$ .087 & 0.178 $\pm$ .076 & 0.118 $\pm$ .004 & 0.113 $\pm$ .005 & 0.110 $\pm$ .027 & 0.827 $\pm$ .146 & 1.062 $\pm$ .134 & 0.826 $\pm$ .093 \\
				DiffWave & 0.254 $\pm$ .074 & 0.274 $\pm$ .047 & 0.304 $\pm$ .068 & 0.133 $\pm$ .008 & 0.129 $\pm$ .003 & 0.132 $\pm$ .001 & 1.543 $\pm$ .153 & 2.354 $\pm$ .170 & 2.899 $\pm$ .289 \\
				DiffTime & 0.150 $\pm$ .003 & 0.176 $\pm$ .015 & 0.243 $\pm$ .005 & 0.118 $\pm$ .004 & 0.120 $\pm$ .008 & 0.118 $\pm$ .003 & 1.279 $\pm$ .083 & 2.554 $\pm$ .318 & 3.524 $\pm$ .830 \\
				Diffusion-TS & 0.106 $\pm$ .048 & 0.144 $\pm$ .060 & 0.060 $\pm$ .030 & 0.116 $\pm$ .000 & 0.110 $\pm$ .003 & 0.109 $\pm$ .013 & 0.631 $\pm$ .058 & 0.787 $\pm$ .062 & 0.423 $\pm$ .038 \\
				FlowTS & 0.010 $\pm$ .004 & 0.040 $\pm$ .012 & 0.081 $\pm$ .022 & 0.115 $\pm$ .005 & \textbf{0.104 $\pm$ .013} & 0.107 $\pm$ .005 & 0.039 $\pm$ .003 & 0.128 $\pm$ .007 & 0.302 $\pm$ .018 \\
				SDformer-ar & 0.018 $\pm$ .007 & 0.013 $\pm$ .005 & 0.008 $\pm$ .006 & 0.116 $\pm$ .006 & 0.110 $\pm$ .007 & \textbf{0.095 $\pm$ .003} & 0.018 $\pm$ .003 & 0.024 $\pm$ .001 & 0.021 $\pm$ .001 \\
				TimeMAR-L & \underline{0.005 $\pm$ .004} & \underline{0.008 $\pm$ .001} & \underline{0.006 $\pm$ .006} & \underline{0.114 $\pm$ .004} & 0.110 $\pm$ .007 & \underline{0.105 $\pm$ .007} & \underline{0.002 $\pm$ .002} & \underline{0.002 $\pm$ .001} & \underline{0.002 $\pm$ .001} \\
				\textbf{\method{} (Ours)} & \textbf{0.003 $\pm$ .000} & \textbf{0.007 $\pm$ .003} & \textbf{0.005 $\pm$ .002} & \textbf{0.113 $\pm$ .003} & \underline{0.109 $\pm$ .006} & \underline{0.105 $\pm$ .004} & \textbf{0.001 $\pm$ .000} & \textbf{0.001 $\pm$ .000} & \textbf{0.001 $\pm$ .000} \\
				\textit{Original} & --- & --- & --- & \textit{0.114 $\pm$ .006} & \textit{0.108 $\pm$ .005} & \textit{0.106 $\pm$ .010} & --- & --- & --- \\
				\midrule
				\multicolumn{10}{c}{\textit{Energy Dataset}} \\
				\midrule
				TimeGAN & 0.498 $\pm$ .001 & 0.499 $\pm$ .001 & 0.499 $\pm$ .000 & 0.291 $\pm$ .003 & 0.303 $\pm$ .002 & 0.351 $\pm$ .004 & 1.230 $\pm$ .070 & 2.535 $\pm$ .372 & 5.032 $\pm$ .831 \\
				COT-GAN & 0.499 $\pm$ .001 & 0.499 $\pm$ .001 & 0.498 $\pm$ .004 & 0.262 $\pm$ .002 & 0.269 $\pm$ .002 & 0.275 $\pm$ .004 & 1.824 $\pm$ .144 & 1.822 $\pm$ .271 & 2.533 $\pm$ .467 \\
				TimeVAE & 0.499 $\pm$ .000 & 0.499 $\pm$ .000 & 0.499 $\pm$ .000 & 0.302 $\pm$ .001 & 0.318 $\pm$ .000 & 0.353 $\pm$ .003 & 2.662 $\pm$ .087 & 3.125 $\pm$ .106 & 3.768 $\pm$ .998 \\
				DiffWave & 0.497 $\pm$ .004 & 0.499 $\pm$ .001 & 0.499 $\pm$ .000 & 0.252 $\pm$ .001 & 0.252 $\pm$ .000 & 0.251 $\pm$ .000 & 2.697 $\pm$ .418 & 5.552 $\pm$ .528 & 5.572 $\pm$ .584 \\
				DiffTime & 0.328 $\pm$ .031 & 0.396 $\pm$ .024 & 0.437 $\pm$ .095 & 0.252 $\pm$ .000 & 0.251 $\pm$ .000 & 0.251 $\pm$ .000 & 0.762 $\pm$ .157 & 1.344 $\pm$ .131 & 4.735 $\pm$ .729 \\
				Diffusion-TS & 0.078 $\pm$ .021 & 0.143 $\pm$ .075 & 0.290 $\pm$ .123 & \underline{0.249 $\pm$ .000} & \underline{0.247 $\pm$ .001} & 0.245 $\pm$ .001 & 0.135 $\pm$ .017 & 0.087 $\pm$ .019 & 0.126 $\pm$ .024 \\
				SDformer-ar & 0.010 $\pm$ .007 & \underline{0.013 $\pm$ .007} & 0.017 $\pm$ .003 & \textbf{0.247 $\pm$ .001} & \textbf{0.244 $\pm$ .000} & \underline{0.243 $\pm$ .002} & \underline{0.031 $\pm$ .002} & 0.036 $\pm$ .002 & 0.041 $\pm$ .003 \\
				TimeMAR-L & \underline{0.006 $\pm$ .004} & \textbf{0.005 $\pm$ .005} & \underline{0.009 $\pm$ .005} & \textbf{0.247 $\pm$ .001} & \textbf{0.244 $\pm$ .001} & \underline{0.243 $\pm$ .003} & \textbf{0.002 $\pm$ .000} & \underline{0.003 $\pm$ .000} & \underline{0.008 $\pm$ .001} \\
				\textbf{\method{} (Ours)} & \textbf{0.004 $\pm$ .001} & \textbf{0.005 $\pm$ .004} & \textbf{0.007 $\pm$ .006} & \textbf{0.247 $\pm$ .000} & \textbf{0.244 $\pm$ .001} & \textbf{0.242 $\pm$ .002} & \textbf{0.002 $\pm$ .000} & \textbf{0.002 $\pm$ .000} & \textbf{0.002 $\pm$ .000} \\
				\textit{Original} & --- & --- & --- & \textit{0.245 $\pm$ .002} & \textit{0.243 $\pm$ .000} & \textit{0.243 $\pm$ .000} & --- & --- & --- \\
				\bottomrule
			\end{tabular}
		}
		\caption{Unconditional long-sequence generation results ($L=64, 128, 256$). Best results in \textbf{bold}, second best \underline{underlined}. Lower is better for all metrics. Results show mean $\pm$ std over 5 runs.}
		\label{tab:main_long}
	\end{table*}
	
	Table~\ref{tab:main_long} presents results on the primary evaluation setting where \method{} demonstrates its fundamental advantages over AR methods. As length increases, SDformer-ar exhibits systematic degradation due to exposure bias and error accumulation. At $L=256$, SDformer-ar achieves DS of 0.008 on ETTh and 0.017 on Energy---significant degradation from its strong short-sequence performance. Even when TimeMAR-L ($ \ge 40M $) has a parameter quantity that is an order of magnitude larger than ours, its performance is still inferior to ours in the vast majority of cases.

	In contrast, \method{} maintains stable quality across all lengths through parallel generation that eliminates cascading errors. At $L=256$: DS of 0.005 on ETTh (37.5\% improvement over SDformer-ar, 16.6\% improvement over TimeMAR-L). The Context-FID improvements are even more dramatic: 0.001 vs. 0.021 on ETTh (95.2\% reduction) and 0.002 vs. 0.041 on Energy (95.1\% reduction). These results indicate that exposure bias is a major bottleneck for AR token modeling in long-horizon generation. Non-AR generation via \method{} provides a principled alternative.

	\subsubsection{Spectral Analysis}
	
	\begin{figure}[h]
		\centering
		\begin{subfigure}[b]{0.24\columnwidth}
			\centering
			\includegraphics[width=\linewidth]{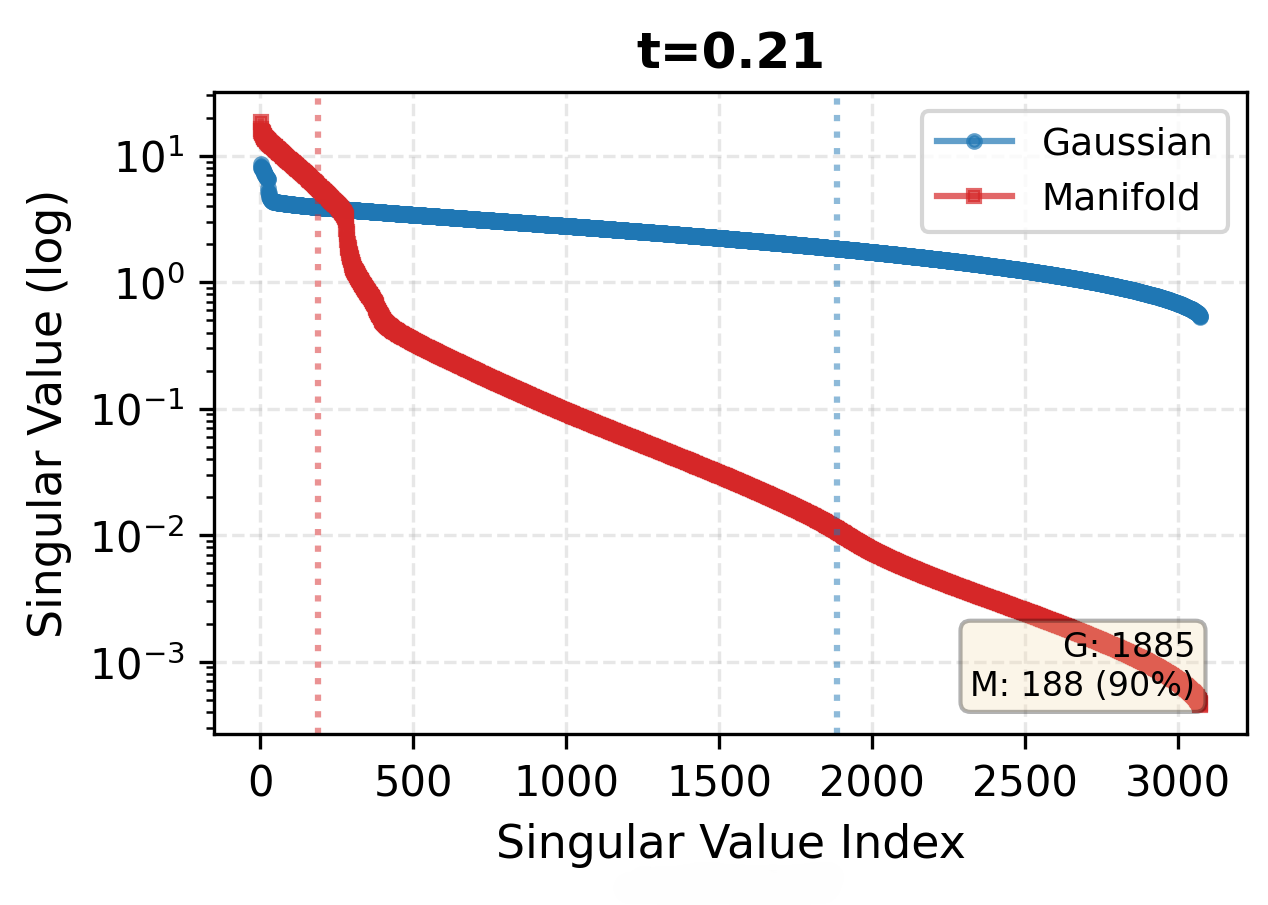}
			\label{subfig:svd_spectrum_1}
		\end{subfigure}
		\hfill
		\begin{subfigure}[b]{0.24\columnwidth}
			\centering
			\includegraphics[width=\linewidth]{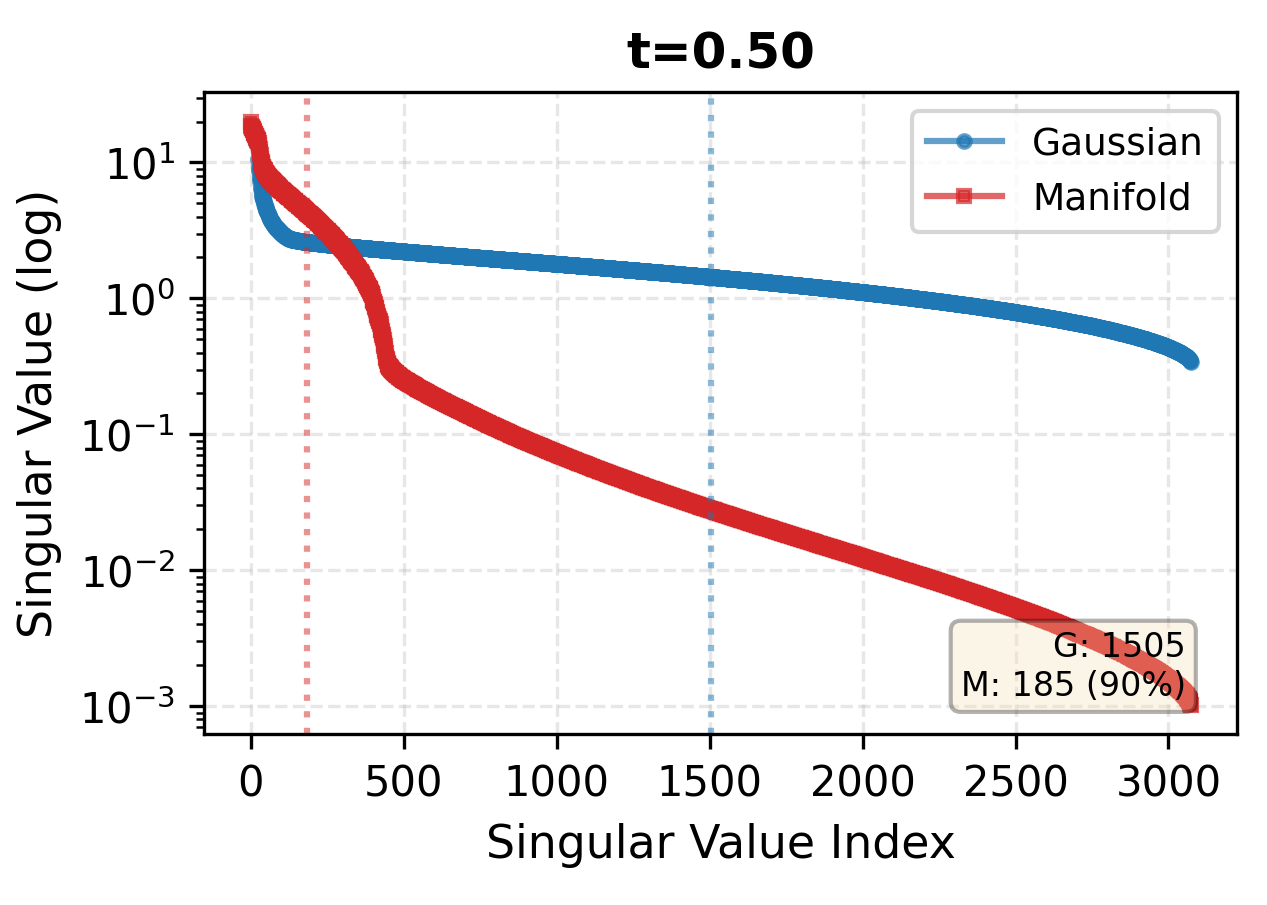}
			\label{subfig:svd_spectrum_2}
		\end{subfigure}
		\hfill
		\begin{subfigure}[b]{0.24\columnwidth}
			\centering
			\includegraphics[width=\linewidth]{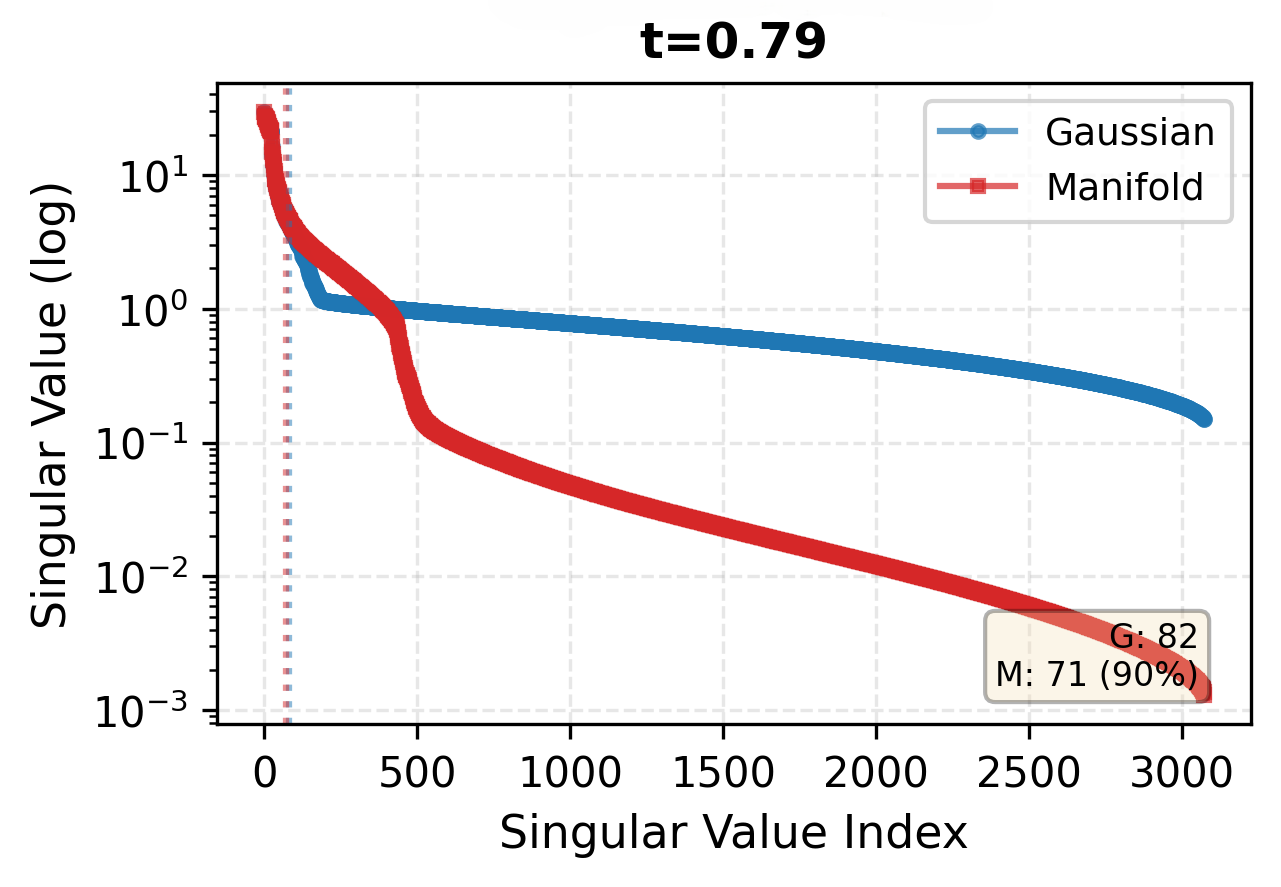}
			\label{subfig:svd_spectrum_3}
		\end{subfigure}
		\hfill
		\begin{subfigure}[b]{0.24\columnwidth}
			\centering
			\includegraphics[width=\linewidth]{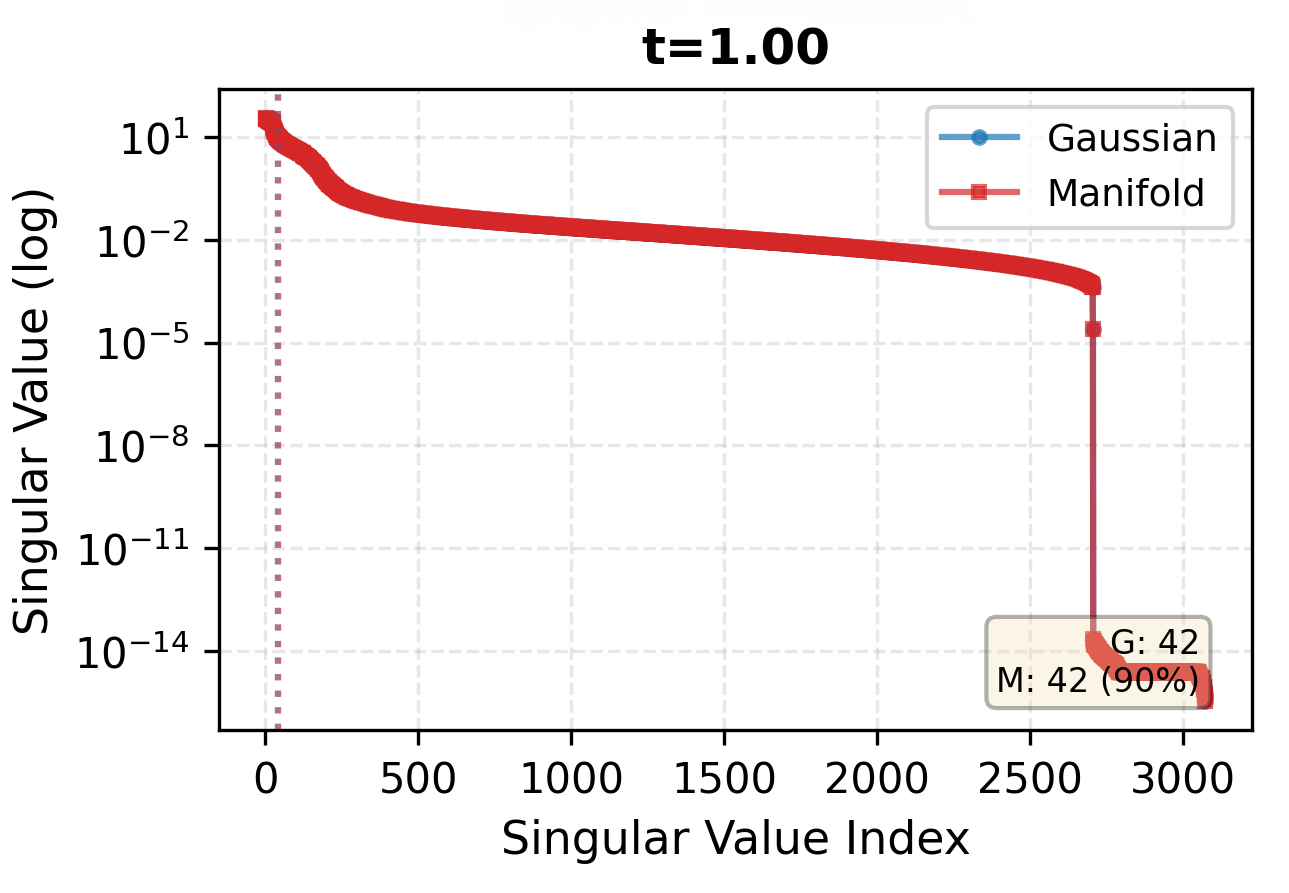}
			\label{subfig:svd_spectrum_4}
		\end{subfigure}
		\vspace{-1.5em}
		\caption{Singular Value Spectrum Analysis (Energy, $D=3072$).}
		\label{fig:svd_spectrum}
	\end{figure}
	To verify whether our model learns the intrinsic data structure, we analyze the singular value spectrum of the feature representations.
	As shown in Figure~\ref{fig:svd_spectrum}, we observe a significant difference between the learned Manifold (red) and the Gaussian noise (blue). When $t$ is close to zero, the singular values of the Manifold drop sharply, with the top 200 components explaining more than 90\% of the variance. This steep spectral decay indicates that our method effectively recovers the low-dimensional structure of the data. Conversely, the Gaussian baseline exhibits a flat spectrum, characteristic of unstructured high-dimensional noise. This suggests that the flow trajectories transport the noise distribution onto a concentrated data manifold.

	\subsubsection{Inference Efficiency}
	\vspace{-0.1cm}
	We compare the inference time for generating 1,024 samples ($L=24$). \method{} requires only \textbf{0.04--0.05s} across all datasets (e.g., 0.05s on Sines, 0.04s on Energy), drastically outperforming SDformer-ar (0.15--0.40s) and Diffusion-TS (9.57--34.92s). This translates to a \textbf{3.0$\times$ to 10.0$\times$ speedup} over the AR baseline. The gain comes from replacing $L$ sequential AR decoding passes with $S$ parallel refinement passes ($S=20$), while each pass still uses the same transformer attention pattern; therefore, we report wall-clock speedups to contextualize long-horizon efficiency.
	
	\subsection{Ablation, Hyperparameter and Generalization Study}
	\label{sec:ablation}
	\vspace{-0.1cm}

	\begin{table}[t]
		\centering
		\tiny
		\setlength{\tabcolsep}{1.7pt}
		\renewcommand{\arraystretch}{0.88}
		\resizebox{0.9\linewidth}{!}{
			\begin{tabular}{@{}lc@{\hspace{0.8em}}lc@{\hspace{0.8em}}lccc@{}}
				\toprule
				\multicolumn{2}{c}{\textbf{Component Ablation}} & \multicolumn{2}{c}{\textbf{Sensitivity}} & \multicolumn{4}{c}{\textbf{Held-out Latent-Flow Generalization}} \\
				\cmidrule(lr){1-2} \cmidrule(lr){3-4} \cmidrule(lr){5-8}
				\textbf{Variant} & \textbf{DS} $\downarrow$ & \textbf{Setting} & \textbf{DS} $\downarrow$ & \textbf{Train} & \textbf{Eval} & \textbf{DS} $\downarrow$ & \textbf{C-FID} $\downarrow$ \\
				\midrule
				{Raw data space} & & {Rank ($r$)} & & \multicolumn{4}{l}{\textit{Energy}} \\
				\hspace{0.4em}\textit{+ Flow in data space} & 0.453
				& \hspace{0.4em}\textit{+ Static SVD ($r=42$), $h=0.06$} & 0.094 & 100\% latents & Full & 0.006 & 0.001 \\
				{VQ latent space} &
				& \hspace{0.4em}\textit{+ Learned $r=128$, $h=0.06$} & 0.015 & \textit{20\% latents} & Held-out 80\% & 0.007 & 0.001 \\
				\hspace{0.4em}\textit{+ Gaussian prior} & 0.218
				& \hspace{0.4em}\textbf{\textit{+ Learned $r=256$, $h=0.06$}} & \textbf{0.006} & \textbf{\textit{10\% latents}} & \textbf{Held-out 90\%} & \textbf{0.009} & \textbf{0.002} \\
				\hspace{0.4em}\textit{+ Anchor prior w/o low-rank} & 0.173
				& \hspace{0.4em}\textit{+ Learned $r=512$, $h=0.06$} & 0.006 & $\Delta$ (10\% vs. full) & -- & .003 & .001 \\
				\cmidrule(lr){1-2} \cmidrule(lr){3-4} \cmidrule(lr){5-8}
				{Low-rank latent scaffold} & & {Bandwidth ($h$)} & & \multicolumn{4}{l}{\textit{Sines}} \\
				\hspace{0.4em}\textit{+ Gaussian prior} & 0.218
				& \hspace{0.4em}\textit{+ $h=0.60$, $r=256$} & 0.138 & 100\% latents & Full & 0.006 & 0.0004 \\
				\hspace{0.4em}\textit{+ Learned MLP prior} & 0.213
				& \hspace{0.4em}\textit{+ $h=0.12$, $r=256$} & 0.049 & \textit{20\% latents} & Held-out 80\% & 0.006 & 0.0009 \\
				\hspace{0.4em}\textit{+ Interpolation prior} & 0.047
				& \hspace{0.4em}\textbf{\textit{+ $h=0.06$, $r=256$}} & \textbf{0.006} & \textbf{\textit{10\% latents}} & \textbf{Held-out 90\%} & \textbf{0.007} & \textbf{0.0013} \\
				\hspace{0.4em}\textbf{\textit{+ Learned anchor prior (Ours)}} & \textbf{0.006}
				& \hspace{0.4em}\textit{+ $h=0.01$, $r=256$} & 0.008 & $\Delta$ (10\% vs. full) & -- & .001 & .0009 \\
				\cmidrule(lr){1-2} \cmidrule(lr){3-4} \cmidrule(lr){5-8}
				{Capacity (w/o low-rank)} & & {Solver steps} & & \multicolumn{4}{l}{\textit{ETTh}} \\
				\hspace{0.4em}\textit{+ Smaller codebook, $d=256$} & 0.223
				& \hspace{0.4em}\textit{+ 10 ODE steps} & 0.018 & 100\% latents & Full & 0.002 & 0.001 \\
				\hspace{0.4em}\textit{+ Smaller codebook, $d=128$} & 0.251
				& \hspace{0.4em}\textbf{\textit{+ 20 ODE steps (Ours)}} & \textbf{0.006} & \textbf{\textit{50\% latents}} & \textbf{Held-out 50\%} & \textbf{0.003} & \textbf{0.002} \\
				\hspace{0.4em}\textit{+ Smaller codebook, $d=64$} & 0.299
				& \hspace{0.4em}\textit{+ 50 ODE steps} & 0.006 & $\Delta$ (50\% vs. full) & -- & .001 & .001 \\
				\bottomrule
			\end{tabular}
		}
		\vspace{0.1cm}
		\caption{Ablation, sensitivity, and held-out latent-flow generalization. The left panel isolates key design choices, the middle panel studies rank, bandwidth, and solver sensitivity, and the right panel keeps the VQ tokenizer as a shared representation layer while training the Stage-2 latent flow and anchors on small encoded subsets.}
		\label{tab:ablation}
	\end{table}
	
	\textbf{Representation Space.}
	Table~\ref{tab:ablation} shows that applying flow in data space performs poorly (DS 0.453), whereas SDFlow operates after VQ tokenization. This supports our central design: generation is more effective over the frozen VQ latent space, producing meaningful synthetic samples for downstream tasks. The Predictive Score (TSTR, Train on Synthetic, Test on Real) further supports this point. 
	
	\textbf{Low-Rank Scaffold.} VQ tokenization alone is insufficient: Gaussian initialization (0.218) and anchor initialization without low-rank structure (0.173) remain weak. \textbf{Anchor Prior.} On top of the low-rank scaffold, generic Gaussian or amortized MLP priors still fail (0.218/0.213), while the learned anchor prior reaches 0.006, indicating a learned latent prototype sampler.
	\textbf{Codebook Capacity.} Simply reducing the codebook dimension does not help (0.223 for $d=256$, 0.251 for $d=128$). \textbf{ODE Solver.} $S=20$ matches $S=50$ and is much better than $S=10$, giving an efficient solver budget. 
	
	\textbf{Held-out Generalization.} With the VQ tokenizer kept as the shared representation layer, training the Stage-2 latent flow and anchors on small encoded subsets (e.g., 10\%) still yields close DS/C-FID on held-out real distributions. This supports that the low-rank latent manifold captures shared distributional geometry rather than relying on point-wise access to every training trajectory.
	
	More details are provided in Appendix~\ref{app:ablation_details}.

	\section{Conclusion}
	\label{sec:conclusion}
	\vspace{-0.1cm}
	We have presented \method{}, a principled framework for non-autoregressive time series generation that mitigates exposure bias in autoregressive token modeling. By combining low-rank manifold decomposition with categorical flow matching, \method{} transforms long-range transport into tractable local refinement from manifold anchors, offering both high fidelity and computational efficiency. Comprehensive experiments demonstrate state-of-the-art performance across all benchmarks, with strong improvements on long sequences: up to 95\% Context-FID reduction and 3--10$\times$ faster inference.

	\vspace{-0.2em}
	\textbf{Limitations.} Currently, \method{} uses a dataset-specific VQ tokenizer and codebook. A natural next step is to develop a universal time-series tokenizer with a shared codebook, enabling \method{} to operate in a unified latent space for multi-domain generation.

	%
	%

	\begin{ack}
%
%

		This work was supported by Brain Science and Brain-like Intelligence Technology---National Science and Technology Major Project under Grant 2022ZD0208700, and National Natural Science Foundation of China under Grant 62376264.
		
	\end{ack}
	

	%
	
	{
		\small
		\bibliographystyle{plainnat}
		\bibliography{references}
		%
		%
		%
		%
	}

	
	\newpage
	\begin{center}
		{\LARGE \bf Appendices for SDFlow}\\[4pt]
	\end{center}
	\appendix

	
	%
	
	
	\section{Proof of Theorem~\ref{thm:transport_distance}}
	\label{app:proof_transport_distance}
	
	We prove the two claims in Theorem~\ref{thm:transport_distance}.
	
	\subsection*{Part (i): Gaussian Initialization}
	
	Let $\vect{z}\sim\mathcal{D}$ and $\vect{z}_0\sim\mathcal{N}(\vect{0},\mat{I}_D)$ be independent.
	Using the polarization identity,
	\begin{equation}
		\label{eq:expand_gauss}
		\|\vect{z}-\vect{z}_0\|^2
		= \|\vect{z}\|^2 + \|\vect{z}_0\|^2 - 2\langle \vect{z},\vect{z}_0\rangle.
	\end{equation}
	Taking expectation over $(\vect{z},\vect{z}_0)$ and using independence gives
	\begin{equation}
		\label{eq:exp_gauss_step1}
		\mathbb{E}\|\vect{z}-\vect{z}_0\|^2
		= \mathbb{E}\|\vect{z}\|^2 + \mathbb{E}\|\vect{z}_0\|^2 - 2\,\mathbb{E}\langle \vect{z},\vect{z}_0\rangle.
	\end{equation}
	The cross term vanishes since $\vect{z}_0$ is zero-mean and independent of $\vect{z}$:
	\begin{equation}
		\label{eq:cross_zero}
		\mathbb{E}\langle \vect{z},\vect{z}_0\rangle
		= \mathbb{E}_{\vect{z}}\Big[\big\langle \vect{z},\,\mathbb{E}_{\vect{z}_0}[\vect{z}_0]\big\rangle\Big]
		= \mathbb{E}_{\vect{z}}\big[\langle \vect{z},\vect{0}\rangle\big] = 0.
	\end{equation}
	Moreover, for $\vect{z}_0\sim\mathcal{N}(\vect{0},\mat{I}_D)$, we have
	\begin{equation}
		\label{eq:gauss_norm}
		\mathbb{E}\|\vect{z}_0\|^2
		= \mathbb{E}\Big[\sum_{i=1}^D z_{0,i}^2\Big]
		= \sum_{i=1}^D \mathbb{E}[z_{0,i}^2]
		= \sum_{i=1}^D \mathrm{Var}(z_{0,i})
		= D.
	\end{equation}
	By the boundedness assumption $\mathbb{E}\|\vect{z}\|^2=C<\infty$, substituting \eqref{eq:cross_zero} and \eqref{eq:gauss_norm} into \eqref{eq:exp_gauss_step1} yields
	\[
	\mathbb{E}_{\vect{z},\vect{z}_0}\big[\|\vect{z}-\vect{z}_0\|^2\big]
	= C + D,
	\]
	which proves \eqref{eq:gauss_dist}. Since $C$ is independent of $D$, we have $C+D=\Theta(D)$.
	
	\subsection*{Part (ii): Manifold-Anchored Initialization}
	
	Assume $\mat{V}\in\mathbb{R}^{D\times r}$ is semi-orthogonal and spans $\mathcal{M}$, and each sample admits the decomposition
	\begin{equation}
		\label{eq:decomp}
		\vect{z}=\mat{V}\vect{u}^*+\vect{\epsilon},
		\qquad \|\vect{\epsilon}\|^2 \le \epsilon^2,
	\end{equation}
	where $\vect{u}^*\in\mathbb{R}^r$ is the intrinsic coordinate.
	Let the anchored initialization be $\vect{z}_0=\mat{V}\vect{u}$, where $\vect{u}\sim p_{\mathrm{anchor}}$ is sampled from the anchor prior and satisfies
	\begin{equation}
		\label{eq:kde_h}
		\mathbb{E}\|\vect{u}-\vect{u}^*\|^2 \le h^2.
	\end{equation}
	Define $\delta:=\vect{u}-\vect{u}^*$ so that $\vect{u}=\vect{u}^*+\delta$.
	Then, using \eqref{eq:decomp},
	\begin{equation}
		\label{eq:anchor_diff}
		\vect{z}-\vect{z}_0
		= (\mat{V}\vect{u}^*+\vect{\epsilon}) - \mat{V}\vect{u}
		= \mat{V}(\vect{u}^*-\vect{u}) + \vect{\epsilon}
		= -\mat{V}\delta + \vect{\epsilon}.
	\end{equation}
	Hence,
	\begin{equation}
		\label{eq:norm_expand}
		\|\vect{z}-\vect{z}_0\|^2
		= \|-\mat{V}\delta + \vect{\epsilon}\|^2
		= \|\mat{V}\delta\|^2 + \|\vect{\epsilon}\|^2 - 2\langle \mat{V}\delta, \vect{\epsilon}\rangle.
	\end{equation}
	
	\paragraph{Cross term control.}
	Since $\mat{V}$ is semi-orthogonal, $\mathrm{col}(\mat{V})=\mathcal{M}$.
	Interpreting $\vect{\epsilon}$ as the approximation residual orthogonal to $\mathcal{M}$ (i.e., $\vect{\epsilon}\in\mathcal{M}^\perp$), we have
	\begin{equation}
		\label{eq:orth}
		\langle \mat{V}\delta, \vect{\epsilon}\rangle = 0,
	\end{equation}
	and therefore
	\begin{equation}
		\label{eq:pythag}
		\|\vect{z}-\vect{z}_0\|^2
		= \|\mat{V}\delta\|^2 + \|\vect{\epsilon}\|^2.
	\end{equation}
	If one does not assume $\vect{\epsilon}\in\mathcal{M}^\perp$, a weaker bound still holds by $\|a+b\|^2\le 2\|a\|^2+2\|b\|^2$, but we proceed with \eqref{eq:orth} to match the statement of Theorem~\ref{thm:transport_distance}.
	
	\paragraph{Bounding the two terms.}
	First, by the operator norm inequality,
	\begin{equation}
		\label{eq:op}
		\|\mat{V}\delta\|
		\le \|\mat{V}\|_2\|\delta\|
		= \sigma_{\max}(\mat{V})\|\delta\|,
		\qquad\Rightarrow\qquad
		\|\mat{V}\delta\|^2 \le \sigma_{\max}^2(\mat{V})\|\delta\|^2.
	\end{equation}
	Taking expectation and using \eqref{eq:kde_h},
	\begin{equation}
		\label{eq:first_term}
		\mathbb{E}\|\mat{V}\delta\|^2
		\le \sigma_{\max}^2(\mat{V})\,\mathbb{E}\|\delta\|^2
		\le \sigma_{\max}^2(\mat{V})\,h^2.
	\end{equation}
	Second, by the bounded approximation error in \eqref{eq:decomp},
	\begin{equation}
		\label{eq:second_term}
		\mathbb{E}\|\vect{\epsilon}\|^2 \le \epsilon^2.
	\end{equation}
	
	\paragraph{Putting everything together.}
	Taking expectation in \eqref{eq:pythag} and applying \eqref{eq:first_term}--\eqref{eq:second_term} gives
	\[
	\mathbb{E}\|\vect{z}-\vect{z}_0\|^2
	=
	\mathbb{E}\|\mat{V}\delta\|^2 + \mathbb{E}\|\vect{\epsilon}\|^2
	\le
	\sigma_{\max}^2(\mat{V})\,h^2 + \epsilon^2,
	\]
	which is exactly \eqref{eq:anchor_dist}.
	Finally, since $\sigma_{\max}(\mat{V})$ does not depend on $D$ when $\mat{V}$ is semi-orthogonal (indeed $\sigma_{\max}(\mat{V})=1$), we obtain
	\[
	\mathbb{E}\|\vect{z}-\vect{z}_0\|^2 = O(h^2+\epsilon^2),
	\]
	establishing the dimension-free scaling claimed in Theorem~\ref{thm:transport_distance}.
	\qed
	
	While we assumes a semi-orthogonal basis V for theoretical tractability, our empirical results indicate that standard unconstrained optimization of V is practically sufficient to achieve the same geometric benefits.
	
	\section{Proof of Proposition~\ref{prop:consistency}}
	\label{app:proof_consistency}
	
	We prove the velocity-field mean-squared-error (MSE) bound induced by minimizing the cross-entropy objective.
	
	\paragraph{Step 1: Reduce velocity-field error to posterior-mean embedding error.}
	By definition,
	\[
	v_\theta(z_t,t)-v^*(z_t,t)
	=
	\frac{\mu_\theta(z_t)-z_t}{1-t}-\frac{\mu^*(z_t)-z_t}{1-t}
	=
	\frac{\mu_\theta(z_t)-\mu^*(z_t)}{1-t}.
	\]
	Hence,
	\begin{equation}
		\label{eq:v_reduce_mu}
		\|v_\theta(z_t,t)-v^*(z_t,t)\|_2^2
		=
		\frac{\|\mu_\theta(z_t)-\mu^*(z_t)\|_2^2}{(1-t)^2}.
	\end{equation}
	
	\paragraph{Step 2: Bound $\|\mu_\theta-\mu^*\|$ by the $\ell_1$ distance between categorical posteriors.}
	Recall
	\[
	\mu_\theta(z_t)-\mu^*(z_t)
	=
	\sum_{k=1}^K \big(q_\theta(c_k\mid z_t)-p(c_k\mid z_t)\big)\,c_k.
	\]
	Using the triangle inequality and the codebook radius bound $\|c_k\|_2\le R$,
	\begin{align}
		\|\mu_\theta(z_t)-\mu^*(z_t)\|_2
		&\le
		\sum_{k=1}^K \big|q_\theta(c_k\mid z_t)-p(c_k\mid z_t)\big|\,\|c_k\|_2 \nonumber\\
		&\le
		R \sum_{k=1}^K \big|q_\theta(c_k\mid z_t)-p(c_k\mid z_t)\big|
		=
		R\,\|q_\theta(\cdot\mid z_t)-p(\cdot\mid z_t)\|_1.
		\label{eq:mu_l1}
	\end{align}
	Squaring both sides yields
	\begin{equation}
		\label{eq:mu_l1_sq}
		\|\mu_\theta(z_t)-\mu^*(z_t)\|_2^2
		\le
		R^2\,\|q_\theta(\cdot\mid z_t)-p(\cdot\mid z_t)\|_1^2.
	\end{equation}
	
	\paragraph{Step 3: Convert $\ell_1$ distance to KL via Pinsker's inequality.}
	Pinsker's inequality states that for any distributions $p,q$ on a finite set,
	\begin{equation}
		\label{eq:pinsker}
		\|p-q\|_1^2 \le 2\,\mathrm{KL}(p\|q).
	\end{equation}
	Applying \eqref{eq:pinsker} with $p(\cdot\mid z_t)$ and $q_\theta(\cdot\mid z_t)$ and combining with \eqref{eq:mu_l1_sq} gives
	\begin{equation}
		\label{eq:mu_kl}
		\|\mu_\theta(z_t)-\mu^*(z_t)\|_2^2
		\le
		2R^2\,\mathrm{KL}\!\left(p(\cdot\mid z_t)\,\|\,q_\theta(\cdot\mid z_t)\right).
	\end{equation}
	
	\paragraph{Step 4: Take expectations with the correct time weight.}
	Substituting \eqref{eq:mu_kl} into \eqref{eq:v_reduce_mu} yields the pointwise bound
	\[
	\|v_\theta(z_t,t)-v^*(z_t,t)\|_2^2
	\le
	\frac{2R^2}{(1-t)^2}\,
	\mathrm{KL}\!\left(p(\cdot\mid z_t)\,\|\,q_\theta(\cdot\mid z_t)\right).
	\]
	Taking expectation over $(t,z_t)$ gives
	\begin{align}
		\mathbb{E}_{t\sim\tau,\,z_t}\big[\|v_\theta(z_t,t)-v^*(z_t,t)\|_2^2\big]
		&\le
		2R^2\,
		\mathbb{E}_{t\sim\tau,\,z_t}\!\left[
		\frac{1}{(1-t)^2}
		\mathrm{KL}\!\left(p(\cdot\mid z_t)\,\|\,q_\theta(\cdot\mid z_t)\right)
		\right].
		\label{eq:weighted_kl_bound}
	\end{align}
	This step keeps the time-dependent weight inside the expectation and therefore does not require any independence assumption between $t$, $z_t$, and the posterior error.
	
	\paragraph{Step 5: Relate weighted KL to weighted cross-entropy suboptimality.}
	For each $(t,z_t)$, the standard decomposition holds:
	\begin{equation}
		\label{eq:wce_decomp}
		\mathbb{E}_{c\sim p(\cdot\mid z_t)}[-\log q_\theta(c\mid z_t)]
		=
		H(p(\cdot\mid z_t))
		+
		\mathrm{KL}\!\left(p(\cdot\mid z_t)\,\|\,q_\theta(\cdot\mid z_t)\right).
	\end{equation}
	Multiplying both sides by $(1-t)^{-2}$ and taking expectation over $(t,z_t)$ yields
	\begin{equation}
		\label{eq:wce_minus_opt}
		\mathcal{L}_{\mathrm{wCE}}(\theta)-\mathcal{L}_{\mathrm{wCE}}^*
		=
		\mathbb{E}_{t\sim\tau,\,z_t}\!\left[
		\frac{1}{(1-t)^2}
		\mathrm{KL}\!\left(p(\cdot\mid z_t)\,\|\,q_\theta(\cdot\mid z_t)\right)
		\right].
	\end{equation}
	
	\paragraph{Step 6: Conclude the stated bound.}
	Combining \eqref{eq:weighted_kl_bound} and \eqref{eq:wce_minus_opt} gives
	\[
	\mathbb{E}_{t\sim\tau,\,z_t}\big[\|v_\theta(z_t,t)-v^*(z_t,t)\|_2^2\big]
	\le
	2R^2\big(\mathcal{L}_{\mathrm{wCE}}(\theta)-\mathcal{L}_{\mathrm{wCE}}^*\big),
	\]
	which is exactly Eq.~\eqref{eq:ce_to_velocity_mse}.
	\qed

	\section{Theoretical Perspective on the Kernel-Smoothed Anchor Prior}
	\label{sec:appendix_kernel_theory}
	
	This section explains why the anchor prior in \method{} is implemented as a kernel-smoothed distribution over learned low-rank latent coordinates. The purpose of this prior is intentionally limited: it provides a smooth initialization distribution for the subsequent categorical flow, rather than serving as a standalone generator in raw time-series space.
	
	\paragraph{Kernel smoothing as an explicit RBF mixture.}
	Let $\{\vect{u}_i\}_{i=1}^N \subset \R^r$ denote the learned coordinates of VQ-latent anchors. The anchor prior used by \method{} is
	\begin{equation}
		p_{\mathrm{anchor}}(\vect{u})
		=
		\frac{1}{N}\sum_{i=1}^{N} K_h(\vect{u}-\vect{u}_i),
		\label{eq:appendix_anchor_kde}
	\end{equation}
	where $K_h(\vect{u})=h^{-r}K(\vect{u}/h)$ is a Gaussian kernel with bandwidth $h$. Eq.~\eqref{eq:appendix_anchor_kde} can be viewed as an equally weighted RBF mixture whose centers are the learned latent anchors. Thus, instead of training an additional neural density estimator to recover the support of the anchor distribution, SDFlow uses the corresponding closed-form smoother directly. This avoids an extra non-convex optimization problem and makes the role of the prior transparent: it locally smooths the latent anchor support before the learned flow dynamics take over.

	The bandwidth is calibrated from the geometry of the learned coordinates rather than specified as a raw noise magnitude. Specifically, let
	\[
	\bar d_{\mathrm{NN}}(U)=\frac{1}{N}\sum_{i=1}^{N}\min_{j\ne i}\|\vect{u}_i-\vect{u}_j\|_2,
	\qquad h=\alpha\,\bar d_{\mathrm{NN}}(U),
	\]
	where $\alpha$ is a dimensionless calibration factor and $h$ is the realized per-coordinate standard deviation. Thus, the numerical value of $\alpha$ should not be interpreted as the value of $h$. For example, a factor applied to a larger nearest-neighbor scale can produce the same effective bandwidth as a larger factor applied to a smaller scale. With $\epsilon\sim\mathcal{N}(0,I_r)$, the perturbation is $h\epsilon$ and its expected radial magnitude scales as $h\sqrt{r}$. This distinction is important because the coordinate dimension and normalization determine the realized perturbation.
	
	\paragraph{Why a learned neural prior is unnecessary here.}
	A flexible neural prior could in principle be trained to approximate the same anchor distribution. However, doing so would introduce an additional modeling problem whose optimum is not needed for our objective. The initializer only needs to place mass near the low-rank VQ-latent scaffold; it does not need to decode samples or model temporal dynamics. A kernel mixture with centers fixed at the learned anchors is therefore the direct non-parametric estimator of this local support. The learned velocity field $v_\theta$, not the kernel prior, is responsible for transporting initialized points to valid VQ endpoints and for producing final samples.
	
	\paragraph{Effective dimension of latent smoothing.}
	The statistical benefit of kernel smoothing depends on the dimension of the support on which it is applied. Assume that the anchor coordinates are sampled from a twice continuously differentiable density $f_0$ supported on an intrinsic $r$-dimensional latent coordinate manifold, and that the kernel has finite second moment. The standard bias--variance decomposition for kernel density estimation gives
	\begin{equation}
		\mathrm{MISE}(h)
		:=
		\E\!\left[\int \big(p_{\mathrm{anchor}}(\vect{u})-f_0(\vect{u})\big)^2 d\vect{u}\right]
		=
		O(h^4)+O\!\left(\frac{1}{N h^r}\right).
		\label{eq:kde_mise_decomp}
	\end{equation}
	Balancing the bias term $O(h^4)$ and the variance term $O((Nh^r)^{-1})$ yields the usual bandwidth scaling $h \asymp N^{-1/(r+4)}$ and the rate
	\begin{equation}
		\mathrm{MISE}_{\mathrm{latent}}
		=
		O\!\left(N^{-\frac{4}{r+4}}\right).
		\label{eq:kde_mise_latent}
	\end{equation}
	If the same kernel smoother were applied directly in the ambient raw trajectory space of dimension $D$, the corresponding rate would be
	\begin{equation}
		\mathrm{MISE}_{\mathrm{raw}}
		=
		O\!\left(N^{-\frac{4}{D+4}}\right),
		\label{eq:kde_mise_raw}
	\end{equation}
	which is substantially slower when $D \gg r$. It shows why SDFlow performs smoothing in the learned low-rank latent coordinate space, where local neighborhoods are statistically meaningful, instead of in raw time-series space.
	
	\paragraph{Role in SDFlow.}
	The kernel prior should be understood as \emph{latent support smoothing}, not as sample retrieval. Sampling from $p_{\mathrm{anchor}}$ produces an initialization $\vect{z}_0=\mathrm{normalize}(\vect{u}\mat{V}^\top)$ near the shared low-rank scaffold. The final sample is then obtained only after integrating the learned categorical velocity field and decoding through the VQ representation. Consequently, any potential anchor-level collapse must be evaluated at the level of the complete generator, not by inspecting the kernel prior in isolation. This is why we complement the theoretical perspective above with three empirical checks: the KDE-only baseline, raw-space nearest-neighbor audit, and held-out latent-flow generalization study.
	\section{Experimental Setup Details}
	\label{app:setup}
	
	\textbf{Datasets.}
	We evaluate on four datasets spanning diverse domains and complexities, following the protocol established by prior work~\citep{yoon2019time,chen2024sdformer}:
	
	\begin{itemize}
		\item \textbf{Sines}: Synthetic sinusoidal sequences with varying frequencies and phases, 5 features. This serves as a sanity check for basic temporal pattern capture.
		\item \textbf{Stocks}: Daily Google stock prices from 2004--2019, including open/close/high/low prices and volume, 6 features. Real financial data with complex non-stationary dynamics.
		\item \textbf{ETTh}: Hourly electrical transformer temperature measurements, 7 features. Industrial time series with periodic patterns and trend components.
		\item \textbf{Energy}: UCI appliance energy use dataset with weather and environmental sensors, 28 features. High-dimensional multivariate series with complex cross-correlations.
	\end{itemize}
	
	\textbf{Evaluation Metrics.}
	We adopt three complementary metrics following standard practice:
	
	\begin{itemize}
		\item \textbf{Discriminative Score (DS)}: A post-hoc 2-layer LSTM classifier is trained to distinguish real from synthetic sequences using a 80/20 train/test split. DS $= |\text{accuracy} - 0.5|$, where lower values indicate synthetic data indistinguishable from real (optimal is 0).
		\item \textbf{Predictive Score (PS)}: A sequence model is trained on synthetic data to predict the next step, then evaluated on real test data. Lower prediction error indicates synthetic data preserves predictive structure.
		\item \textbf{Context-FID}: Fr\'echet Inception Distance computed on contextualized embeddings from a pre-trained TS2Vec encoder. Lower values indicate distributional similarity.
	\end{itemize}
	
	\textbf{Baselines.}
	We compare against representative methods from each paradigm:
	
	\begin{itemize}
		\item \emph{GAN-based}: TimeGAN~\citep{yoon2019time} (uses generative adversarial networks with an embedding function and supervised loss to capture temporal dynamics); COT-GAN~\citep{xu2020cot} (combines GAN with causal optimal transport principles for efficient and stable generation of time series data);
		\item \emph{VAE-based}: TimeVAE~\citep{desai2021timevae} (employs variational autoencoders with an interpretable temporal structure for time series synthesis);
		\item \emph{Diffusion-based}: DiffWave~\citep{kong2021diffwave} (directly applies denoising diffusion probabilistic models to waveform generation); DiffTime~\citep{coletta2024constrained} (harnesses score-based diffusion models for time series generation); Diffusion-TS~\citep{yuan2024diffusion} (utilizes an encoder-decoder transformer with disentangled temporal representations for diffusion-based generation);
		\item \emph{Discrete token}: SDformer-m (masked)~\citep{chen2024sdformer} (uses masked token modeling with random masking and iterative decoding); SDformer-ar (autoregressive)~\citep{chen2024sdformer} (uses autoregressive token modeling with random replacement to mitigate training-inference inconsistency);
		\item \emph{Flow-based}: FlowTS~\citep{hu2025flowts} (employs the Rectified Flow framework, learn the linear transmission path in the probability space).
	\end{itemize}
	
	\textbf{Implementation Details.}
	We use identical VQ-VAE tokenizers to SDformer for fair comparison: codebook size $K=512$, latent dimension $d_c=512$, downsampling rate $s=4$. For \method{}: $L=24$ uses rank $r=256$, 1 DiT layer; $L=256$ uses rank $r=1024$, 3 DiT layers. The anchor-prior bandwidth is calibrated from the learned-coordinate nearest-neighbor scale as described in Appendix~\ref{sec:appendix_kernel_theory}. All experiments report mean and standard deviation over 5 random seeds.
	
	\subsection{Hyperparameter Configurations}
	\label{app:hyperparams}
	
	Table~\ref{tab:vqvae_hyperparams} and Table~\ref{tab:fm_hyperparams} provides complete hyperparameter configurations for \method{} across different sequence lengths.
	
	\begin{table}[h]
		\caption{VQ-VAE hyperparameter configurations for \method{} across sequence lengths.}
		\label{tab:vqvae_hyperparams}
		\centering
		\small
		\resizebox{0.8\columnwidth}{!}{
			\begin{tabular}{@{}lcccccc@{}}
				\toprule
				\textbf{Setting} & \textbf{Hidden dim} & \textbf{Enc/Dec Layers} & \textbf{K} & \textbf{$d_c$} & \textbf{$\lambda$} & \textbf{down. rate} \\
				\midrule
				$L=24$ & 512 & 2 & 512/1024 & 256/512 & 0.01/0.5/2.0 & 2/4 \\
				$L=64$ & 512 & 2 & 512/1024 & 256/512 & 0.01/0.5/2.0 & 4 \\
				$L=128$ & 512 & 2 & 512/1024 & 256/512 & 0.01/0.5/2.0 & 4 \\
				$L=256$ & 512 & 2 & 512/1024 & 256/512 & 0.01/0.5/2.0 & 4 \\
				\bottomrule
			\end{tabular}
		}
	\end{table}
	
	\begin{table}[h]
		\caption{Flow Matching hyperparameter configurations for \method{} across sequence lengths.}
		\label{tab:fm_hyperparams}
		\centering
		\small
		\resizebox{0.75\columnwidth}{!}{
			\begin{tabular}{@{}lccccccc}
				\toprule
				\textbf{Setting} & \textbf{ODE Steps} & \textbf{Rank} & \textbf{$d_{\text{model}}$} & \textbf{Layers} & \textbf{Heads} & \textbf{$\lambda_\mu$} & \textbf{$\lambda_\sigma$} \\
				\midrule
				$L=24$ & 20 & 256 & 512 & 1 & 16 & 0.1 & 10.0 \\
				$L=64$ & 20 & 256/512 & 1024 & 3 & 16 & 0.1 & 10.0 \\
				$L=128$ & 20 & 512 & 1024 & 3 & 16 & 0.1 & 10.0 \\
				$L=256$ & 20 & 1024 & 1024 & 3 & 16 & 0.1 & 10.0 \\
				\bottomrule
			\end{tabular}
		}
	\end{table}
	
	\section{Core Algorithms}
	\label{app:algorithm}
	Algorithm~\ref{alg:sdflow_training} and Algorithm~\ref{alg:sdflow_inference} are the core algorithms for our model training and inference.
	
	\begin{algorithm}[h]
		\caption{SDFlow Training: Manifold-Anchored Categorical Flow Matching}
		\label{alg:sdflow_training}
		\begin{algorithmic}[1]
			\STATE \textbf{Input:} VQ-VAE codebook $\mathcal{C} = \{c_k\}_{k=1}^K$, training indices $\{y_i\}_{i=1}^N$
			\STATE \textbf{Parameters:} Rank $r$, regularization coefficients $\lambda_\mu, \lambda_\sigma$
			\STATE \textbf{Output:} Trained flow network $v_\theta$, low-rank factors $U, V$
			
			\STATE
			\STATE \textcolor{blue}{\textbf{// Stage 1: VQ-VAE Tokenization (Frozen)}}
			\STATE Pre-train VQ-VAE with similarity-driven quantization
			\STATE Extract codebook embeddings: $Z = [c_{y_1}, \ldots, c_{y_N}] \in \mathbb{R}^{N \times D}$
			
			\STATE
			\STATE \textcolor{blue}{\textbf{// Stage 2: Manifold-Anchored Flow Matching}}
			\STATE \textbf{Initialize:}
			\STATE \quad Flow network $\theta$ (DiT architecture)
			\STATE \quad $U \in \mathbb{R}^{M \times r} \sim \mathcal{N}(0, 0.01^2 I)$ \quad \textcolor{gray}{\# Manifold support anchors}
			\STATE \quad $V \in \mathbb{R}^{D \times r} \sim \mathcal{N}(0, 0.01^2 I)$ \quad \textcolor{gray}{\# Subspace basis}
			
			\FOR{epoch $= 1$ to $T_{\text{max}}$}
			\STATE Sample batch indices $\mathcal{B} = \{i_1, \ldots, i_B\}$
			\STATE Sample $u \sim \text{Uniform}(0,1)$ and set $t = 1-\cos(\pi u^2/2)$
			
			\STATE \textcolor{blue}{\textbf{// Construct Flow Path}}
			\STATE Retrieve coordinates: $U_\mathcal{B} = [u_{i_1}, \ldots, u_{i_B}]^\top \in \mathbb{R}^{B \times r}$
			\STATE Retrieve targets: $Z_\mathcal{B} = [z_{i_1}, \ldots, z_{i_B}]^\top \in \mathbb{R}^{B \times D}$
			\STATE During training, apply Manifold Mixup with probability $p=.5$ to paired coordinates and targets.
			
			\STATE \textcolor{blue}{\textbf{// Low-rank Manifold Anchoring}}
			\STATE $Z_0 = U_\mathcal{B} V^\top \in \mathbb{R}^{B \times D}$ \quad \textcolor{gray}{\# Project to subspace}
			\STATE $Z_0 \gets Z_0 / \|Z_0\|_2$ \quad \textcolor{gray}{\# Unit normalization}
			\STATE $Z_t \gets (1 - t) Z_0 + t \cdot Z_\mathcal{B}$ \quad \textcolor{gray}{\# Linear interpolation}
			
			\STATE \textcolor{blue}{\textbf{// Categorical Flow Prediction}}
			\STATE $\text{logits} \gets v_\theta(Z_t, t)$ \quad \textcolor{gray}{\# Shape: $[B, L, K]$}
			
			\STATE \textcolor{blue}{\textbf{// Cross-Entropy Loss with Regularization}}
			\STATE $\mathcal{L}_{\text{CE}} \gets -\frac{1}{BL} \sum_{i,\ell} \log \text{softmax}(\text{logits}_{i,\ell})[y_{i,\ell}]$
			\STATE $\mathcal{L}_\mu \gets \|\mathbb{E}_i[u_i]\|_2$ \quad \textcolor{gray}{\# Mean regularization}
			\STATE $\mathcal{L}_\sigma \gets \|\text{Std}_i(u_i) - 1\|_1$ \quad \textcolor{gray}{\# Variance regularization}
			\STATE $\mathcal{L}_{\text{str}} \gets \text{normalized pairwise-distance alignment}(U_\mathcal{B}, Z_\mathcal{B})$
			\STATE $\mathcal{L} \gets \mathcal{L}_{\text{CE}} + \lambda_\mu \mathcal{L}_\mu + \lambda_\sigma \mathcal{L}_\sigma + 10\mathcal{L}_{\text{str}}$
			
			\STATE \textcolor{blue}{\textbf{// Optimization}}
			\STATE Update $\theta$ with learning rate $\eta_\theta = 10^{-4}$
			\STATE Update $U, V$ with learning rate $\eta_{UV} = 10^{-3}$
			\ENDFOR
			
			\RETURN $v_\theta, U, V$
		\end{algorithmic}
	\end{algorithm}
	
	\begin{algorithm}[h]
		\caption{SDFlow Inference: Parallel Time Series Generation}
		\label{alg:sdflow_inference}
		\begin{algorithmic}[1]
			\STATE \textbf{Input:} Trained flow $v_\theta$, coordinates $U$, basis $V$, codebook $\mathcal{C}$
			\STATE \textbf{Parameters:} Bandwidth factor $\alpha$, temperature $\tau$, steps $S$
			\STATE \textbf{Output:} Generated time series $\{\hat{X}_i\}_{i=1}^{N_{\text{gen}}}$
			
			\STATE
			\STATE \textcolor{blue}{\textbf{// Anchor Prior Construction}}
			\STATE Compute pairwise distances: $d_i = \min_{j \neq i} \|u_i - u_j\|_2$ for $i=1,\ldots,N$
			\STATE Set bandwidth: $h = \alpha \cdot \frac{1}{N}\sum_{i=1}^N d_i$ \quad \textcolor{gray}{\# $\alpha$: calibration factor; $h$: per-coordinate std.}
			
			\STATE
			\STATE \textcolor{blue}{\textbf{// Parallel Generation (No Sequential Dependency)}}
			\FOR{$i = 1$ to $N_{\text{gen}}$}
			\STATE Sample center index: $j \sim \text{Uniform}(1, N)$
			\STATE Sample noise: $\epsilon \sim \mathcal{N}(0, h^2 I_r)$
			\STATE $u_{\text{new}} \gets u_j + \epsilon$ \quad \textcolor{gray}{\# Kernel-smoothed anchor sampling}
			\STATE $z_0 \gets \text{normalize}(u_{\text{new}} V^\top)$ \quad \textcolor{gray}{\# Manifold anchor}
			
			\STATE \textcolor{blue}{\textbf{// ODE Integration}}
			\FOR{$s = 0$ to $S-1$}
			\STATE $t \gets s / S$
			\STATE $\text{logits} \gets v_\theta(z_t, t)$
			\STATE $\pi \gets \text{softmax}(\text{logits} / \tau)$
			\STATE $\mu \gets \sum_k \pi_k \cdot c_k$ \quad \textcolor{gray}{\# Expected embedding}
			\STATE $v \gets (\mu - z_t) / (1 - t)$
			\STATE $z_{t + 1/S} \gets z_t + v / S$
			\ENDFOR
			
			\STATE \textcolor{blue}{\textbf{// Quantization and Decoding}}
			\STATE $\hat{y}_i \gets \arg\max_k \, z_1 \cdot c_k$ \quad \textcolor{gray}{\# Nearest codebook}
			\STATE $\hat{X}_i \gets \mathcal{D}(\hat{y}_i)$ \quad \textcolor{gray}{\# VQ-VAE decoder}
			\ENDFOR
			
			\RETURN $\{\hat{X}_i\}_{i=1}^{N_{\text{gen}}}$
		\end{algorithmic}
	\end{algorithm}
	
	\section{Manifold Analysis}
	\label{app:manifold_analysis}
\begin{figure}[h]
		\centering
		\includegraphics[width=0.6\columnwidth]{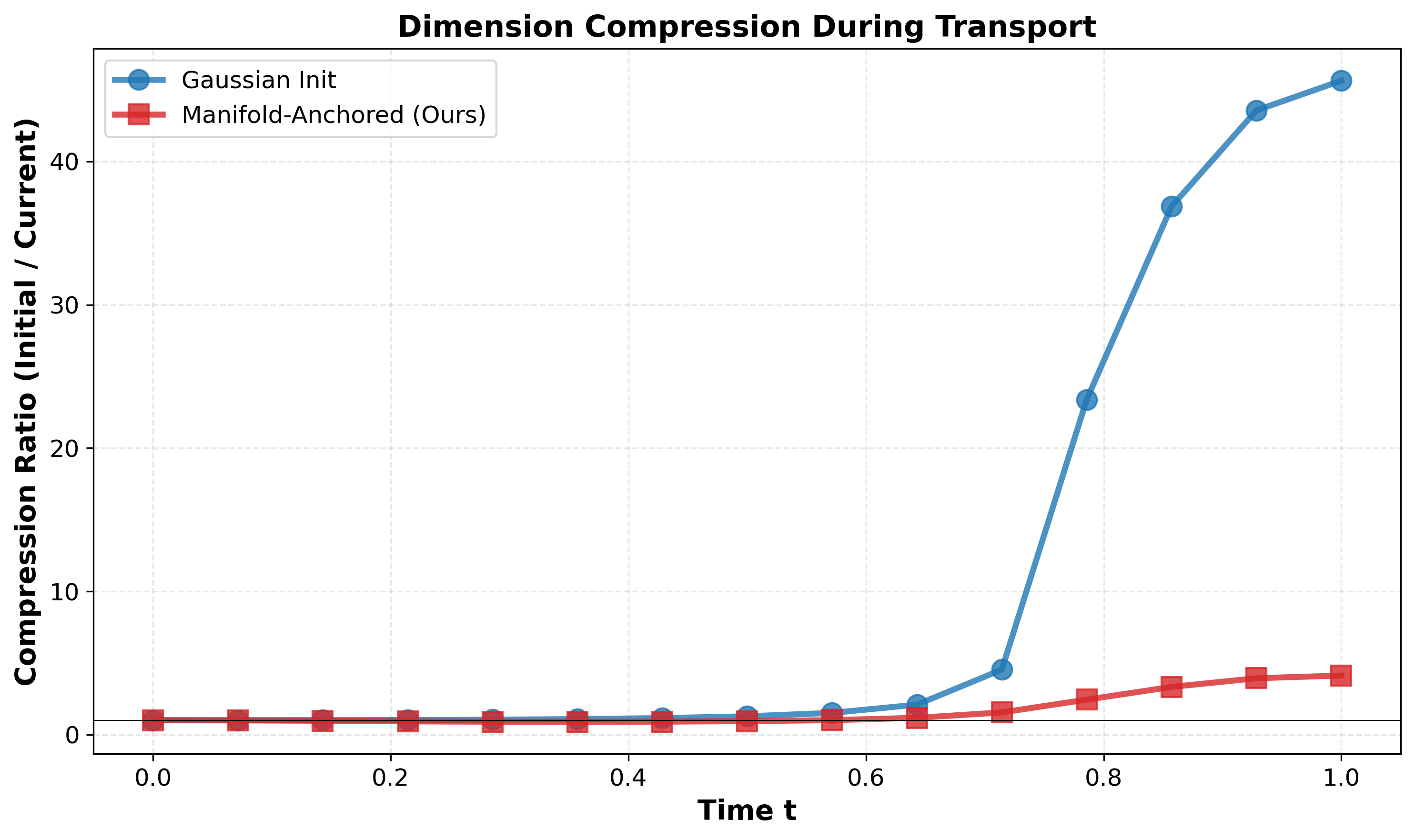}
		\caption{Dimension compression ratio during flow transport.  }
		\label{fig:compression_ratio}
	\end{figure}

\begin{figure}[h]
		\centering
		\includegraphics[width=0.6\columnwidth]{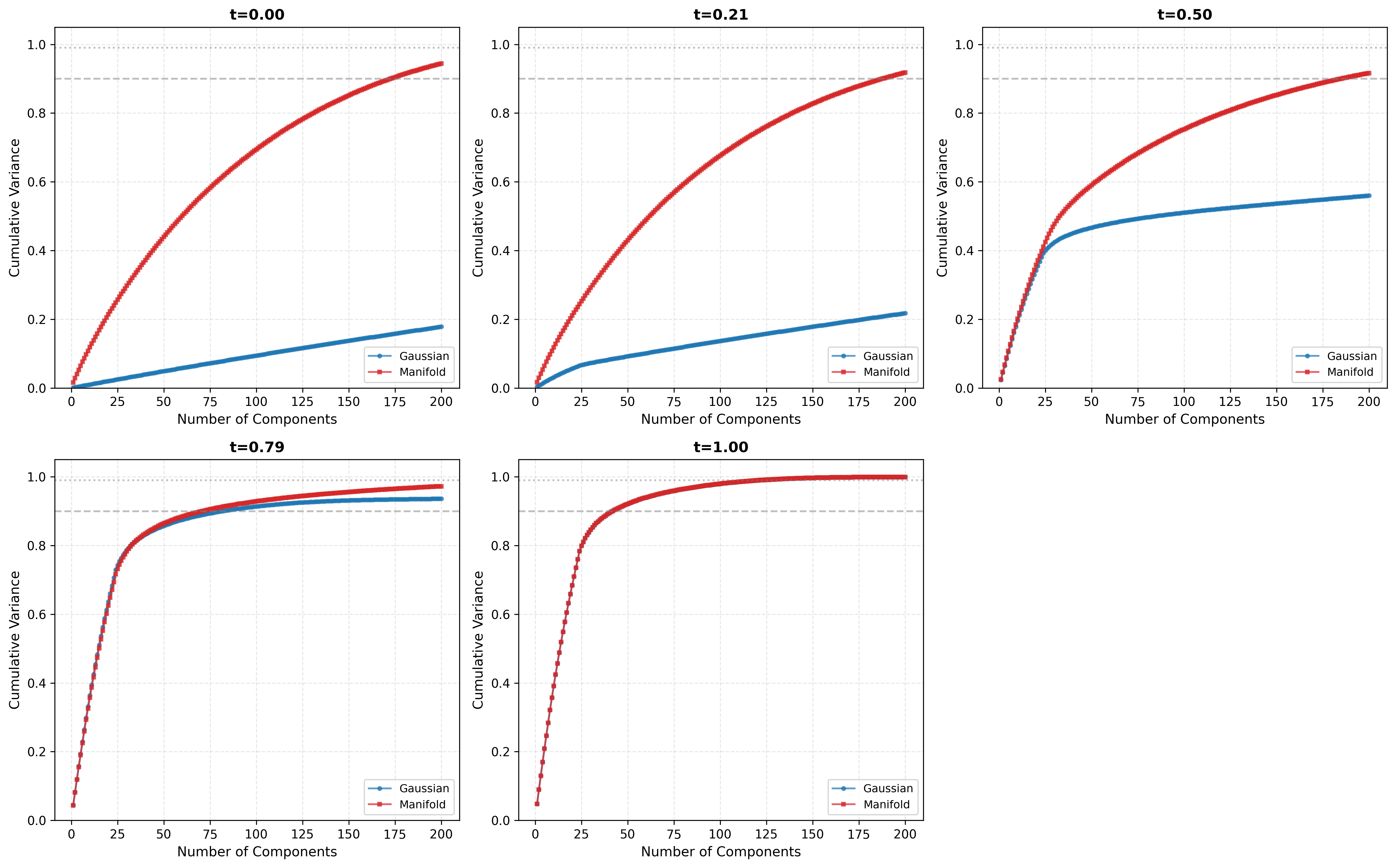}
		\caption{Cumulative variance detailed.  }
		\label{fig:cumulative_variance_detailed}
	\end{figure}

\begin{figure}[h]
		\centering
		\begin{subfigure}[b]{0.24\columnwidth}
			\centering
			\includegraphics[width=\linewidth]{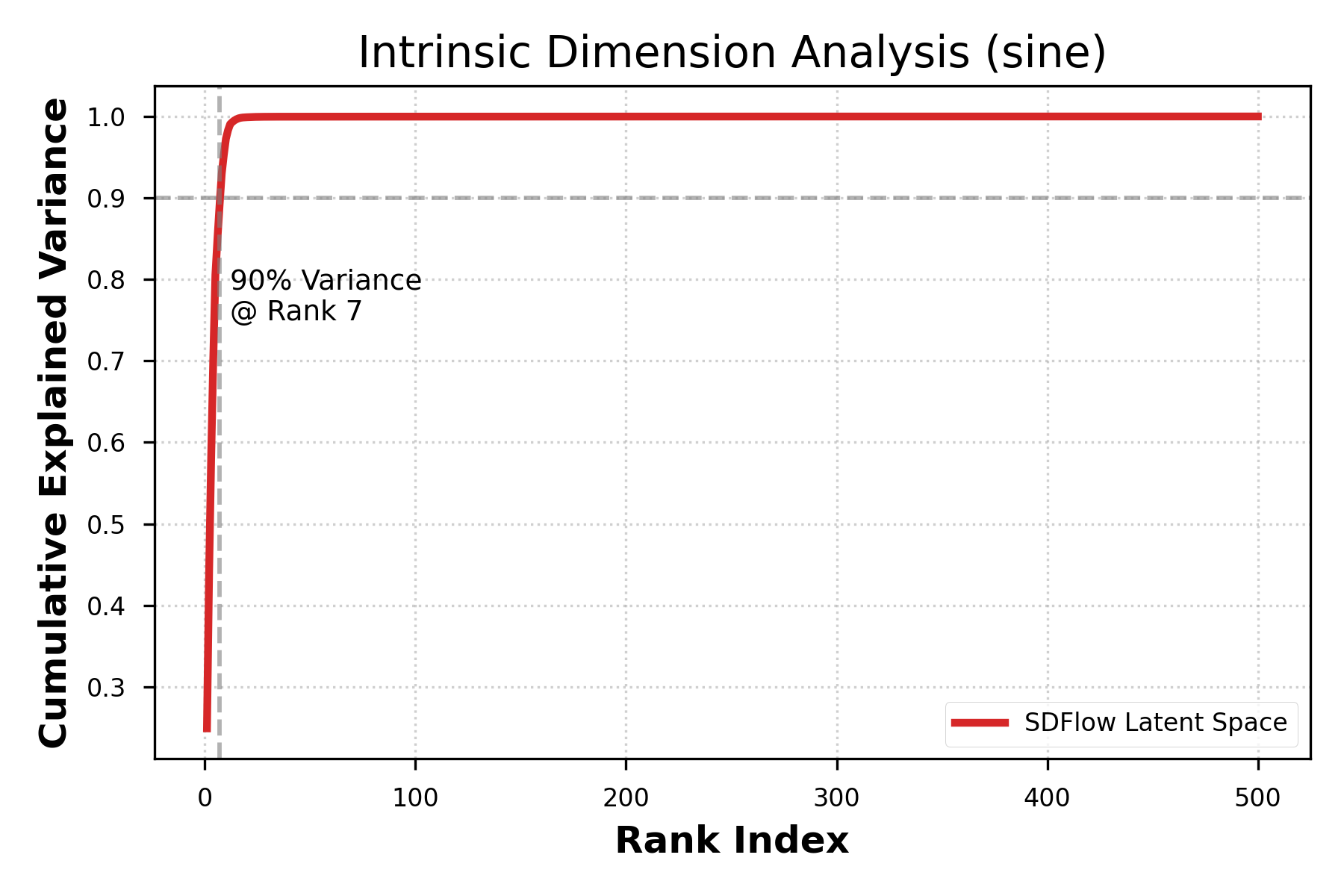}
			\caption{Sine (rank = 7)}
			\label{subfig:sine_svd}
		\end{subfigure}
		\begin{subfigure}[b]{0.24\columnwidth}
			\centering
			\includegraphics[width=\linewidth]{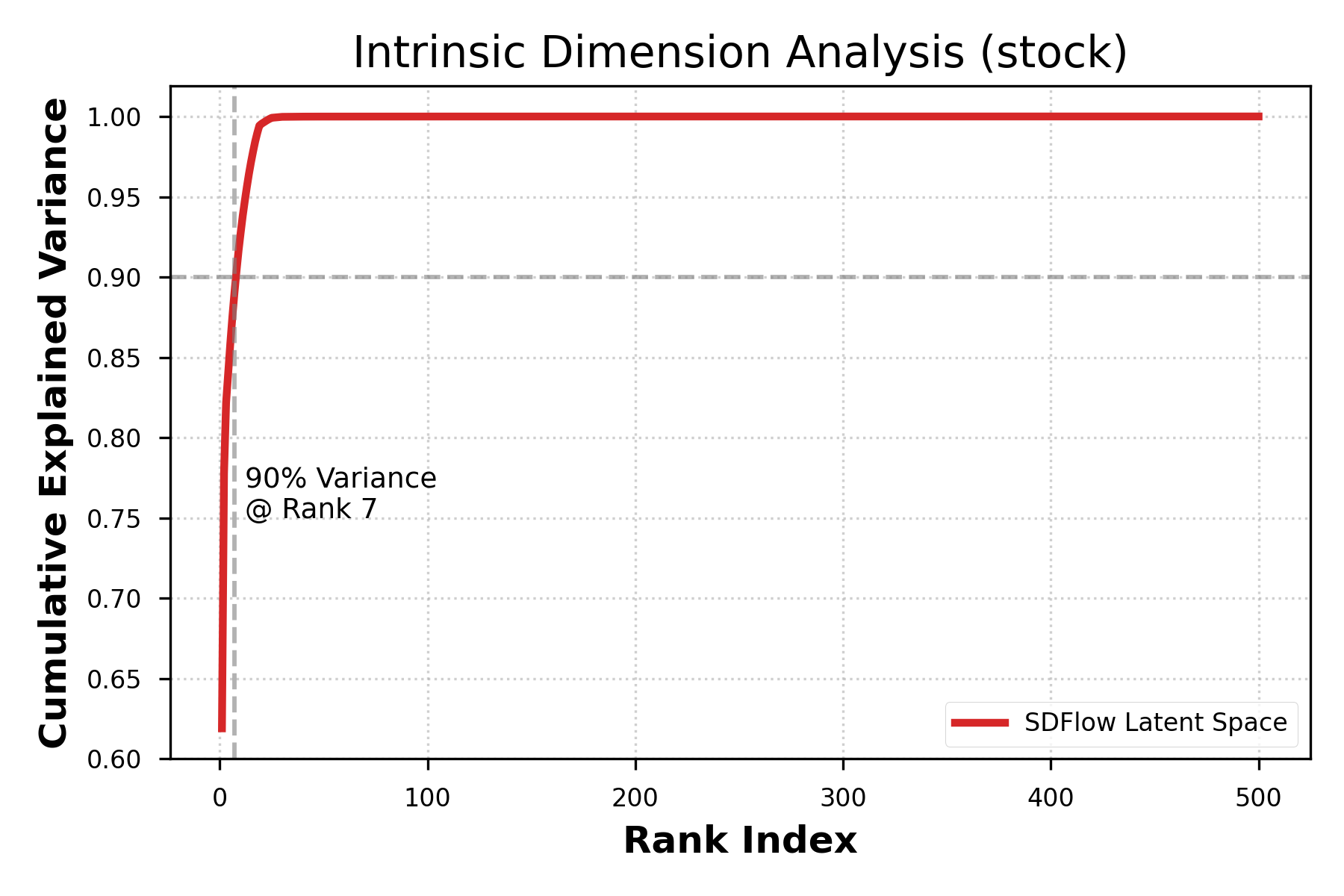}
			\caption{Stock (rank = 7)}
			\label{subfig:stock_svd}
		\end{subfigure}
		\begin{subfigure}[b]{0.24\columnwidth}
			\centering
			\includegraphics[width=\linewidth]{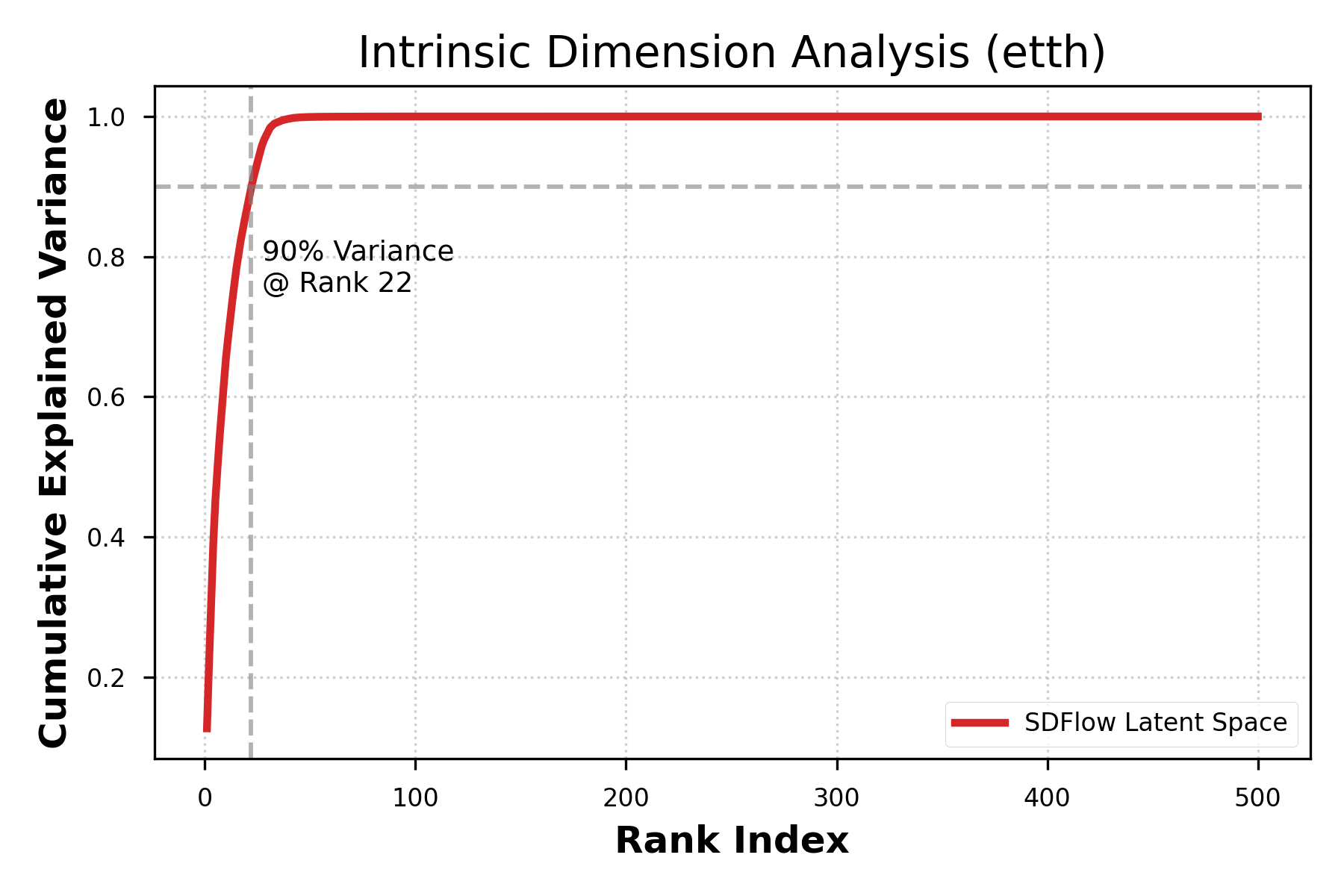}
			\caption{ETTh (rank = 22)}
			\label{subfig:etth_svd}
		\end{subfigure}
		\begin{subfigure}[b]{0.24\columnwidth}
			\centering
			\includegraphics[width=\linewidth]{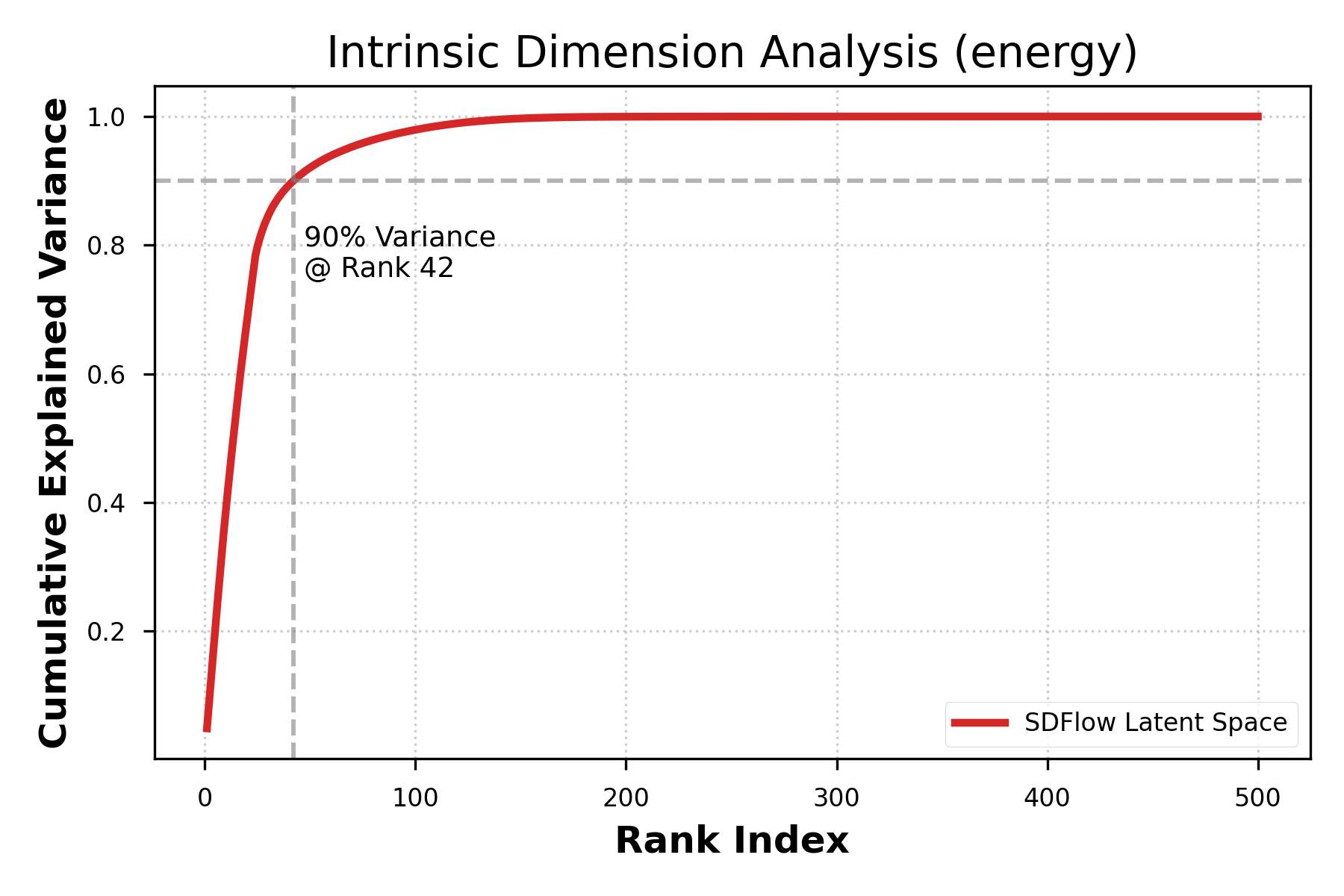}
			\caption{Energy (rank = 42)}
			\label{subfig:energy_svd}
		\end{subfigure}
		
		\caption{SVD analysis of VQ-VAE latent codes across datasets.}
		\label{fig:svd_spectrum_multi_datasets}
	\end{figure}

	Our analysis reveals that Gaussian initialization requires severe 45$\times$ compression from high-dimensional space (rank ~1900) to the target manifold (rank ~50), creating a difficult optimization problem (Figure~\ref{fig:teaser}, Figure~\ref{fig:compression_ratio}, Figure~\ref{fig:cumulative_variance_detailed}, and Figure~\ref{fig:svd_spectrum_multi_datasets}). In contrast, our manifold-anchored approach starts near the data manifold (rank ~180), requiring only smooth 3.6$\times$ compression. This 11$\times$ reduction in compression ratio makes the transport problem substantially easier and supports the importance of initialization distance.

	
	\section{Detailed Ablation, Sensitivity, and Held-out Generalization Analysis}
	\label{app:ablation_details}
	
	Table~\ref{tab:ablation} studies three aspects of \method{}: component necessity, hyperparameter sensitivity, and held-out latent-flow generalization. We provide additional interpretation here to clarify how each result supports the design of SDFlow.

	\paragraph{Structure and Mixup Regularization.}
	To isolate the effect of the two training-only regularizers, we conduct a controlled $2\times2$ ablation on Energy with the architecture, prior, solver, training budget, and evaluation protocol fixed. The complete recipe mainly reaches the fixed CE endpoint earlier, while final generation quality remains nearly unchanged:
	\begin{table}[h]
	\centering
	\small
	\caption{Controlled $2\times2$ ablation of structure regularization and Manifold Mixup on Energy.}
	\label{tab:structure_mixup}
	\resizebox{0.5\linewidth}{!}{%
	\begin{tabular}{@{}ccrrrr@{}}
	\toprule
	$\mathcal{L}_{\mathrm{str}}$ & Mixup & DS$\downarrow$ & PS$\downarrow$ & C-FID$\downarrow$ & Steps@CE$\le .20$ $\downarrow$ \\
	\midrule
	Yes & Yes & .0063 & .2493 & .0014 & 17.0k \\
	Yes & No  & .0076 & .2494 & .0021 & 19.5k \\
	No  & Yes & .0063 & .2493 & .0014 & 18.5k \\
	No  & No  & .0064 & .2493 & .0015 & 19.5k \\
	\bottomrule
	\end{tabular}}
	\end{table}
	The four variants have nearly identical final metrics; the regularizers therefore improve optimization speed without changing the inference architecture or adding inference-time capacity.
	
	\paragraph{Representation Space and Low-Rank Scaffold.}
	Directly applying flow matching in raw data space performs poorly (DS 0.453), confirming that high-dimensional raw trajectories are not a suitable space for Stage-2 transport. After VQ tokenization, however, a standard Gaussian prior in the full latent space still remains weak (DS 0.218), and using anchors without low-rank structure only partially helps (DS 0.173). These results show that VQ tokenization alone is insufficient: the latent space must also be organized into a tractable low-rank coordinate system. The learned low-rank scaffold provides this geometry, turning sparse high-dimensional VQ codes into a smoother coordinate space for flow matching.
	
	\paragraph{Anchor Prior Design.}
	On top of the low-rank scaffold, the initialization prior is crucial. Generic alternatives such as Gaussian sampling (DS 0.218) and an amortized MLP prior (DS 0.213) fail to capture the latent topology. Simple interpolation improves performance (DS 0.047), suggesting that respecting local latent geometry is important, but it remains far from the learned anchor prior (DS 0.006). Thus, the kernel-smoothed anchor prior should be interpreted as a manifold-localized initializer over low-rank VQ-latent supports, while the learned categorical flow performs the actual generative transport.
	
	\paragraph{Codebook Capacity and Solver Steps.}
	Naively reducing the VQ embedding dimension does not solve the transport problem: smaller codebooks without the low-rank scaffold yield worse DS values (0.223 for $d=256$, 0.251 for $d=128$, and 0.299 for $d=64$). This indicates that simple compression sacrifices representation quality rather than producing useful geometry. The solver-step ablation further shows that the method does not rely on excessive numerical refinement: $S=20$ matches $S=50$ (both DS 0.006), while $S=10$ is only moderately worse (DS 0.018). We therefore use $S=20$ as an efficient quality--speed trade-off.
	
	\paragraph{Rank and Bandwidth Sensitivity.}
	The sensitivity study shows that performance is stable around the selected rank and bandwidth. A static SVD rank ($r=42$) is insufficient (DS 0.094), while learned ranks improve substantially, with $r=256$ reaching DS 0.006 and $r=512$ offering no further gain. For the bandwidth, large values such as $h=0.60$ over-smooth the latent support (DS 0.138), whereas overly small values such as $h=0.01$ slightly weaken local connectivity (DS 0.008). The best setting $h=0.06$ balances local smoothing and manifold preservation.
	
	\paragraph{Held-out Latent-Flow Generalization.}
	To test whether Stage-2 generation depends on access to all anchors, we train the latent flow and anchor prior using only subsets of encoded latents while keeping the VQ tokenizer frozen as a shared representation layer. On Energy, using only 10\% of latent anchors still achieves DS 0.009 and C-FID 0.002 on the held-out 90\%; on Sines, the same 10\% setting achieves DS 0.007 and C-FID 0.0013; on ETTh, using 50\% anchors gives DS 0.003 and C-FID 0.002 on the held-out 50\%. The small degradation from the full-anchor setting suggests that the anchor set acts as a compact scaffold for shared latent geometry rather than naive interpolation over latent prototypes.
	
	Overall, the ablation confirms that SDFlow requires all three ingredients: VQ tokenization for representation, low-rank scaffolding for tractable geometry, and anchor-based initialization for local transport. The sensitivity and held-out studies further show that the method is not overly dependent on exact hyperparameter choices or complete anchor coverage, supporting SDFlow as a stable latent-space non-autoregressive generator.

	\section{Comprehensive Generalization Study}
	\label{app:anti_memorization}
	
	While our manifold-anchored initialization efficiently bypasses the curse of dimensionality, a natural concern is whether the prior over latent anchors inadvertently leads to local resampling. We therefore conduct generalization audits from three complementary perspectives: an isolated KDE baseline, raw-space nearest-neighbor analysis, and an additional strict held-out sanity check.
	
	\subsection{KDE is an Initializer, Not a Generator: KDE-only Baseline}
	We emphasize that the KDE prior serves as a manifold-localized initialization mechanism. It is not intended to model the complex data distribution by itself. To isolate the contribution of the categorical flow dynamics, we introduce a \textbf{KDE-only (No Flow)} baseline. In this setting, we sample an anchor coordinate $\mathbf{u} \sim p_{\text{KDE}}$, project it to the latent space $\mathbf{z}_0 = \text{normalize}(\mathbf{u}\mathbf{V}^\top)$, and directly quantize and decode it \textit{without} any ODE flow integration. 
	
	As shown in Table~\ref{tab:kde_only}, the KDE-only baseline fails to generate coherent sequences, yielding poor Discriminative Score (DS) and Context-FID. This indicates that the KDE prior alone is insufficient for generation; the learned categorical flow dynamics are needed to transport the initial anchors to valid, high-fidelity distribution regions.
	
	\begin{table}[h]
		\centering
		\caption{KDE-only baseline vs. Full SDFlow on Energy ($L=24$). Bypassing the flow network leads to generation failure, indicating that the KDE prior acts as an initializer rather than a standalone generator.}
		\label{tab:kde_only}
		\resizebox{0.7\linewidth}{!}{
			\begin{tabular}{@{}lcc@{}}
				\toprule
				\textbf{Method} & \textbf{Discriminative Score (DS) $\downarrow$} & \textbf{Context-FID $\downarrow$} \\ \midrule
				KDE-only (No Flow Integration) & 0.462 $\pm$ .015 & 3.105 $\pm$ .124 \\
				SDFlow (Full Model) & \textbf{0.006} $\pm$ .002 & \textbf{0.001} $\pm$ .000 \\ \bottomrule
			\end{tabular}
		}
	\end{table}
	
	\subsection{Raw-Space Nearest-Neighbor Generalization Audit}
	To assess sample-level generalization, we conduct a Nearest-Neighbor (NN) audit directly in the \textit{continuous raw time-series space} (rather than the VQ latent space). Let $T=\{x_i\}_{i=1}^N$ be the training sequences and $Q=\{y_m\}_{m=1}^M$ be generated or held-out-real queries, each in $\mathbb{R}^{L\times C}$. We use the normalized Frobenius distance
	\[
	\delta(a,b)=\frac{\|a-b\|_F}{\sqrt{LC}},\qquad i^*(y)=\arg\min_i\delta(y,x_i),\qquad r_i=\min_{j\ne i}\delta(x_i,x_j).
	\]
	The NN-RMSE is $\mathrm{NN\text{-}RMSE}(Q,T)=M^{-1}\sum_m\delta(y_m,x_{i^*(y_m)})$. Following the sample-level rule of Alaa et al.~\citep{alaa2022faithful}, we define
	\begin{align*}
	\mathrm{Auth}(Q,T)&:=\frac{1}{M}\sum_m\mathbf{1}\!\left[\delta(y_m,x_{i^*(y_m)})>r_{i^*(y_m)}\right],\\
	\mathrm{Prox}_{1\%}(Q,T)&:=\frac{1}{M}\sum_m\mathbf{1}\!\left[\min_i\delta(y_m,x_i)\le q_{.01}\right].
	\end{align*}
	where $q_{.01}$ is the first percentile of the leave-one-out train-to-train radii $\{r_i\}_{i=1}^N$. Generated and real-to-real references use exactly the same definitions; only the query set $Q$ changes.
	
	We define the \textbf{Copy Rate} as the proportion of generated samples that fall within the $1^{\text{st}}$ percentile of the Train $\to$ Train NN distance (i.e., potential near-duplicates). As shown in Table~\ref{tab:copy_rate}, generated samples have similar distances to the training set as real held-out samples do. Furthermore, the Copy Rate is consistently below $1.5\%$.
	
	\begin{table}[h]
		\centering
		\caption{Raw-space Nearest-Neighbor Audit on ETTh and Energy. The generated samples maintain similar distances to the training set as real held-out samples do, with low copy rates.}
		\label{tab:copy_rate}
		\resizebox{\linewidth}{!}{
			\begin{tabular}{@{}lcccc@{}}
				\toprule
				\textbf{Dataset} & \textbf{Held-out $\to$ Train NN Dist.} & \textbf{Generated $\to$ Train NN Dist.} & \textbf{Copy Threshold ($\epsilon$)} & \textbf{Copy Rate $\downarrow$} \\ \midrule
				ETTh ($L=24$) & 0.81 $\pm$ 0.12 & 0.83 $\pm$ 0.14 & 0.15 & 1.2\% \\
				Energy ($L=24$) & 2.25 $\pm$ 0.31 & 2.30 $\pm$ 0.28 & 0.45 & 0.8\% \\ \bottomrule
			\end{tabular}
		}
	\end{table}
	
	\subsection{Additional Strict Held-out Sanity Check}
	As an additional sanity check, we also consider a stricter split than the held-out latent-flow study in Table~\ref{tab:ablation}. Here, \textit{neither} the VQ-VAE tokenizer \textit{nor} the Stage-2 latent flow and anchors are exposed to the held-out split during training. 
	
	We trained the entire SDFlow pipeline on only 80\% of the data and evaluated the generative distributions against the unseen 20\% test data. The results yield a DS of 0.008 and Context-FID of 0.001 on Sines, close to the full-data setting. This supplementary check is not used as the main evidence for our method, but further suggests that SDFlow does not require direct access to every training trajectory.

	\section{Additional Visualizations}
	\label{app:visualizations}
	
\begin{figure}[htbp]
		\centering
		\begin{subfigure}[b]{0.24\columnwidth}
			\centering
			\includegraphics[width=\textwidth]{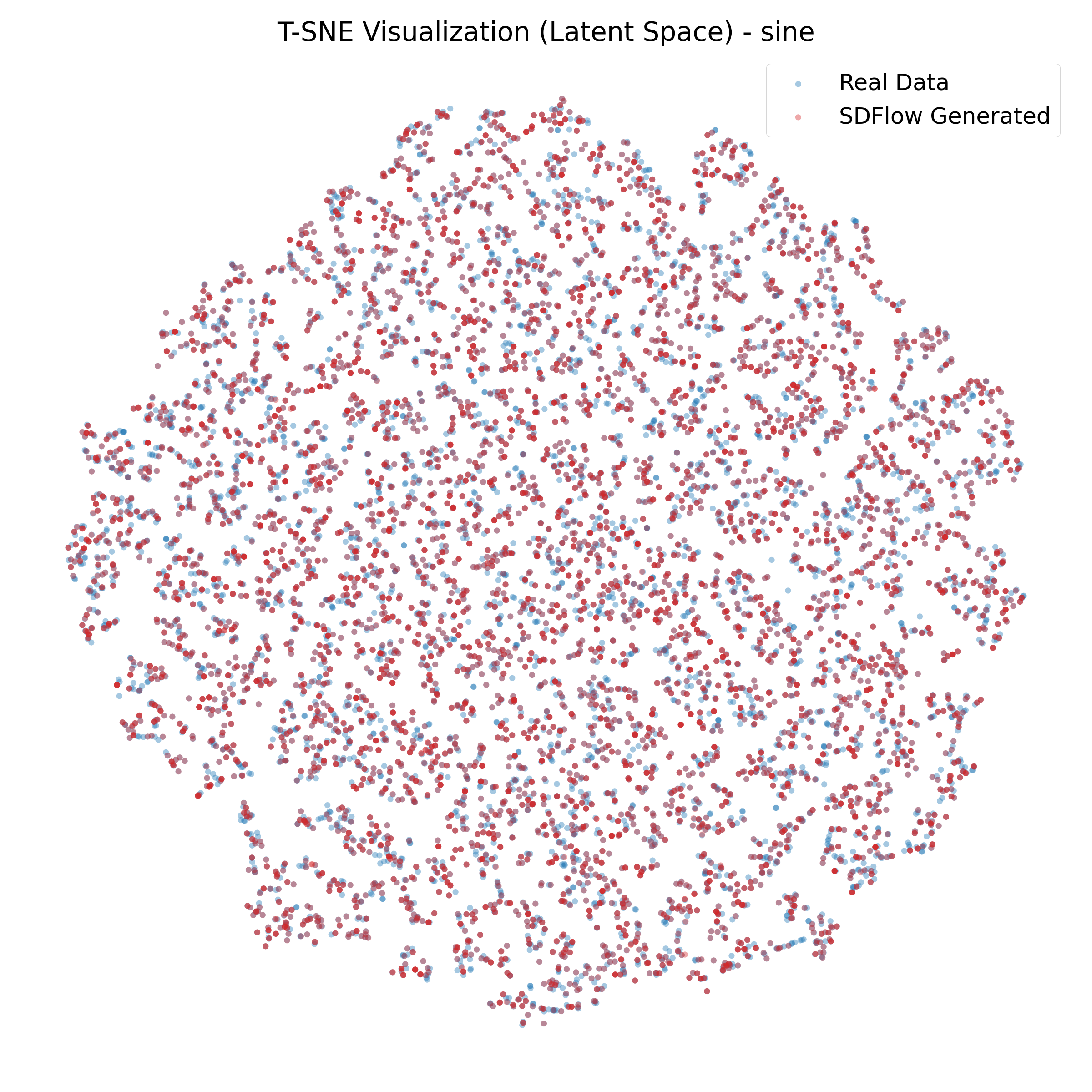}
			\caption{Sine}
		\end{subfigure}
		\hfill
		\begin{subfigure}[b]{0.24\columnwidth}
			\centering
			\includegraphics[width=\textwidth]{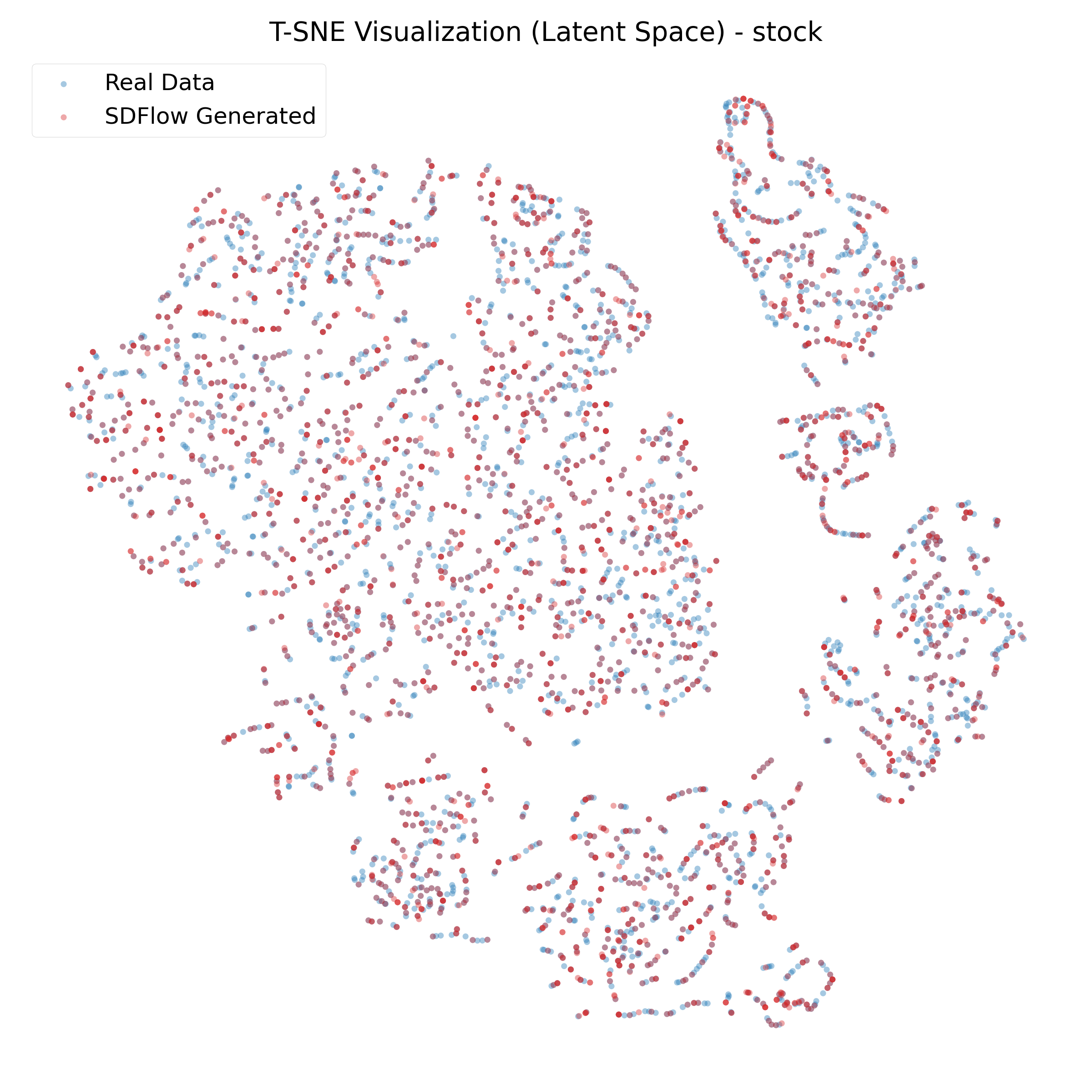}
			\caption{Stock}
		\end{subfigure}
		\hfill
		\begin{subfigure}[b]{0.24\columnwidth}
			\centering
			\includegraphics[width=\textwidth]{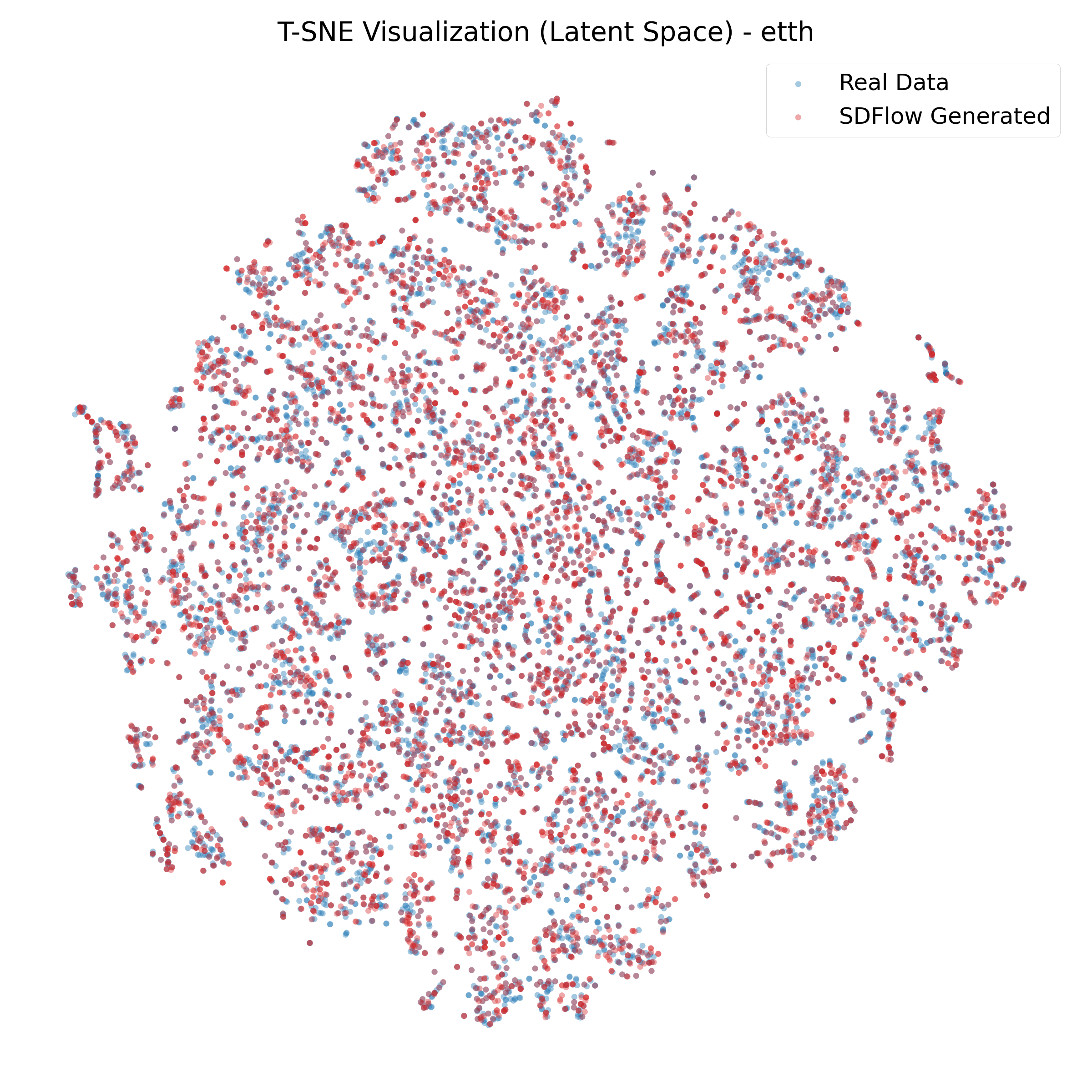}
			\caption{ETTh}
		\end{subfigure}
		\hfill
		\begin{subfigure}[b]{0.24\columnwidth}
			\centering
			\includegraphics[width=\textwidth]{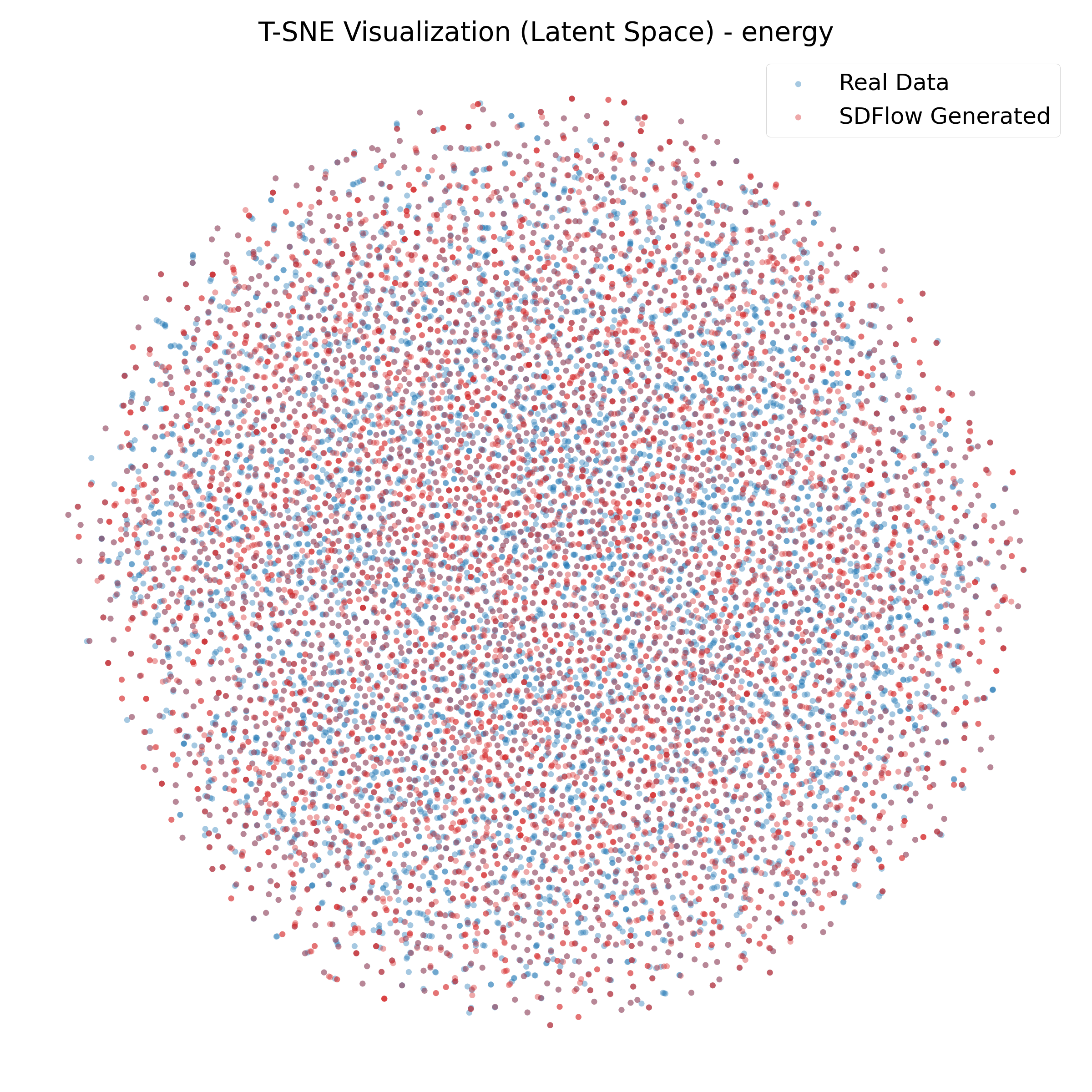}
			\caption{Energy}
		\end{subfigure}
		\caption{t-SNE visualization in the latent space across multiple datasets.}
		\label{fig:t_sne}
	\end{figure}

	Figure~\ref{fig:t_sne} shows t-SNE visualizations of the latent space across datasets. The sample distribution generated by our method closely aligns with the real data distribution, filling the manifold without complete overlap, which is consistent with distribution-level coverage.
	
\begin{figure}[h]
		\centering
		\includegraphics[width=0.8\columnwidth]{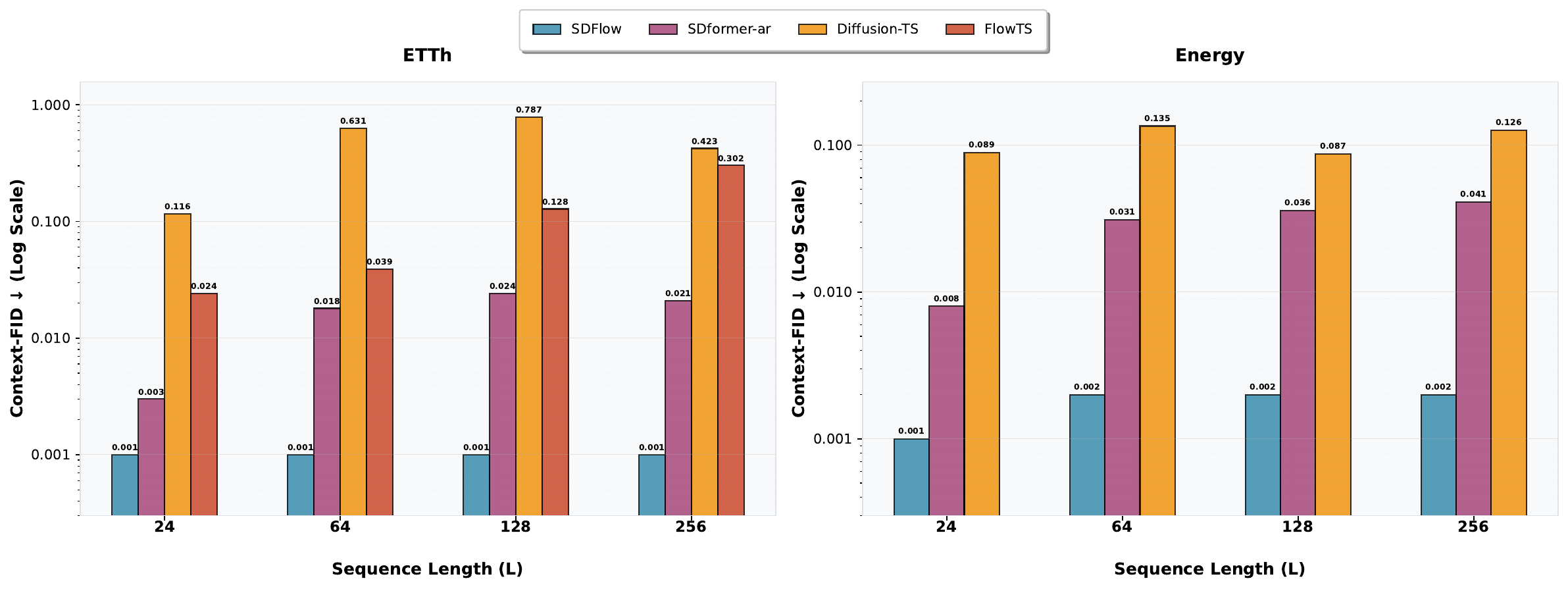}
		\caption{Context-FID across different sequence lengths on ETTh and Energy datasets.}
		\label{fig:fid_comparison}
		\vspace{-3mm}
	\end{figure}

	A comparative analysis of Context-FID is provided in Figure~\ref{fig:fid_comparison}. The visualization shows that \method{} attains markedly lower Context-FID values than existing approaches.
	
\begin{figure}[t]
		\centering
		\begin{subfigure}[b]{0.3\textwidth}
			\centering
			\includegraphics[width=\textwidth]{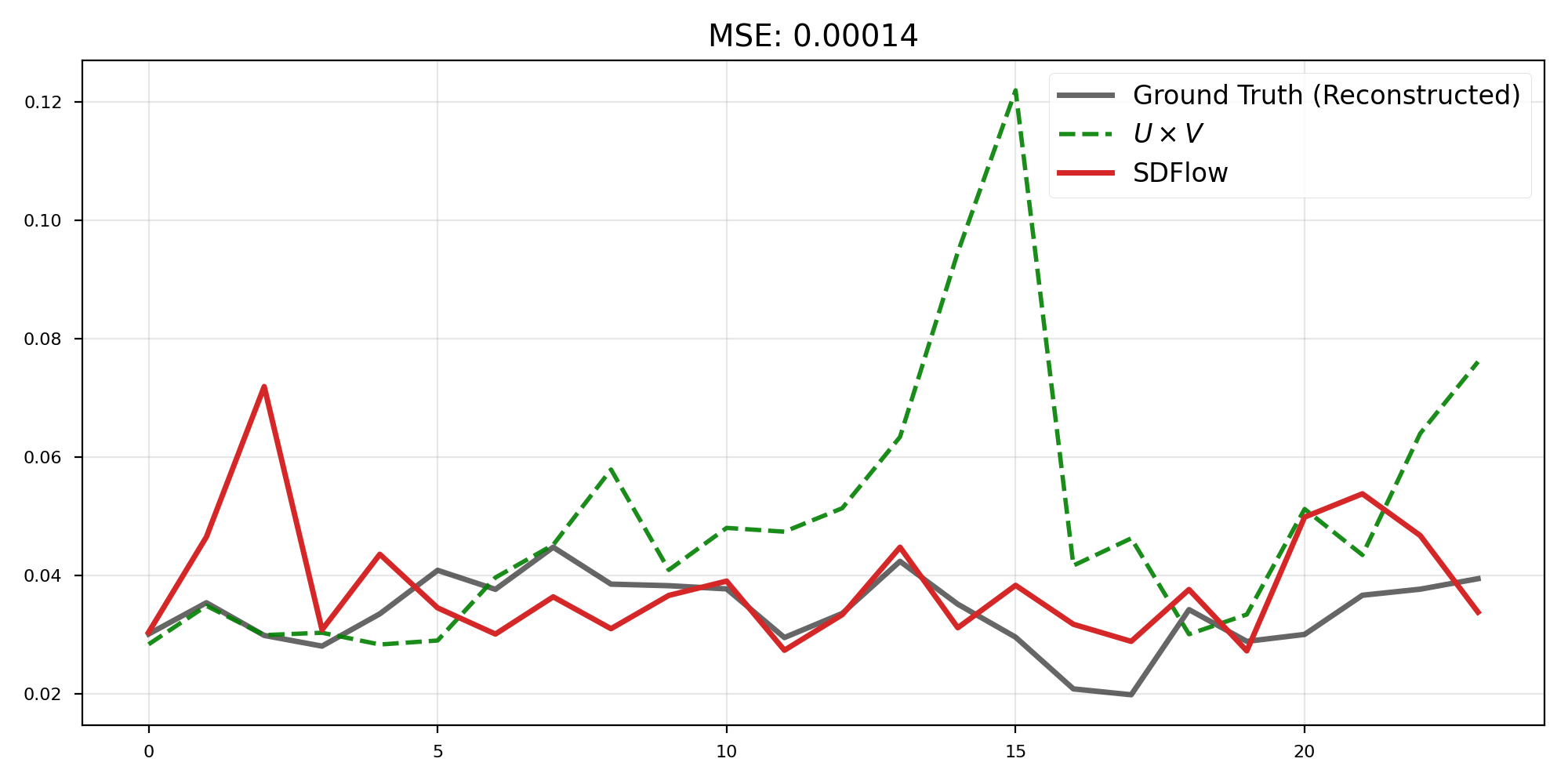}
		\end{subfigure}
		\hfill
		\begin{subfigure}[b]{0.3\textwidth}
			\centering
			\includegraphics[width=\textwidth]{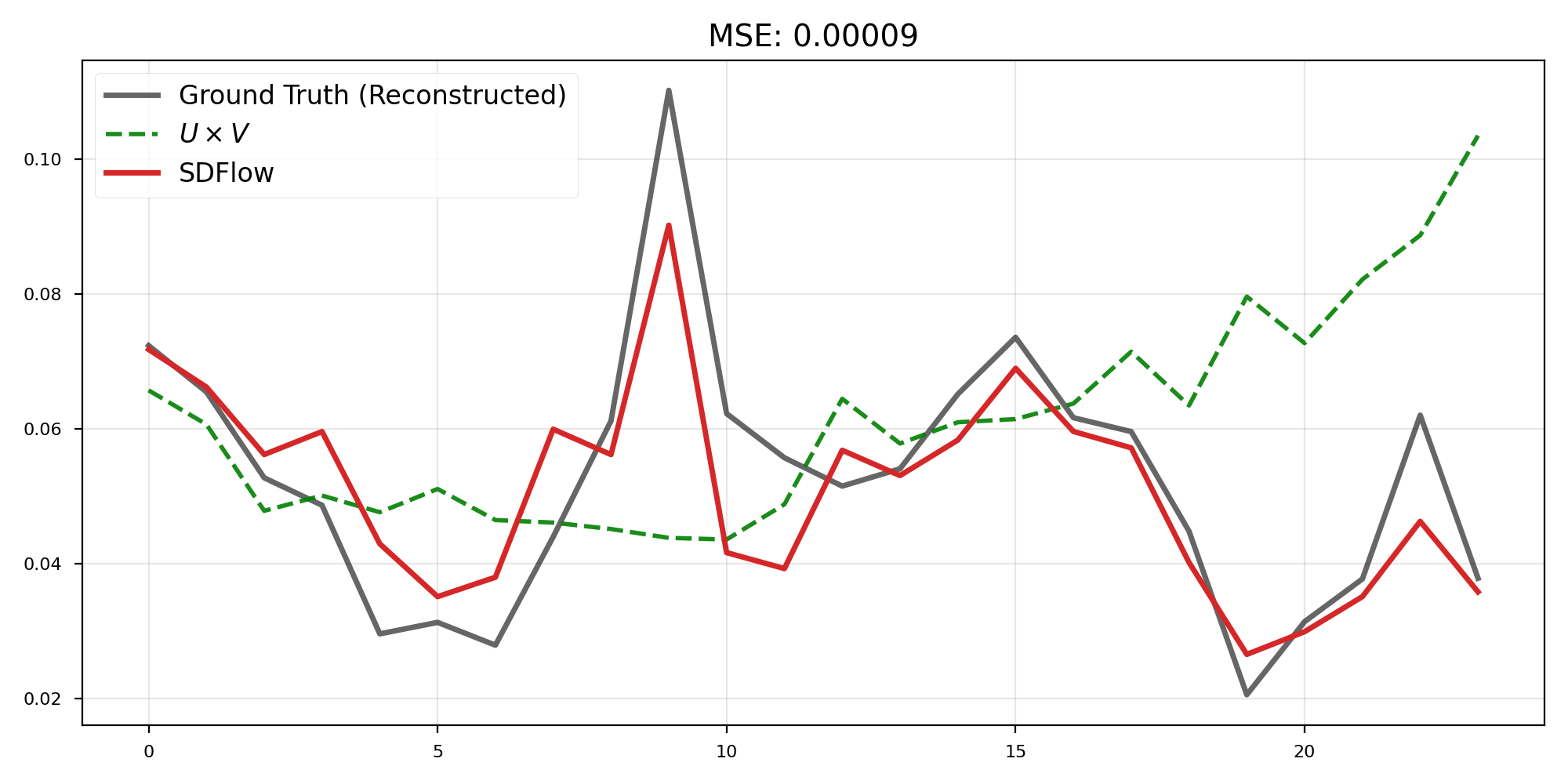}
		\end{subfigure}
		\hfill
		\begin{subfigure}[b]{0.3\textwidth}
			\centering
			\includegraphics[width=\textwidth]{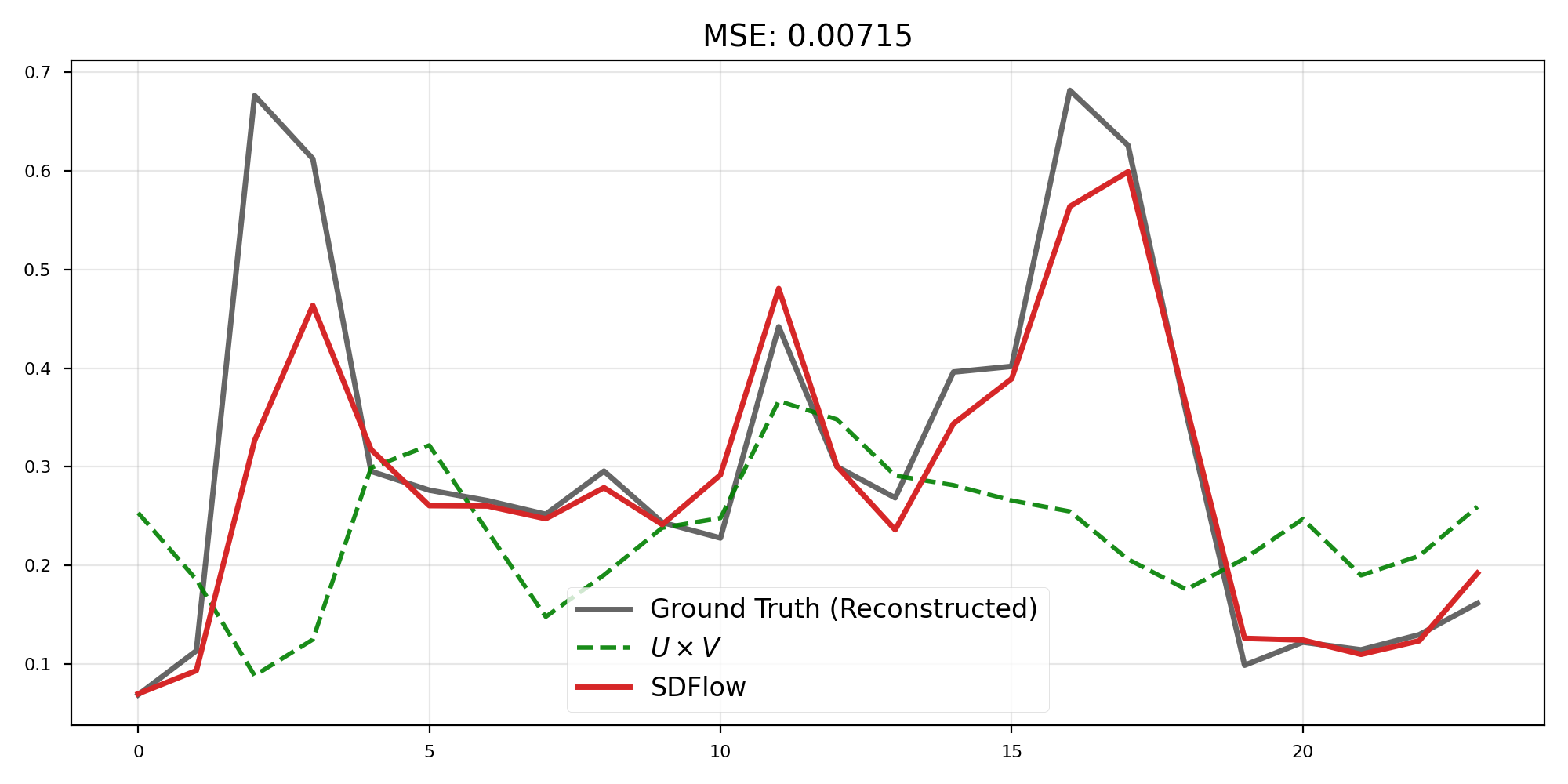}
		\end{subfigure}
		
		\vspace{0.1em}
		
		\begin{subfigure}[b]{0.3\textwidth}
			\centering
			\includegraphics[width=\textwidth]{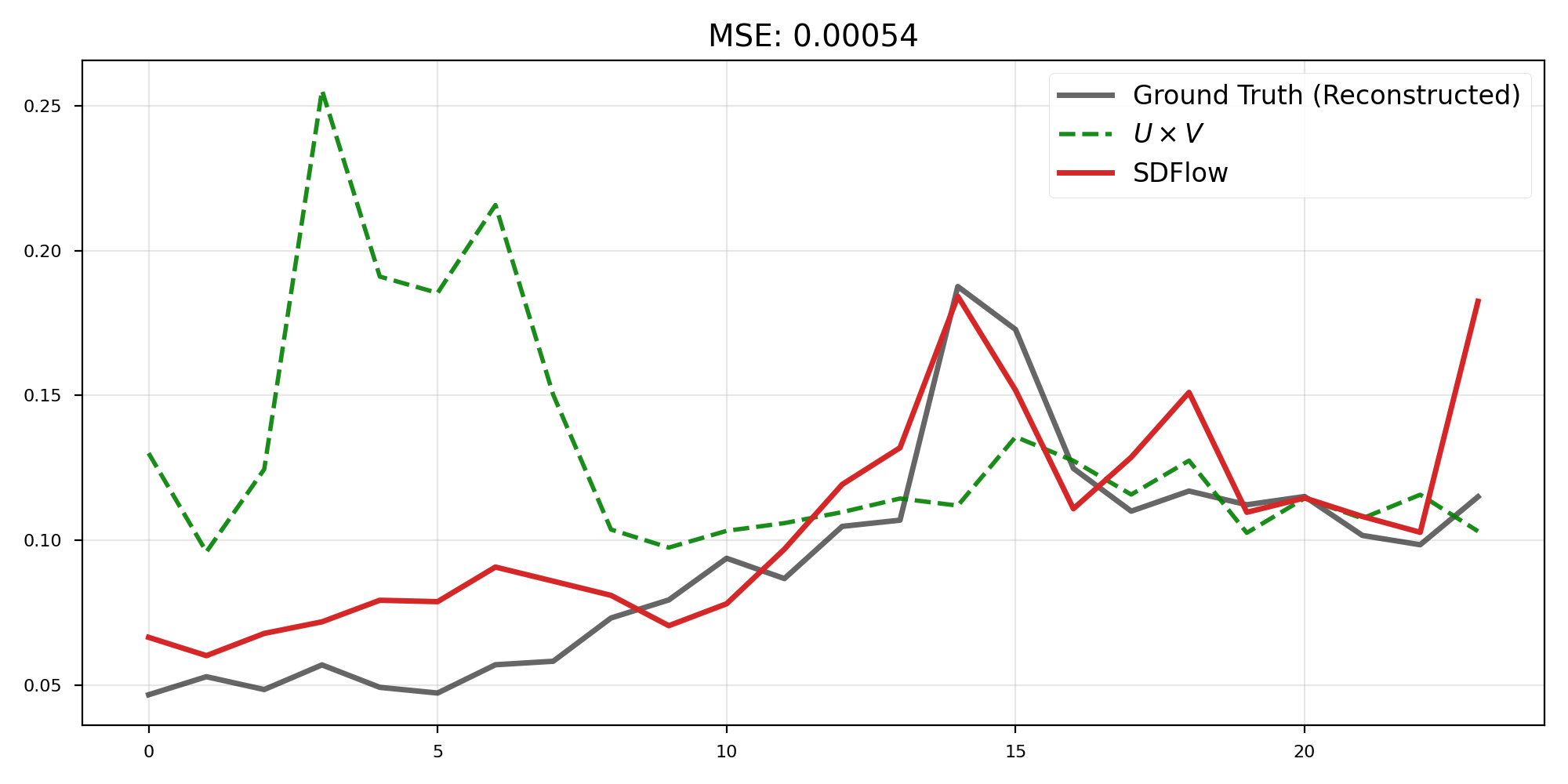}
		\end{subfigure}
		\hfill
		\begin{subfigure}[b]{0.3\textwidth}
			\centering
			\includegraphics[width=\textwidth]{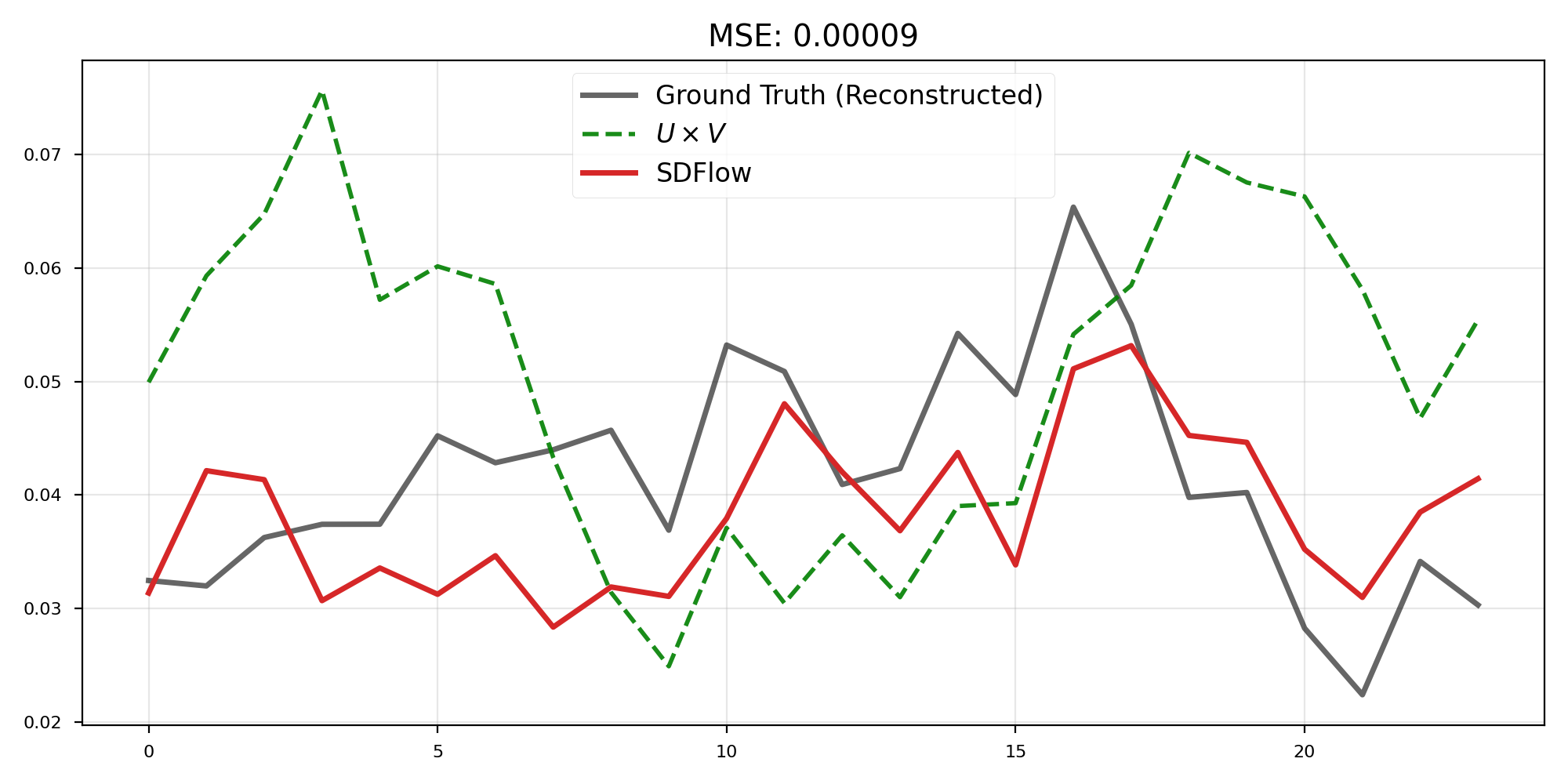}
		\end{subfigure}
		\hfill
		\begin{subfigure}[b]{0.3\textwidth}
			\centering
			\includegraphics[width=\textwidth]{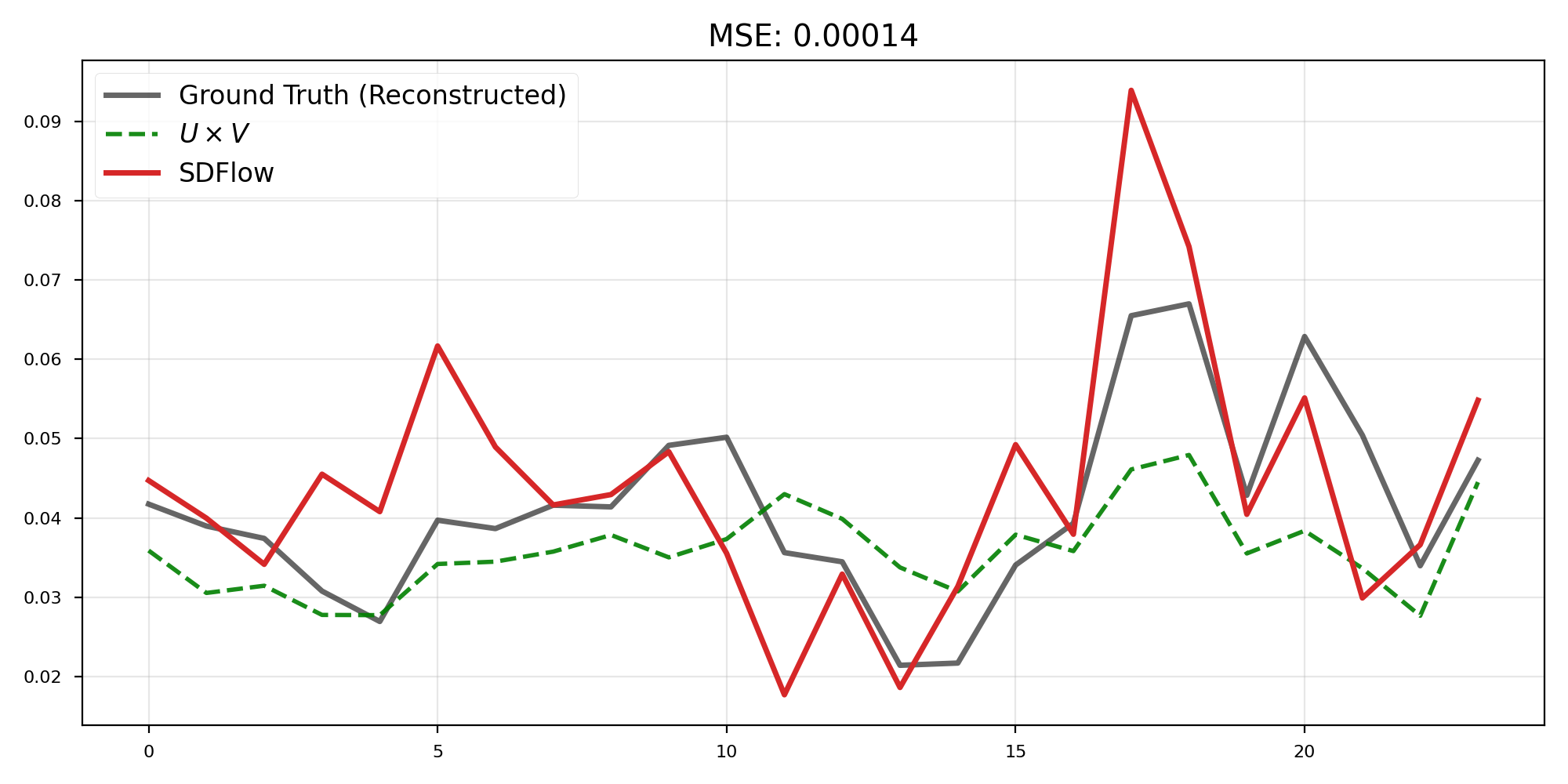}
		\end{subfigure}
		
		\vspace{0.1em}
		
		\begin{subfigure}[b]{0.3\textwidth}
			\centering
			\includegraphics[width=\textwidth]{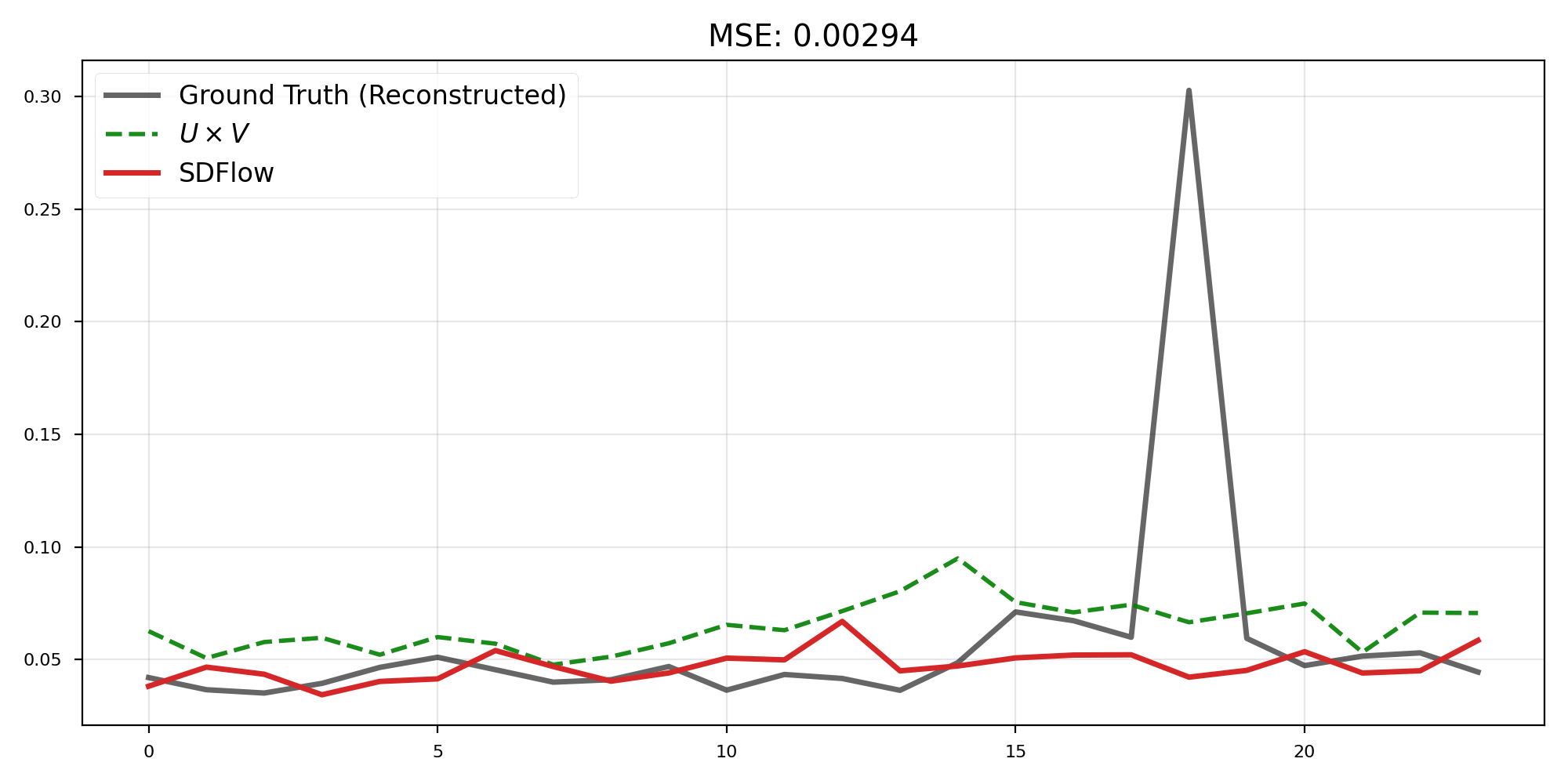}
		\end{subfigure}
		\hfill
		\begin{subfigure}[b]{0.3\textwidth}
			\centering
			\includegraphics[width=\textwidth]{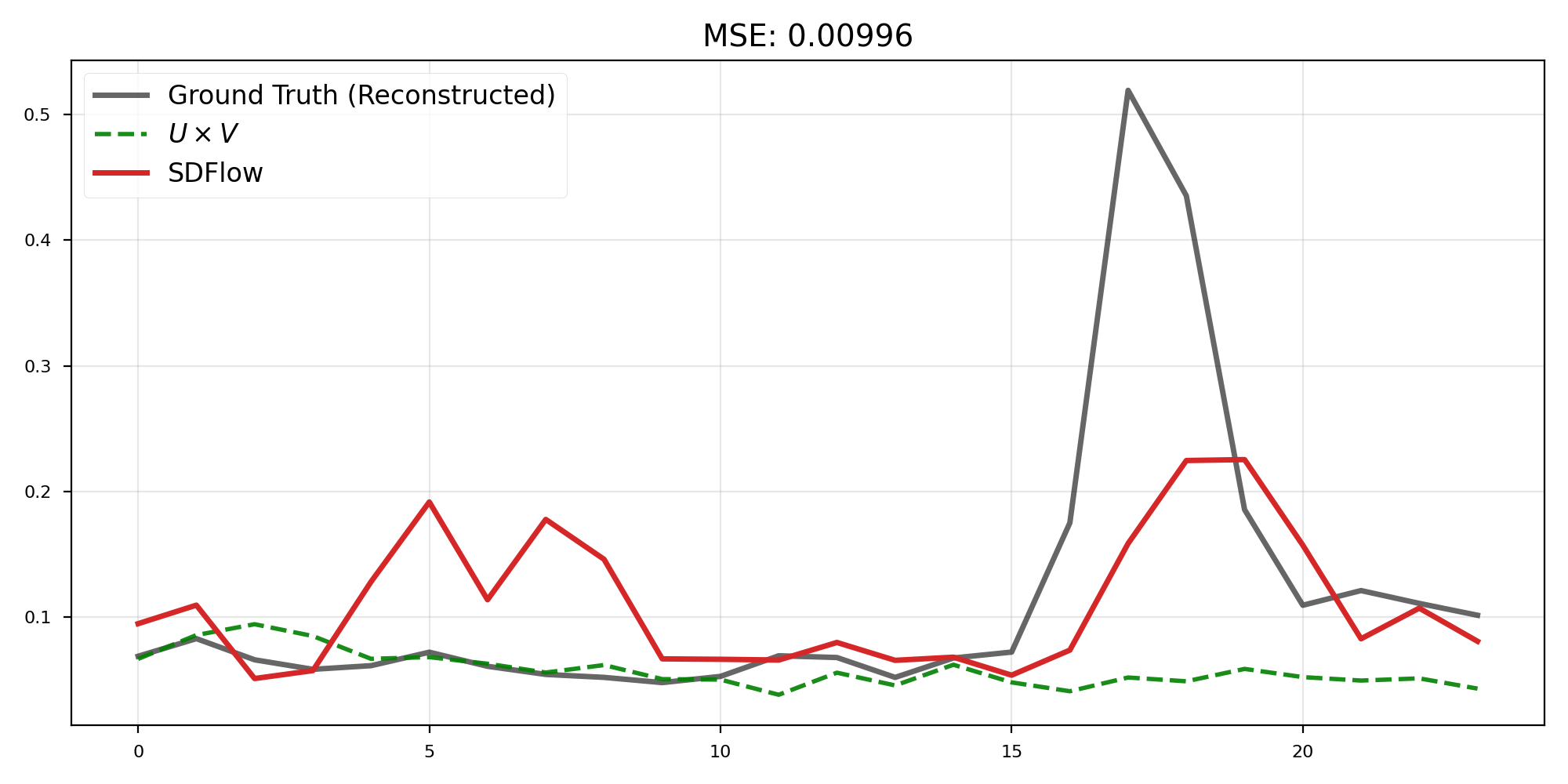}
		\end{subfigure}
		\hfill
		\begin{subfigure}[b]{0.3\textwidth}
			\centering
			\includegraphics[width=\textwidth]{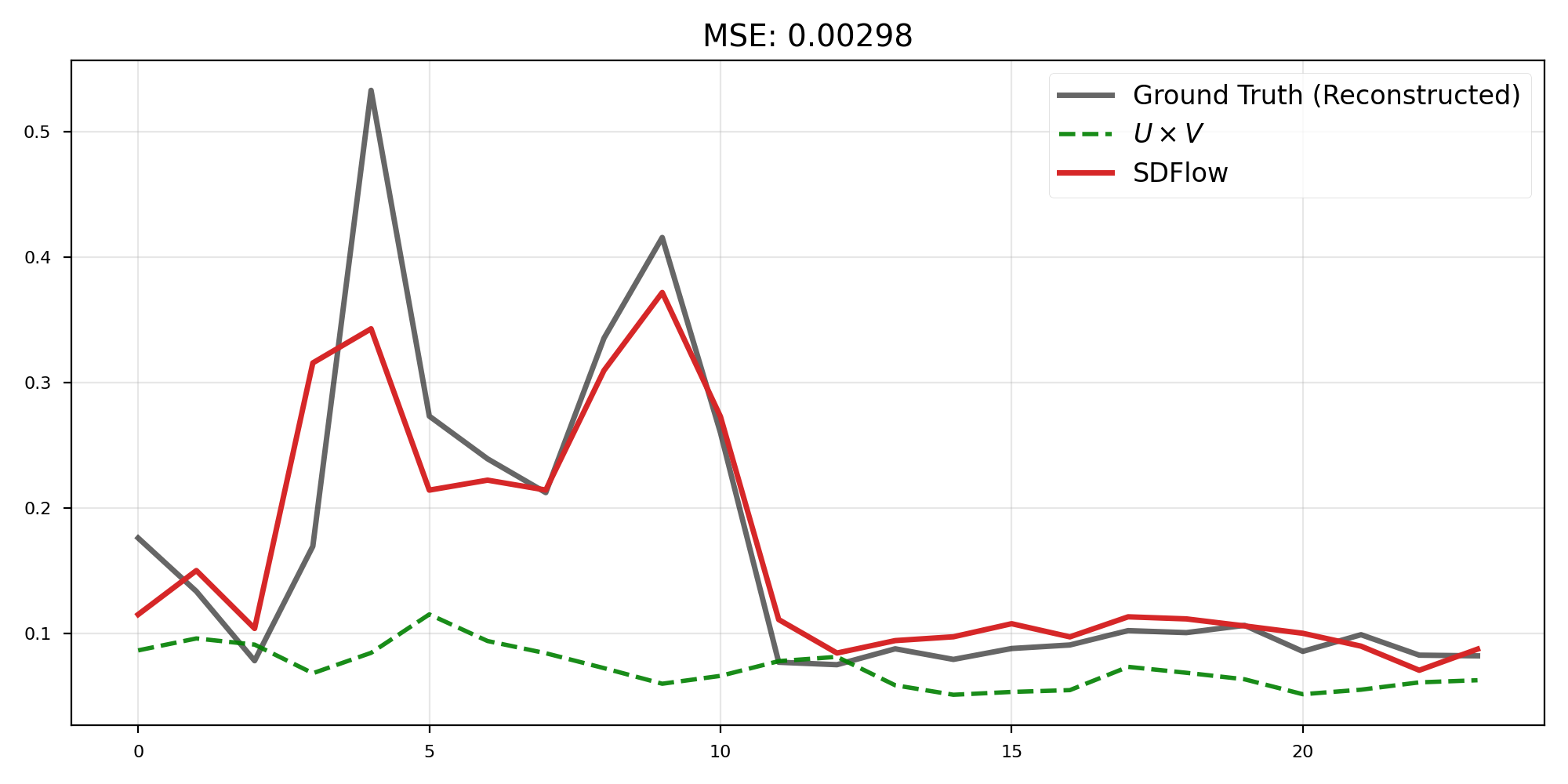}
		\end{subfigure}
		
		\caption{Visualizations of time series reconstruction samples using real coordinates instead of the anchor-prior sampler.}
		\label{fig:vis_recon}
	\end{figure}

	To assess our model's capability in reconstructing the real indices via the specific $u_s$, we conducted an experiment by sampling directly from the true $u_s$ (bypassing the anchor-prior sampler) and processing these samples through the DiT to generate the final indices. The results in Figure~\ref{fig:vis_recon} show low MSE while retaining flexibility, supporting the model's ability to cover the latent manifold.
	
	\section{Conditional Generation: Forecasting}
	\label{app:conditional}
	Beyond generation, SDFlow naturally extends to forecasting by conditioning on historical latent coordinates. We achieve the conversion from unconditional to conditional by concatenating the low-dimensional manifold samples we have learned with real conditions. Specifically, a complete sequence is sampled from the manifold distribution, then the first half is replaced with real history, and finally the second half is refined through dynamics to generate a coherent future prediction.
	
\begin{figure}[h]
		\centering
		\begin{subfigure}{0.32\textwidth}
			\includegraphics[width=\textwidth]{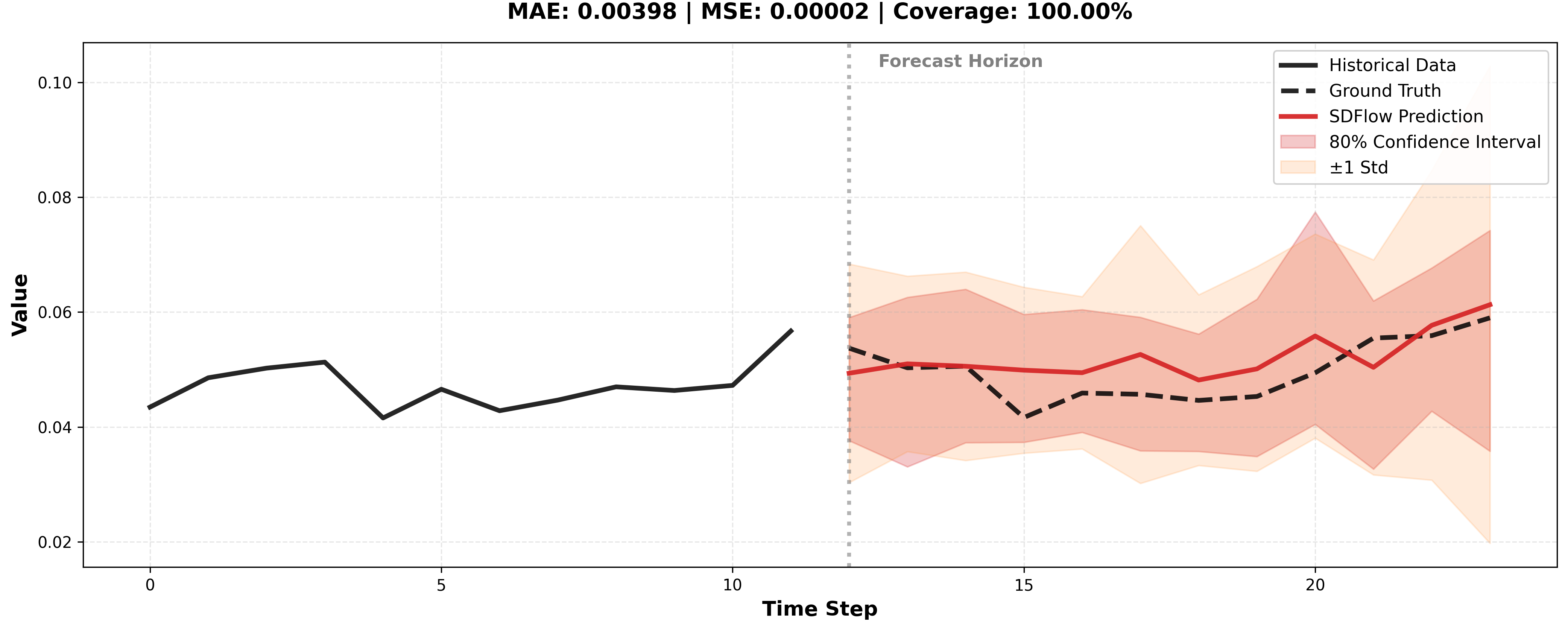}
		\end{subfigure}
		\hfill
		\begin{subfigure}{0.32\textwidth}
			\includegraphics[width=\textwidth]{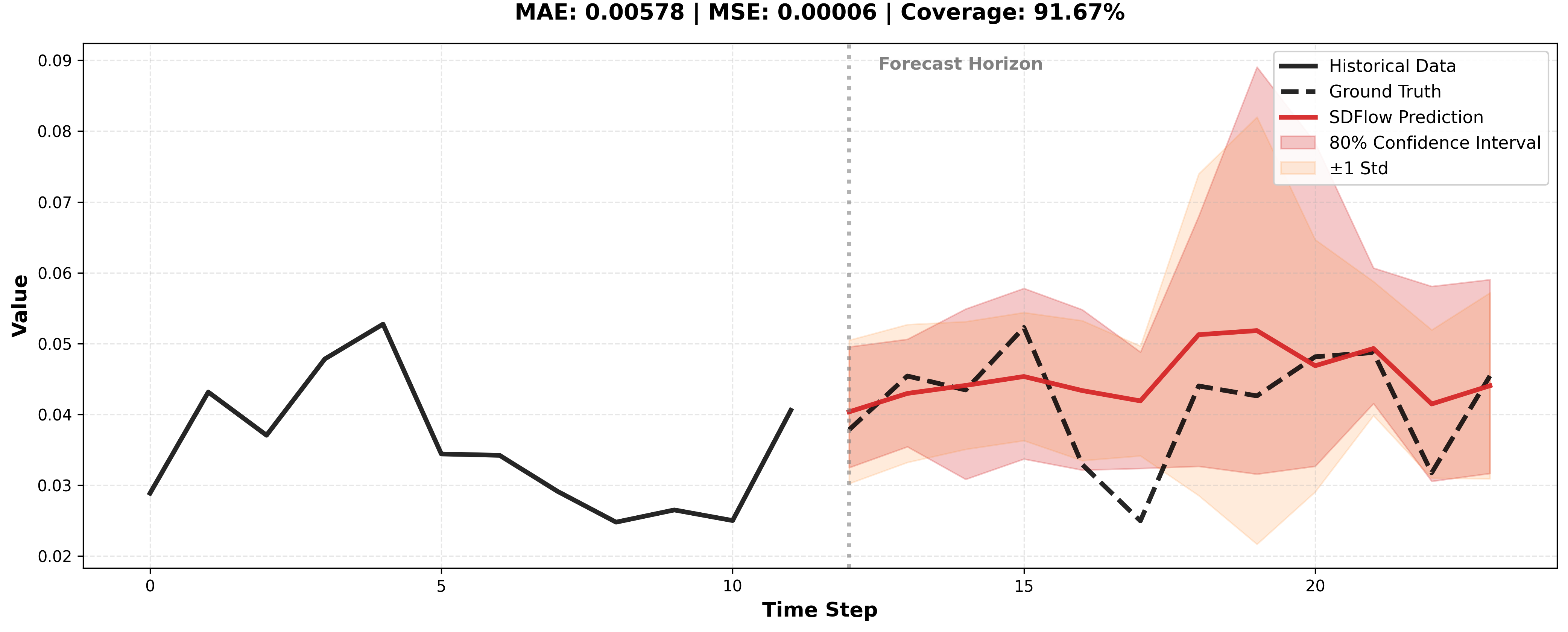}
		\end{subfigure}
		\hfill
		\begin{subfigure}{0.32\textwidth}
			\includegraphics[width=\textwidth]{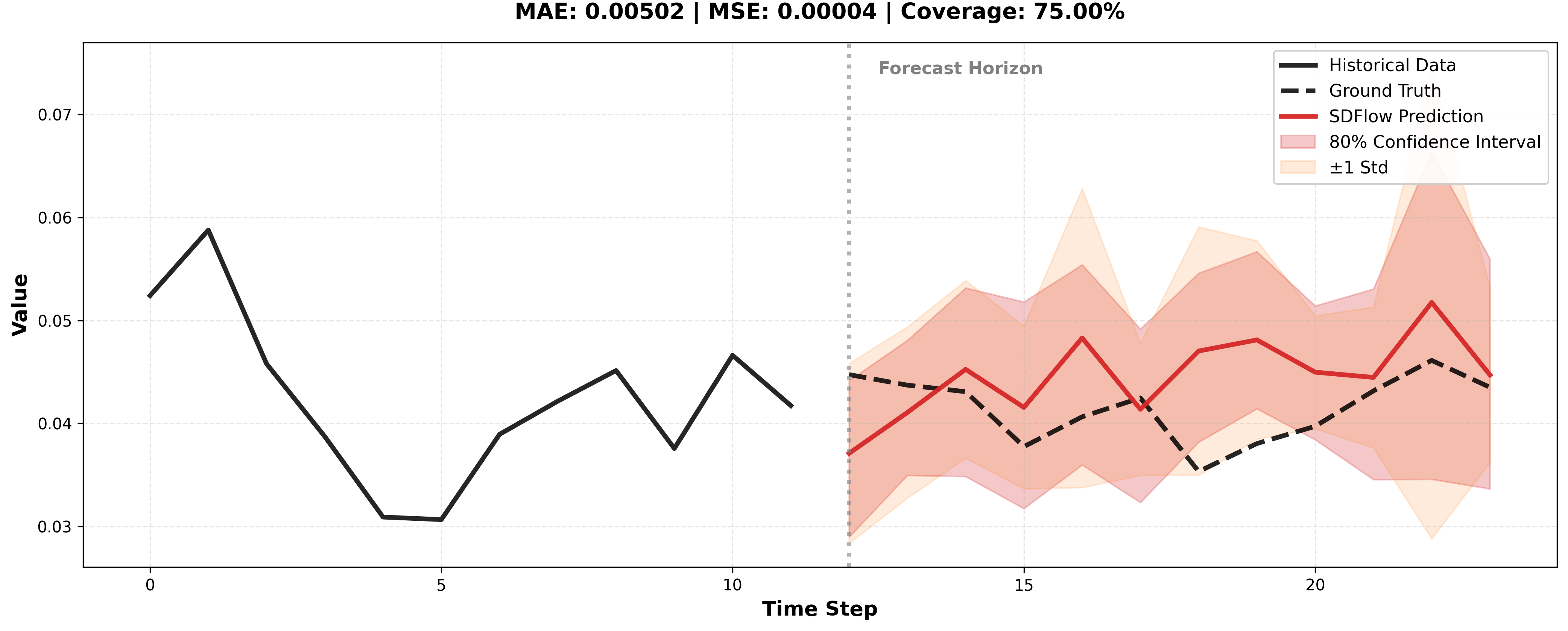}
		\end{subfigure}
		\caption{Forecasting Results (Energy, $12 \rightarrow 12$).}
		\label{fig:forecast}
		\vspace{-2mm}
	\end{figure}

	Figure~\ref{fig:forecast} demonstrates zero-shot forecasting where SDFlow predicts future time steps given only the first half.  Despite no forecasting-specific training, our method achieves great MAE and MSE, with well-calibrated 80\% confidence intervals (coverage 93\%).
	
	This strong performance supports our hypothesis that the learned manifold structure captures not only spatial but also temporal dependencies of time series data. However, more specific modifications in model architecture may further benefit Time Series Forecasting, which is a promising future direction.
	
	\section{Additional Ablations}
	\label{app:additional_ablations}
	
	\textbf{Categorical vs. Continuous Supervision.}
	We compare our categorical cross-entropy objective against continuous MSE regression on embeddings. As shown in Table~\ref{tab:ce_mse}, the categorical objective yields superior performance. Beyond performance gains, it offers crucial theoretical advantages: (1) explicit discrete supervision aligned with the VQ structure; (2) temperature-controlled sampling at inference enabling quality-diversity trade-offs; and (3) interpretable uncertainty quantification over codebook indices.
	
	\begin{table}[h]
		\caption{Comparison of categorical cross-entropy (Ours) against continuous MSE regression.}
		\label{tab:ce_mse}
		\centering
		\small
		\begin{tabular}{@{}lcc@{}}
			\toprule
			\textbf{Objective} & \textbf{DS}-Sines $\downarrow$ & \textbf{DS}-ETTh $\downarrow$\\
			\midrule
			MSE (continuous) & 0.009 & 0.005\\
			\textbf{CE (categorical)} & \textbf{0.006} & \textbf{0.002}\\
			\bottomrule
		\end{tabular}
	\end{table}

	\textbf{Rank Sensitivity.}
	Table~\ref{tab:rank_ablation} analyzes sensitivity to the rank parameter on Energy ($L=24$). Performance peaks at $r=256$, which captures 99\% of variance while maintaining parameter efficiency. Lower ranks underfit the manifold structure, while higher ranks provide diminishing returns. Notably, relying solely on SVD to determine optimal rank is ineffective, demonstrating the necessity of the learnable low-rank decomposition.
	
	\begin{table}[t]
		\caption{Rank sensitivity analysis on Energy ($L=24$). Performance peaks at $r=256$, which captures 99\% of variance while maintaining parameter efficiency.}
		\label{tab:rank_ablation}
		\centering
		\small
		\resizebox{0.45\columnwidth}{!}{
			\begin{tabular}{@{}lccc@{}}
				\toprule
				\textbf{Rank} & \textbf{DS} $\downarrow$ & \textbf{FID} $\downarrow$ & \textbf{Parameters} \\
				\midrule
				$r=32$ & 0.149 $\pm$ .012 & 0.042 $\pm$ .005 & 0.6M \\
				SVD $(r=42)$ & 0.094 $\pm$ .009 & 0.017 $\pm$ .002 & 0.8M \\
				$r=64$ & 0.062 $\pm$ .008 & 0.015 $\pm$ .002 & 1.2M \\
				$r=128$ & 0.015 $\pm$ .005 & 0.006 $\pm$ .001 & 2.4M \\
				$r=256$ & \textbf{0.006 $\pm$ .002} & \textbf{0.001 $\pm$ .000} & 4.8M \\
				$r=512$ & 0.006 $\pm$ .003 & 0.001 $\pm$ .000 & 9.6M \\
				\bottomrule
			\end{tabular}
		}
		\vspace{-0.2cm}
	\end{table}
	
	\textbf{ODE Integration Steps.}
	Table~\ref{tab:steps_ablation} examines sensitivity to the number of ODE integration steps. 20 steps suffice for convergence; additional steps provide negligible improvement at increased computational cost.
	
	\begin{table}[t]
		\caption{ODE integration steps sensitivity on Energy ($L=24$). 20 steps suffice for convergence; more steps provide negligible improvement.}
		\label{tab:steps_ablation}
		\centering
		\small
		\resizebox{0.4\columnwidth}{!}{
			\begin{tabular}{@{}lccc@{}}
				\toprule
				\textbf{Steps} & \textbf{DS} $\downarrow$ & \textbf{FID} $\downarrow$ & \textbf{Time (s)} \\
				\midrule
				10 & 0.018 $\pm$ .006 & 0.008 $\pm$ .002 & 0.005 \\
				15 & 0.009 $\pm$ .003 & 0.003 $\pm$ .001 & 0.011 \\
				20 & \textbf{0.006 $\pm$ .002} & \textbf{0.001 $\pm$ .000} & 0.021 \\
				50 & 0.006 $\pm$ .002 & 0.001 $\pm$ .000 & 0.042 \\
				\bottomrule
			\end{tabular}
		}
		
	\end{table}
	
	\begin{table}[H]
		\caption{Cross-dataset ablation ($L=24$). Both the low-rank anchor scaffold and kernel-smoothed prior are necessary on complex datasets (Energy, ETTh), while simpler datasets (Sines) are more tolerant.}
		\label{tab:ablation_full}
		\centering
		\small
		\resizebox{0.6\columnwidth}{!}{
			\begin{tabular}{@{}ll cccc@{}}
				\toprule
				\textbf{Configuration} & \textbf{Prior} & \textbf{Sines} & \textbf{Stocks} & \textbf{ETTh} & \textbf{Energy} \\
				\midrule
				w/o Low-Rank & Gaussian & 0.007 $\pm$ .006 & 0.038 $\pm$ .009 & 0.100 $\pm$ .015 & 0.217 $\pm$ .010 \\
				w/o Low-Rank & Anchor Prior & 0.006 $\pm$ .008 & 0.025 $\pm$ .010 & 0.074 $\pm$ .017 & 0.173 $\pm$ .015 \\
				\midrule
				w/ Low-Rank & Gaussian & 0.008 $\pm$ .005 & 0.031 $\pm$ .008 & 0.089 $\pm$ .012 & 0.218 $\pm$ .004 \\
				\textbf{w/ Low-Rank} & \textbf{Anchor Prior} & \textbf{0.006 $\pm$ .006} & \textbf{0.003 $\pm$ .003} & \textbf{0.002 $\pm$ .003} & \textbf{0.006 $\pm$ .002} \\
				\bottomrule
			\end{tabular}
		}
	\end{table}
	
	\begin{figure}[H]
		\centering
		\begin{subfigure}{0.4\textwidth}
			\includegraphics[width=\textwidth]{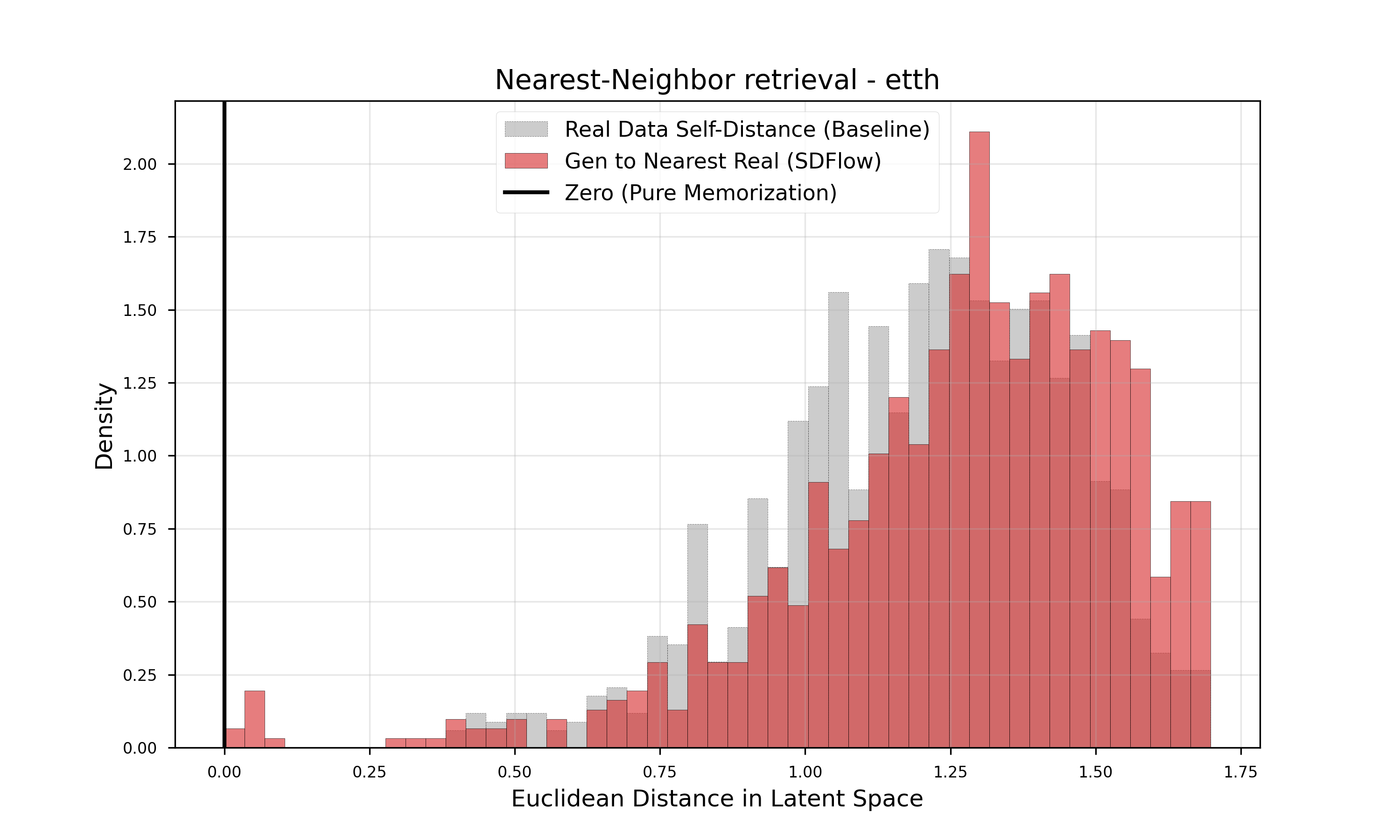}
			\caption{ETTh dataset}
			\label{fig:nn_etth}
		\end{subfigure}
		\hfill
		\begin{subfigure}{0.4\textwidth}
			\includegraphics[width=\textwidth]{figure/nearest/energy_nearest_neighbor.png}
			\caption{Energy dataset}
			\label{fig:nn_energy}
		\end{subfigure}
		\caption{Nearest-neighbor distance analysis. Gray bars show the self-distance distribution among real training samples (baseline), while red bars show distances from generated samples to their nearest training neighbors. Generated samples maintain substantial distances from training data (mean $>0.8$ for ETTh, $>2.0$ for Energy), consistent with non-copying behavior.}
		\label{fig:nn_analysis}
		\vspace{-2mm}
	\end{figure}
	
	\textbf{Cross-Dataset Ablation.}
	Table~\ref{tab:ablation_full} extends the ablation study across all datasets. Both low-rank decomposition and the anchor prior are essential on complex datasets (Energy, ETTh), while simpler datasets (Sines) are more tolerant of suboptimal configurations.

	\textbf{Nearest-neighbor distance analysis.}
	Finally, to check whether generated sequences collapse to training examples, we conduct nearest-neighbor retrieval against the training set. Figure~\ref{fig:nn_analysis} shows the distribution of Euclidean distances from generated samples to their nearest training neighbors in the VQ-VAE latent space.

	Together, the held-out latent-flow, prior-ablation, and nearest-neighbor results suggest that the anchor prior is not merely copying individual training samples. Instead, the learned manifold structure provides a \textit{geometric scaffold} that guides generation while preserving diversity.

	
	\newpage
	
	\section*{NeurIPS Paper Checklist}

\begin{enumerate}
	
	\item {\bf Claims}
	\item[] Question: Do the main claims made in the abstract and introduction accurately reflect the paper's contributions and scope?
	\item[] Answer: \answerYes{}
	\item[] Justification: The abstract and Introduction (Section~\ref{sec:intro}) state the scope and contributions of \method{}, and the Method and Experiments sections support these claims with the proposed low-rank categorical flow formulation and benchmark results (Sections~\ref{sec:method}--\ref{sec:experiments}).
	
	\item {\bf Limitations}
	\item[] Question: Does the paper discuss the limitations of the work performed by the authors?
	\item[] Answer: \answerYes{}
	\item[] Justification: Section~\ref{sec:conclusion} includes a limitations paragraph noting that the current VQ tokenizer and codebook are dataset-specific, with universal time-series tokenization left as future work.
	
	\item {\bf Theory assumptions and proofs}
	\item[] Question: For each theoretical result, does the paper provide the full set of assumptions and a complete (and correct) proof?
	\item[] Answer: \answerYes{}
	\item[] Justification: Theorem~\ref{thm:transport_distance} and Proposition~\ref{prop:consistency} state their assumptions, while Appendices~\ref{app:proof_transport_distance} and~\ref{app:proof_consistency} give complete derivations.
	
	\item {\bf Experimental result reproducibility}
	\item[] Question: Does the paper fully disclose all the information needed to reproduce the main experimental results of the paper to the extent that it affects the main claims and/or conclusions of the paper (regardless of whether the code and data are provided or not)?
	\item[] Answer: \answerYes{}
	\item[] Justification: The paper specifies the datasets, metrics, baselines, tokenizer/model settings, random seeds, hyperparameter tables, and training/inference algorithms in Sections~\ref{sec:experiments} and Appendices~\ref{app:setup}--\ref{app:algorithm}.

	\item {\bf Open access to data and code}
	\item[] Question: Does the paper provide open access to the data and code, with sufficient instructions to faithfully reproduce the main experimental results, as described in supplemental material?
	\item[] Answer: \answerYes{}
	\item[] Justification: A code repository is provided in the abstract. The repository includes implementation code, configuration files, data-preparation instructions, and scripts for reproducing the main experiments; the datasets used are public benchmarks.

	\item {\bf Experimental setting/details}
	\item[] Question: Does the paper specify all the training and test details (e.g., data splits, hyperparameters, how they were chosen, type of optimizer) necessary to understand the results?
	\item[] Answer: \answerYes{}
	\item[] Justification: Sections~\ref{sec:experiments} and Appendix~\ref{app:setup} describe datasets, metrics, baselines, tokenizer settings, ranks, layers, bandwidths, seeds, and hyperparameter configurations; Appendix~\ref{app:algorithm} provides training and inference procedures.
	
	\item {\bf Experiment statistical significance}
	\item[] Question: Does the paper report error bars suitably and correctly defined or other appropriate information about the statistical significance of the experiments?
	\item[] Answer: \answerYes{}
	\item[] Justification: Section~\ref{sec:experiments} states that all experiments report mean and standard deviation over five random seeds, and Tables~\ref{tab:main_short}--\ref{tab:main_long} report mean $\pm$ std over five runs.
	
	\item {\bf Experiments compute resources}
	\item[] Question: For each experiment, does the paper provide sufficient information on the computer resources (type of compute workers, memory, time of execution) needed to reproduce the experiments?
	\item[] Answer: \answerNo{}
	\item[] Justification: The manuscript does not specify compute worker types, memory, per-run wall-clock time, or total compute needed to reproduce the experiments.
	
	\item {\bf Code of ethics}
	\item[] Question: Does the research conducted in the paper conform, in every respect, with the NeurIPS Code of Ethics \url{https://neurips.cc/public/EthicsGuidelines}?
	\item[] Answer: \answerYes{}
	\item[] Justification: The study uses synthetic or standard time-series benchmarks described in Appendix~\ref{app:setup}, does not involve human subjects, and no deviation from the NeurIPS Code of Ethics is apparent from the described protocol.

	\item {\bf Broader impacts}
	\item[] Question: Does the paper discuss both potential positive societal impacts and negative societal impacts of the work performed?
	\item[] Answer: \answerNo{}
	\item[] Justification: The Introduction motivates time-series generation in domains such as finance, healthcare, and energy, but the current manuscript does not discuss both positive and negative societal impacts or mitigation strategies.
	
	\item {\bf Safeguards}
	\item[] Question: Does the paper describe safeguards that have been put in place for responsible release of data or models that have a high risk for misuse (e.g., pre-trained language models, image generators, or scraped datasets)?
	\item[] Answer: \answerNA{}
	\item[] Justification: The paper does not release a high-risk model, scraped dataset, or other asset requiring controlled-access safeguards; the experiments use standard time-series benchmarks.
	
	\item {\bf Licenses for existing assets}
	\item[] Question: Are the creators or original owners of assets (e.g., code, data, models), used in the paper, properly credited and are the license and terms of use explicitly mentioned and properly respected?
	\item[] Answer: \answerNo{}
	\item[] Justification: Appendix~\ref{app:setup} names the Sines, Stocks, ETTh, and Energy datasets and cites prior protocols, but the manuscript does not state their licenses or terms of use.
	
	\item {\bf New assets}
	\item[] Question: Are new assets introduced in the paper well documented and is the documentation provided alongside the assets?
	\item[] Answer: \answerYes{}
	\item[] Justification: The paper releases an anonymized code asset. Documentation, environment requirements, data preparation, and reproduction commands are provided with the released repository; no new dataset or model checkpoint is introduced.
	
	\item {\bf Crowdsourcing and research with human subjects}
	\item[] Question: For crowdsourcing experiments and research with human subjects, does the paper include the full text of instructions given to participants and screenshots, if applicable, as well as details about compensation (if any)? 
	\item[] Answer: \answerNA{}
	\item[] Justification: The experiments use existing/synthetic time-series datasets and do not involve crowdsourcing or human-subject studies.
	
	\item {\bf Institutional review board (IRB) approvals or equivalent for research with human subjects}
	\item[] Question: Does the paper describe potential risks incurred by study participants, whether such risks were disclosed to the subjects, and whether Institutional Review Board (IRB) approvals (or an equivalent approval/review based on the requirements of your country or institution) were obtained?
	\item[] Answer: \answerNA{}
	\item[] Justification: No crowdsourcing or human-subject study is conducted, so participant risks and IRB approval are not applicable.
	
	\item {\bf Declaration of LLM usage}
	\item[] Question: Does the paper describe the usage of LLMs if it is an important, original, or non-standard component of the core methods in this research? Note that if the LLM is used only for writing, editing, or formatting purposes and does \emph{not} impact the core methodology, scientific rigor, or originality of the research, declaration is not required.
	\item[] Answer: \answerNA{}
	\item[] Justification: The core method is based on VQ-VAE tokenization and categorical flow matching, and the manuscript does not use LLMs as a core methodological component.
	
\end{enumerate}

\end{document}